%% file: main.tex
\documentclass{article}
\usepackage{amsmath}

\usepackage{arxiv}

\usepackage[utf8]{inputenc} 
\usepackage[T1]{fontenc}    
\usepackage{hyperref}       
\usepackage{url}            
\usepackage{booktabs}       
\usepackage{amsfonts}       
\usepackage{nicefrac}       
\usepackage{microtype}      
\usepackage{cleveref}       
\usepackage{lipsum}         
\usepackage{graphicx}
\usepackage{doi}
\usepackage{subcaption}
\usepackage{ragged2e}

\title{\input{section/title}}
\usepackage{authblk}

\newbox{\orcid}\sbox{\orcid}{\includegraphics[scale=0.06]{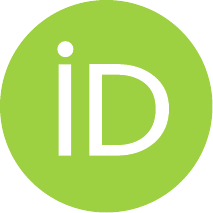}} 
\author[1]{%
	{\hspace{1mm}Tatthapong~Srikitrungruang}%
}
\author[1]{%
	\href{https://orcid.org/0000-0001-7533-1865}{\usebox{\orcid}\hspace{1mm}Jaesung~Lee\thanks{\texttt{j.lee@tamu.edu}}}%
}

\affil[1]{Wm Michael Barnes ’64 Department of Industrial and Systems Engineering, Texas A\&M University, College Station, TX, 77843, USA}
\renewcommand{\shorttitle}{\input{Contents/_title_short}}
\hypersetup{
pdftitle={\input{section/title}},
pdfsubject={},
pdfauthor={Tatthapong~Srikitrungruang, Jaesung~Lee},
}

\date{}

\input{Contents/_prefix}

\begin{document}

\title{\input{Contents/_title}}

\maketitle
\begin{abstract}

     \large\input{Contents/0_abstract}

\end{abstract}


\clearpage

\setcounter{page}{1}

\input{Contents/_document}

\newpage

\bibliographystyle{agsm}
\begingroup
    \setstretch{1}
    \bibliography{Contents/_ref}
\endgroup

\newpage

\input{Contents/9__supplementary}

\end{document}

%% file: Contents/_title_short.tex
Probabilistic Physics-Informed Neural Networks (PIE-PINN)

%% file: Contents/_prefix.tex
\input{Contents/_definition}

\usepackage{amsmath}
\usepackage{amsfonts}
\usepackage{bm}
\usepackage{graphicx}
\usepackage{enumerate}
\usepackage{xcolor}
\usepackage{natbib} 
\usepackage{url} 
\usepackage{xcolor}
\usepackage{tcolorbox} 
\usepackage{tocloft} 
\usepackage{ragged2e}
\usepackage{subcaption} 
\usepackage{multirow}
\usepackage{booktabs} 
\usepackage{float} 
\usepackage{multirow} 
\usepackage{makecell} 
\usepackage{array}
\usepackage{setspace}
\usepackage{float}
\usepackage{algorithm}
\usepackage{algorithmic}
\usepackage{sansmath}
\usepackage{etoolbox}

%% file: Contents/_definition.tex
\newcommand{\Lap}{\mathrm{Laplace}}

\newcommand{\X}{\mathbf{x}}

\newcommand{\nDisp}{N_u}
\newcommand{\nStrain}{N_\varepsilon}
\newcommand{\nPDE}{N_f}

\newcommand{\nnParaU}{\bm{\theta}_u}
\newcommand{\nnParae}{\bm{\theta}_\varepsilon}
\newcommand{\nnParaE}{\bm{\theta}_E}
\newcommand{\splParaU}{\mathbf{C}}

\newcommand{\LOSS}{{\mathcal{L}}}

\newcommand{\bPDE}{\bm{b}_f}
\newcommand{\bSTRAIN}{\bm{b}_\varepsilon}

\newcommand{\uEta}{\bm{\eta}}
\newcommand{\uTau}{\bm{\tau}}
\newcommand{\bDISP}{\bm{b}_{u}}
\newcommand{\bDISPi}{\bm{b}_{u,i}}

\newcommand{\sbPDE}{{b}_{f,j}}
\newcommand{\sbSTRAIN}{{b}_{\varepsilon,k}}

\newcommand{\suEta}{{\eta_{i,j}}}
\newcommand{\suTau}{{\tau_j}}
\newcommand{\sbDISP}{{b}_{u,i,j}}

\newcommand{\CAUHCHY}{\mathcal{C}^+}

\newcommand{\suEtaSqn}{{\eta^2_{i,j}}}

\newcommand{\suTauSqn}{{\tau^2_j}}

\newcommand{\diag}{\operatorname{diag}}

%% file: Contents/_title.tex
Probabilistic Physics-Informed Neural Networks for Estimating Heterogeneous Elastic Properties from Low-Resolution and Noisy Displacement Data

%% file: Contents/0_abstract.tex
Estimating spatially heterogeneous elastic properties from low-resolution displacement measurements is a severely ill-posed inverse elasticity problem because low resolution obscures spatial details needed to distinguish heterogeneous property variations, and small measurement perturbations or fitting errors are amplified through inverse estimation. Existing inverse methods often rely on high-fidelity observations and manually prespecified loss weights, limiting their adaptability and making them sensitive to noise and resolution degradation. We propose a Probabilistic Inverse Elasticity Physics-Informed Neural Network (PIE-PINN) framework for robust estimation of Young's modulus and Poisson's ratio from noisy, low-resolution displacement data. PIE-PINN models displacement observation, strain-discrepancy, and equilibrium residuals using Laplace distributions within a unified probabilistic model. To improve robustness, the framework combines a B-spline-guided displacement network with a hierarchical half-Cauchy model for displacement residual scales. The B-spline provides a smooth global representation of the displacement field, while the neural network correction captures local variations. The hierarchical scale model adaptively downweights severe displacement fitting errors, enabling more robust recovery of the latent mean displacement field. An alternating maximum-likelihood training strategy updates the mean through weighted residual minimization and updates the scales to adjust the loss weights. Systematic case studies across varying noise levels and observation resolutions demonstrate the robustness of PIE-PINN.

%% file: Contents/_keyword.tex
Inverse problem, physic-informed neural network, heterogeneous elasticity estimation

%% file: Contents/_document.tex
\input{Contents/1_introduction}

\input{Contents/2__background}

\input{Contents/3__model}

\input{Contents/4_ParamEst}
\input{Contents/4_caseStudies}

\input{Contents/5_conclusion}
\input{Contents/6_suffix}

%% file: Contents/1_introduction.tex
\section{Introduction} \label{sec:intro}

Accurate estimation of elastic properties, particularly Young’s modulus and Poisson’s ratio, is essential for various applications, including medical imaging and material characterization. Young’s modulus determines a material’s resistance to elastic deformation, while Poisson’s ratio describes the relationship between axial and transverse strains. When external loads are applied, elastic materials develop internal stresses, leading to displacement fields that are directly influenced by the distribution of these elastic properties. Reliable estimation of elastic properties is crucial for several tasks, such as diagnosing diseases (including cancer detection \citep{Goenezen2012, Quan2016, Wu2019} and identifying cardiovascular abnormalities \citep{boutouyrie2021arterial}), assessing structural integrity in bone health \citep{wang2013trabecular}, and evaluating additively manufactured components\citep{Choren2013, Roy2020, Yan2024}. In practical scenarios, elastic properties are often spatially heterogeneous, posing significant challenges for conventional estimation techniques. This spatial heterogeneity necessitates advanced computational methods capable of accurately recovering elastic property distributions from noisy and limited observations.

Elasticity estimation techniques can generally be classified into two primary approaches: direct and indirect methods. The direct approach applies a known force directly to the surface of a material and infers its elasticity from the resulting deformation. Well-known examples include nanoindentation \citep{Oliver1992, Oliver2004} and atomic force microscopy \citep{Xu2022,zhou2022atomic}. However, this approach is inherently destructive and limited to small, localized regions, restricting its applicability in many contexts. In contrast, the indirect approach estimates the elasticity distribution from displacement fields induced by an applied surface force. This category includes Magnetic Resonance Elastography (MRE) \citep{muthupillai1995magnetic, manduca2021mr} and Ultrasound Elastography \citep{Lang1970, Buntin1990}, both of which are widely used in medical applications. 
It is important to note that displacement measurements obtained through these techniques are often contaminated by noise and constrained by resolution limitations inherent to the measurement processes, which can adversely affect the accuracy and reliability of elasticity estimation.

Inferring elasticity from displacement fields can be broadly categorized into three approaches. The first is a data-driven approach, which relies solely on historical data \citep{Nguyen2022, Ni2021, mouloodi2020prediction}. This approach requires sufficient high-quality training data, which is often unavailable, and its applicability is typically limited to the training domain, reducing reliability under unseen conditions. In contrast, physics-based models infer elasticity from governing physical laws, often expressed as partial differential equations (PDEs), and are commonly divided into direct and iterative methods. Direct methods convert elasticity equations into analytically tractable forms but are generally restricted to simple geometries or idealized assumptions \citep{Barbone2007, Babaniyi2017}. Iterative methods repeatedly run computationally intensive physics-based simulations, most commonly finite element models (FEM), to find elastic parameters that reproduce observed displacement fields \citep{Doyley2000, Smyl2019}. As a result, these methods can be computationally expensive and sensitive to initialization. Physics-Informed Neural Networks (PINNs) provide a hybrid approach by integrating data-driven learning with PDE-based physical constraints, combining neural-network expressiveness with governing laws for elasticity estimation \citep{Raissi2019, Raissi2020, Song2022}.

PINNs have been widely applied to forward problems, where displacement or strain fields are predicted from known elastic property distributions \citep{Yadav2022, Chen2023}. In contrast, the inverse problem of estimating elastic distributions from displacement observations is substantially more challenging. Early studies on inverse elasticity estimation typically assumed spatially uniform material properties, represented by constant elastic parameters \citep{Haghighat2021, Gao2022, Hamel2022}. 
More recent studies have focused on estimating both Young’s modulus and Poisson’s ratio in heterogeneous materials, but these approaches relied on restrictive assumptions, such as known internal or boundary stress distributions \citep{ Shukla2022, Ragoza2023, Kamali2023, Kamali2024} and a known mean Young’s modulus \citep{Chen2021, Chen2023}. Moreover, these studies generally assumed noise-free displacement observations, a condition rarely satisfied in practical applications. Recently, the Inverse Elasticity Physics-Informed Neural Network (IE-PINN) model successfully estimated elastic distributions from noisy displacement data by a deterministic approach \citep{Srikitrungruang2025}. While demonstrating promising performance, IE-PINN requires high-resolution displacement observations to construct accurate elasticity maps. In practice, however, measurements are often noisy and low-resolution, significantly limiting the accuracy of elasticity estimation.

PINNs, including IE-PINN, typically combine data-fidelity and physics-based loss components, requiring weights that balance observational data and governing physical laws. These weights are commonly specified empirically \citep{wang2022and}, which can make estimation sensitive to the chosen weights and poorly adaptive to dataset characteristics and learning dynamics across loss components. To address this limitation, dynamic weighting strategies have been proposed to assign adaptive weights either to individual loss components \citep{Deresse2025,Gao2025,Xiang2022, Anagnostopoulos2024} or to pointwise loss terms \citep{McClenny2023, Song2024, Anagnostopoulos2024}. A notable example is the Self-Adaptive PINN (SA-PINN), which uses a soft attention mechanism to learn loss weights jointly with neural network parameters within a minimax optimization framework \citep{McClenny2023}. By assigning larger weights to larger residuals, SA-PINN emphasizes terms that are harder to fit. However, this strategy can perform poorly under noisy conditions, as shown in the IE-PINN study \citep{Srikitrungruang2025}. Although Bayesian PINNs provide a principled framework for uncertainty quantification, they are often computationally prohibitive for large-scale inverse problems \citep{Yang2021}.

To address the challenge of inferring elastic properties from noisy, low-resolution displacement observations, we propose a novel framework, the Probabilistic Inverse Elasticity Physics-Informed Neural Network (PIE-PINN), for learning from two-dimensional displacement data, which is central to many practical applications, including ultrasound elastography \citep{Arda2011,Berko2013}. This probabilistic framework ensures consistency with both the observed data and the underlying physical laws, even in the presence of noise and limited spatial resolution. The main contributions of this work are summarized as follows:

\begin{itemize}
    \item \textbf{Robust probabilistic physics-informed neural network framework:}
    Existing PINNs often struggle with the ill-posedness of inverse elasticity problems, particularly when the material parameters are spatially heterogeneous, the data are noisy, and the observation resolution is low. We propose an effective probabilistic framework in which each residual term, including both data fidelity and physics consistency, is modeled using a Laplace distribution to improve robustness under heavy-tailed errors. We further establish a hierarchical statistical model by adapting the horseshoe+ prior to the Laplace residual formulation, thereby enhancing robustness to residual errors and stabilizing inference under severe ill-posedness exacerbated by low-resolution observations. This strategy mitigates underfitting induced by strong physics constraints and improves the accuracy of elasticity estimation.
	
    \item \textbf{B-spline-guided neural network:}
    When observations are of low resolution, interpolation errors between data points become large and are further amplified during numerical differentiation to compute the PDE residuals. To address this issue, we propose a hybrid functional representation that combines B-splines and neural networks, where the B-splines capture the global displacement trends, and the neural networks represent the local variations.

    \item \textbf{Adaptive weighting algorithm:}
    By jointly estimating the mean and scale parameters within a unified probabilistic model, the proposed framework naturally provides a principled adaptive weighting strategy that balances data fidelity and physics consistency. However, the ill-posedness of this inverse problem makes simultaneous estimation of all model parameters unstable, particularly when the observations are noisy and low-resolution. For stable estimation, we use an alternating maximum-likelihood procedure that iteratively updates the mean parameters and the scale parameters. In this formulation, updating the mean parameters corresponds to weighted residual minimization, as in conventional PINN training, whereas updating the scale parameters corresponds to updating the loss weights for the data and physics losses, as in adaptive loss-weighting schemes for PINNs. This approach is particularly effective for noisy, low-resolution data, where existing adaptive loss-weighting methods for PINNs often struggle. A comparative analysis with existing adaptive loss-weighting methods for PINNs is also provided.
\end{itemize}
The robustness and effectiveness of the proposed framework are demonstrated across multiple datasets with varying noise levels and observation resolutions.

The remainder of this article is organized as follows. Section~\ref{sec:background} introduces the background and fundamental concepts of PINNs for elasticity. Section~\ref{sec:proposedmodel} presents the proposed PIE-PINN framework. In Section ~\ref{sec:estimation}, the estimation procedure is described and compared with an existing adaptive weighting scheme. Section~\ref{sec:caseStudies} presents numerical case studies, including comparisons with benchmark methods, robustness evaluations under varying noise levels and observation resolutions, and ablation studies assessing the contributions of key components. Finally, Section~\ref{sec:conclusion} summarizes the main findings of the paper and outlines directions for future research.

%% file: Contents/2__background.tex
\section{Background} \label{sec:background}

This section introduces the theoretical and methodological foundations of the proposed framework. It begins with the underlying physics model, namely, the linear elasticity formulation, in Section~\ref{ssec:lemodel}, which establishes the fundamental relationship between displacement fields and elastic material properties. Building upon this foundation, Section~\ref{ssec:pinn} presents the PINN formulation for elasticity, describing how governing physics equations and observational data are integrated within a unified learning framework. The discussion then focuses on the inverse elasticity problem, in which spatially heterogeneous elastic parameters are inferred from displacement measurements, and reviews the IE-PINN formulation, which serves as the baseline for the proposed method.

\input{Contents/2_1_LEmodel}

\input{Contents/2_2_PINN}

%% file: Contents/2_1_LEmodel.tex
\subsection{Physics model: Linear elasticity }\label{ssec:lemodel}


\begin{figure}[!t]
    \centering
    \includegraphics[width=\textwidth]{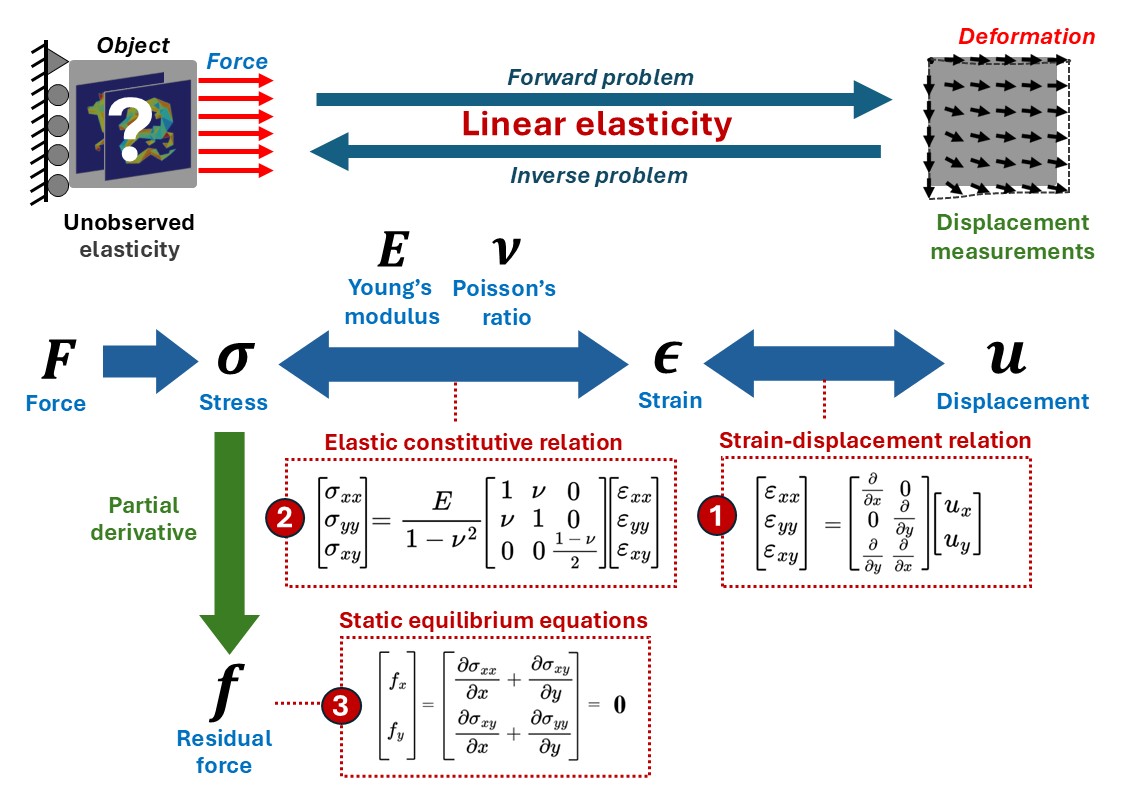}
    \caption{Governing relationships of the linear elasticity model, including strain-displacement, constitutive, and equilibrium equations.}
\label{fig:LE_model}
\end{figure}

The linear elasticity model is a physics-based mathematical framework that describes how solid materials deform and develop internal stresses under applied loads. In this work, we consider a two-dimensional inverse linear elasticity problem in which unknown elastic properties are inferred from observed displacement data. As illustrated in Figure~\ref{fig:LE_model}, external forces produce deformation that is represented by a displacement vector field. 
We assume a spatially heterogeneous isotropic material, for which the mechanical properties may vary across locations while remaining identical in all directions at each point. A plane stress condition is adopted, which is appropriate for thin structures subjected to in-plane loading.
The formulation consists of three key components: the strain-displacement relation, the elastic constitutive relation, and the static equilibrium equations. These relations are outlined below.

First, the strain-displacement relation characterizes material deformation within the framework of continuum mechanics. The motion of material points is represented by the displacement vector $\bm{u} = [u_x, u_y]^\top$. The relative spatial change in displacement is quantified by the strain vector $\bm{\varepsilon} = [\varepsilon_{xx}, \varepsilon_{yy}, \varepsilon_{xy}]^\top$. Under the small-strain assumption, the relation between displacement and strain is expressed through a linear first-order differential operator, $\mathcal{E}$, which maps the displacement field to the strain field.

\begin{equation}
    \mathcal{E}(\bm{u}) = 
    \begin{bmatrix}
        \varepsilon_{xx} \\
        \varepsilon_{yy} \\
        \varepsilon_{xy} \\
    \end{bmatrix}
    =
    \begin{bmatrix}
        \frac{\partial }{\partial x} & 0\\
        0 & \frac{\partial }{\partial y} \\
        \frac{\partial }{\partial y}  & \frac{\partial }{\partial x} \\
    \end{bmatrix}
    \begin{bmatrix}
        u_{x} \\
        u_{y} \\
    \end{bmatrix}
    \label{eq:displacementStrainRelation}
\end{equation}
where $\varepsilon_{xx}$ and $\varepsilon_{yy}$ denote the normal strain components, $\varepsilon_{xy}$ denotes the shear strain, and $\partial/\partial x$ and $\partial/\partial y$ denote partial derivatives with respect to $x$ and $y$, respectively. 

Secondly, the elastic constitutive relation describes the relationship between stress and strain in a material through Young's modulus ($E$) and Poisson's ratio ($\nu$). The stress vector, denoted by $\bm{\sigma}$, quantifies the internal force per unit area within the material. Both Young's modulus and Poisson's ratio are commonly used to characterize the elastic behavior of the material and are represented as a material parameter vector $\bm{E} = [E,\nu ]^\top$. 
Specifically, Young's modulus indicates a material's resistance to elastic deformation along the direction of the applied force, while Poisson's ratio describes how deformation in one direction relates to deformation in the perpendicular direction.
This relationship is expressed as:

\begin{equation}
    \bm{\sigma} = 
    \begin{bmatrix}
        \sigma_{xx} \\
        \sigma_{yy} \\
        \sigma_{xy} \\
    \end{bmatrix}
    =
    \frac{E}{1-\nu^2}
    \begin{bmatrix}
        1 & \nu & 0\\
        \nu & 1 & 0\\
        0 & 0 & \frac{1-\nu }{2} \\
    \end{bmatrix}
    \begin{bmatrix}
        \varepsilon_{xx} \\
        \varepsilon_{yy} \\
        \varepsilon_{xy} \\
    \end{bmatrix}
    \label{eq:elasticCon}
\end{equation}
where $\sigma_{xx}$ and $\sigma_{yy}$ denote the normal stress components, and $\sigma_{xy}$ denotes the shear stress component.

Third, the static equilibrium equations enforce pointwise force balance within the material domain. Under static conditions, the residual force vector $\bm{f} = [f_x,f_y]^\top$, defined as the divergence of the stress tensor, must vanish at every point in the domain:
\begin{equation}
    \bm{f} = 
    \begin{bmatrix}
        f_{x} \\
        f_{y} \\
    \end{bmatrix}
    =
    \begin{bmatrix}
        \dfrac{\partial \sigma_{xx}}{\partial x} + \dfrac{\partial \sigma_{xy}}{\partial y} \\
        \dfrac{\partial \sigma_{xy}}{\partial x} + \dfrac{\partial \sigma_{yy}}{\partial y} \\
    \end{bmatrix}=
    \bm{0}
    \label{eq:equilibrium}
\end{equation}

The three relations above jointly define the governing equations of linear elasticity. These equations may be written compactly in a residual operator as
\begin{align}
    \bm{f} = \mathcal{F}_u(\bm{u}, \bm{E}) = \mathcal{F}_\varepsilon(\bm{\varepsilon}, \bm{E})
    \label{eq:residualOperator}
\end{align}
where $\mathcal{F}_u$ and $\mathcal{F}_\varepsilon$ denote the equilibrium residual operators expressed in terms of the displacement field $\bm{u}$ and the strain field $\bm{\varepsilon}$, respectively. In both cases, the operator maps the field variable and material parameter vector $\bm{E}$ to the residual force vector $\bm{f}$.

%% file: Contents/2_2_PINN.tex
\subsection{Physics-Informed Neural Network for Elasticity}\label{ssec:pinn}

PINNs are a class of deep learning frameworks that integrate data-driven learning with known physical laws, typically expressed as PDEs \citep{Raissi2019}. In the context of elasticity, the governing equations introduced in Section~\ref{ssec:lemodel} are incorporated into the learning framework as physical constraints by enforcing the equilibrium condition (Equation~\eqref{eq:equilibrium}) through the residual operator defined in Equation~\eqref{eq:residualOperator}.

\begin{figure}[!t]
    \centering
    \includegraphics[width=\textwidth]{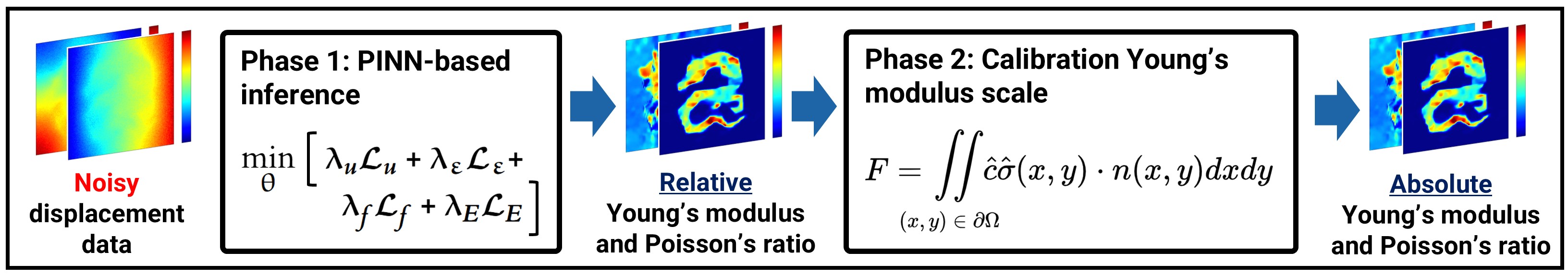}
    \justifying
    \caption{Conceptual flow of the two-phase strategy used in the IE-PINN framework for estimating spatially heterogeneous elastic properties from noisy displacement data.}
\label{fig:PIEPINN_Framework_A}
\end{figure}

The inverse elasticity problem is inherently ill-posed and often suffers from instability and non-uniqueness in its solution \citep{Kabanikhin2008}. A recent study demonstrated promising results in addressing these challenges by learning elastic properties from noisy displacement fields using the IE-PINN framework \citep{Srikitrungruang2025}. The framework consists of two phases: (i) a deterministic PINN-based inference phase for estimating the Poisson's ratio and relative Young's modulus, and (ii) a calibration phase for recovering the true-scale Young's modulus. 
An overview of the framework is shown in Figure~\ref{fig:PIEPINN_Framework_A}. 
Because the true-scale Young's modulus can be recovered from the estimated relative Young's modulus using boundary conditions, this work focuses on Phase 1, namely the effective estimation of relative Young's modulus and Poisson's ratio.

To enable the estimation of spatially heterogeneous elastic parameters from noisy observations, IE-PINN departs from conventional PINNs in three main ways: it (1) introduces additional neural networks for strain and material parameters, with a modified composite loss, (2) uses $L_1$-norm minimization instead of $L_2$-norm minimization, and (3) evaluates spatial derivatives by finite differences rather than automatic differentiation.

Specifically, IE-PINN uses separate neural networks to approximate the displacement field $\bm{u}$, strain field $\bm{\varepsilon}$, Young's modulus $E$, and Poisson's ratio $\nu$, denoted by $\hat{\bm{u}}$, $\hat{\bm{\varepsilon}}$, $\hat{E}$, and $\hat{\nu}$, respectively.
Accordingly, the model minimizes a composite loss function consisting of four terms: a data-fidelity term, a strain-discrepancy term, a physics-consistency term, and a mean-modulus constraint term for estimating the relative Young's modulus, as described below.

The first loss term enforces data fidelity by minimizing the discrepancy between the observed displacement measurements $\bm{u}_i^*=[u_{i,x}^*, u_{i,y}^*]^\top$ and the predicted values $\hat{\bm{u}}(\X_i)=[\hat{u}_{i,x}, \hat{u}_{i,y}]^\top$, where $\X_i$ denotes the coordinate of the $i$th observation point. The corresponding data loss is given by
\begin{align}
    \LOSS_{u} = \sum_{j\in \{x,y\}} \sum_{i=1}^{N_u}\left|{u}_{i,j}^* - \hat{u}_{i,j}\right|
    \label{eq:iepinn_disp}
\end{align}
where $N_u$ denotes the number of displacement observation points.


The second loss term quantifies the discrepancy between the strain field predicted by the strain network, $\hat{\bm{\varepsilon}}(\X_i) = [\hat{\varepsilon}_{i,xx}, \hat{\varepsilon}_{i,yy}, \hat{\varepsilon}_{i,xy}]^\top$, and the strain field obtained by applying the strain operator to the predicted displacement field, $\mathcal{E}(\hat{\bm{u}}(\X_i))=  [\varepsilon^u_{i,xx}, \varepsilon^u_{i,yy}, \varepsilon^u_{i,xy}]^\top$, as defined in Equation~\eqref{eq:displacementStrainRelation}. The strain discrepancy loss is given by
\begin{align}
    \LOSS_{\varepsilon} = \sum_{k\in \{xx,yy,xy\}} \sum_{i=1}^{N_\varepsilon}\left| \hat{{\varepsilon}}_{i,k} - {{\varepsilon}}^u_{i,k} \right| 
    \label{eq:iepinn_strain}
\end{align}
where $N_\varepsilon$ denotes the number of evaluation points for the strain discrepancy term.

The third loss enforces physics consistency by minimizing the residual force associated with the governing PDE, given by
$\mathcal{F}_\varepsilon(\hat{\bm{\varepsilon}}(\X_i), \hat{\bm{E}}(\X_i)) = [f_{i,x}, f_{i,y}]^\top$.
This residual measures the degree of violation of the equilibrium equations and is theoretically zero when the true elastic parameters are used together with an exact solution. The corresponding physics loss is defined as
\begin{align}
    \LOSS_{f} = \sum_{j\in \{x,y\}}\sum_{i=1}^{N_f} \left|\frac{f_{i,j}}{\tilde{E}(\X_i)}\right|
    \label{eq:iepinn_pde}
\end{align}
where $N_f$ denotes the number of PDE evaluation points, and $\tilde{E}(\X_i)$ is a local aggregation of predicted Young's modulus values at $(\X_i)$, computed as the average over a $3\times3$ neighborhood surrounding each evaluation point and used for PDE-residual normalization to improve numerical stability.

The fourth loss term constrains the mean predicted elastic modulus $\bar{E}$ toward a prescribed reference value $E_c$ to prevent the estimated modulus field from collapsing toward zero because of the ill-posedness of the inverse problem. This mean-modulus constraint is defined as
\begin{align}
    \LOSS_{E} = \left| \bar{E} - E_c \right|
\end{align}

The spatial derivatives appearing in the strain-displacement relation (Equation~\eqref{eq:displacementStrainRelation}) and the static equilibrium equations (Equation~\eqref{eq:equilibrium}) are evaluated numerically using finite differences implemented through convolution operations, as detailed in Supplementary Note \ref{sup:FD}. 

All loss components are constructed using $L_1$-norm minimization. Compared with $L_2$-norm minimization, which is more sensitive to large residuals, the $L_1$ norm provides improved robustness in the presence of gross measurement errors \citep{Bektas2010}. This property is particularly advantageous when the observational data are noisy, as in the IE-PINN setting.

Therefore, a multi-objective loss is minimized through a weighted sum of the individual loss terms,
\begin{equation}
     \mathcal{L}_{\text{total}}  =
     {\lambda}_{u} \LOSS_{u}
     + {\lambda}_{\varepsilon} \LOSS_{\varepsilon}
     + {\lambda}_{f} \LOSS_{f}
     + {\lambda}_{E} \LOSS_{E}
     \label{eq:convPINNloss}
\end{equation}
where $\lambda_{u}, \lambda_{\varepsilon}, \lambda_{f},$ and $\lambda_{E}$ denote predefined weighting coefficients for the respective loss components.  

This conventional PINN formulation has two practical limitations for inverse elasticity with noisy, low-resolution displacement measurements. First, these weights are often selected empirically, and inadequate loss balancing can cause training instability, overfitting or underfitting, and even failed recovery of elastic properties. Second, displacement observations are typically treated through deterministic residual losses, without explicitly modeling observational noise, which limits robustness in such settings.

%% file: Contents/3__model.tex
\section{Probabilistic Inverse Elasticity Physics-Informed Neural Network} \label{sec:proposedmodel}

\input{Contents/3_1_ProposedModel}

%% file: Contents/3_1_ProposedModel.tex
\begin{figure}[!t]
    \centering
    \includegraphics[width=\textwidth]{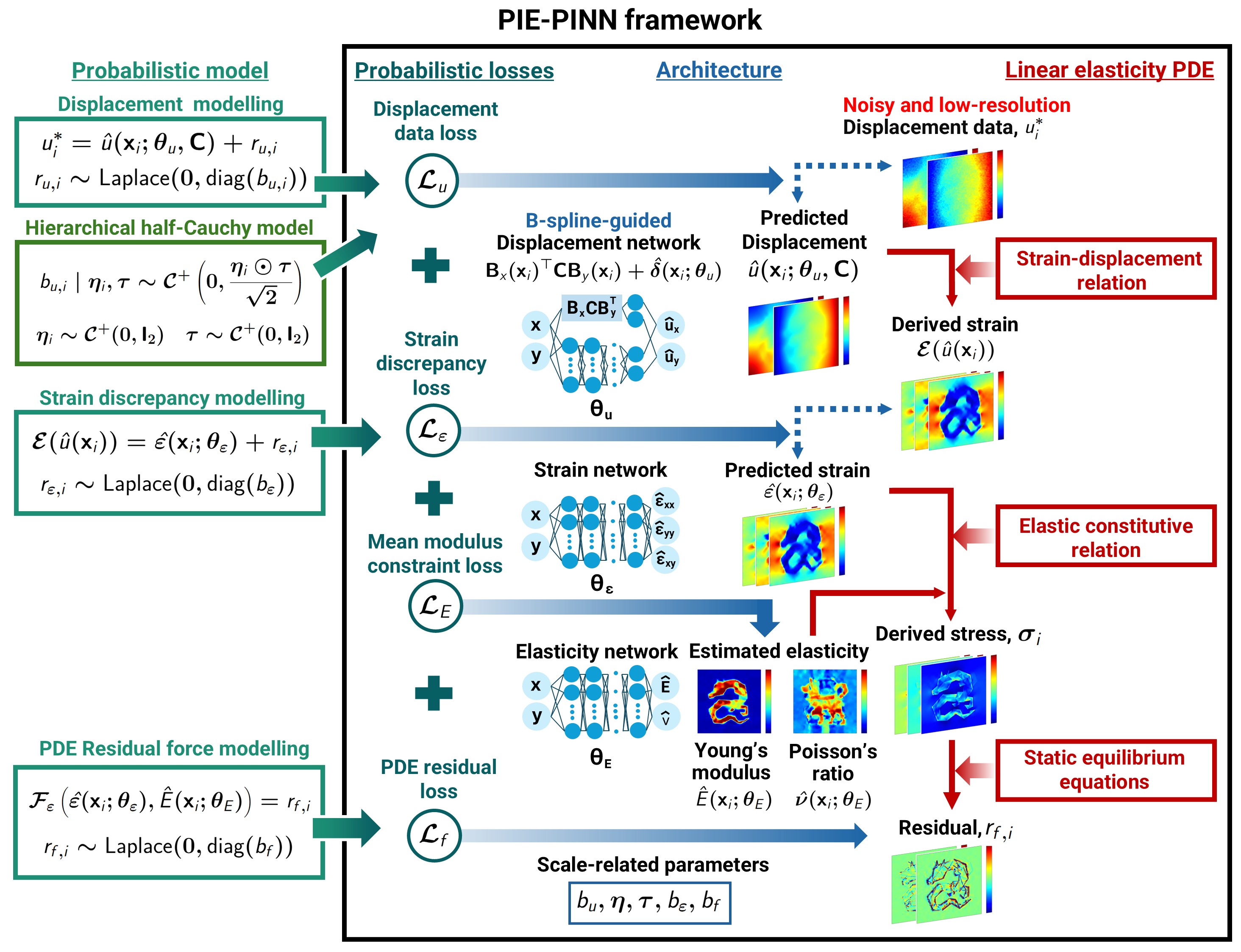}
    \justifying
    \caption{Framework Probabilistic Physics-Informed Neural Networks (PIE-PINN) for heterogeneous elasticity estimation from noisy Low-Resolution and displacement data.}
\label{fig:PIEPINN_framework_B}
\end{figure}

We propose a novel framework called the Probabilistic Inverse Elasticity Physics-Informed Neural Network (PIE-PINN) for estimating elastic properties, including Young's modulus and Poisson’s ratio, from noisy and low-resolution displacement observations, as illustrated in Figure~\ref{fig:PIEPINN_framework_B}. The proposed framework explicitly models the noise in the displacement measurements and its propagation through the governing equations, providing a rigorous interpretation of the underlying noise and error sources while naturally enabling adaptive weighting of the loss terms. 
In particular, we assign a tailored probabilistic model to the residual associated with each loss term that captures its specific objective, including enforcing displacement data fidelity, minimizing strain discrepancies, and ensuring physical consistency. 

Following the notation introduced in the previous sections, the underlying unobserved pointwise elastic properties are represented by $\bm{E} = [E, \nu]^\top$ and are inferred from the observed displacement field $(\bm{u}^*)$,  where 
$\bm{u}_i^* \in \mathbb{R}^{2}$ 
denotes the displacement vector measured at the uniformly spaced spatial location $\X_i=[x_i,y_i]^\top$, and $i$ denotes the observation index. The corresponding strain and stress fields are denoted by $\bm{\varepsilon}$ and $\bm{\sigma}$, respectively. 
These unknown fields are modeled by function approximators denoted by 
$\hat{\bm{E}}(\X_i; \theta_{\bm{E}}) \in \mathbb{R}^{2}$, $\hat{\bm{u}}(\X_i; \theta_{\bm{u}},\splParaU)\in \mathbb{R}^{2}$, and $\hat{\bm{\varepsilon}}(\X_i; \theta_{\bm{\varepsilon}}) \in \mathbb{R}^{3}$. 
In addition, the strain field derived from the predicted displacement field is given by 
$\mathcal{E}(\hat{\bm{u}}(\X_i)) \in \mathbb{R}^{3}$. 
The stress field $\hat{\bm{\sigma}}$ is subsequently computed through the constitutive relation and used to evaluate the static equilibrium residual as \eqref{eq:equilibrium}.
The overall hierarchical physics-informed probabilistic model is shown below.



\newpage
\noindent\textbf{Displacement modeling}
\vspace{-0.5em}
\begin{alignat}{2}
&\textit{Observation:}\quad
& \bm{u}_i^*
&= \hat{\bm{u}}(\X_{i};\nnParaU,\splParaU) + \bm{r}_{u,i},
\qquad
\bm{r}_{u,i} \sim \Lap(\bm{0},\diag({\bm{b}_{u,i}}))
\label{eq:model_disp}
\\
&\textit{Mean:}\quad
& \hat{\bm{u}}(\X_i;\nnParaU,\splParaU)
&= \mathbf{B}_x(\X_i)^\top \splParaU \mathbf{B}_y(\X_i)
+ \hat{\delta}(\X_i;\nnParaU)
\label{eq:pred_disp}
\\
&\textit{Error scale:}\quad
& \bm{b}_{u,i} \mid \uEta_i, \uTau
&\sim \CAUHCHY\left(\bm{0}, \frac{\uEta_i  \odot \uTau}{\sqrt{2}}\right),
\qquad
\uEta_i \sim \CAUHCHY(\bm{0},\mathbf{I}_2),
\quad
\uTau \sim \CAUHCHY(\bm{0},\mathbf{I}_2)
\label{eq:hs_disp}
\end{alignat}

\noindent
\textbf{Strain discrepancy modeling}
\vspace{-0.5em}
\begin{align}
    \hspace{1em}\mathcal{E}(\hat{\bm{u}}(\X_i))  &= \bm{\hat{\varepsilon}} (\X_i;\nnParae)  + \bm{r}_{\varepsilon,i}
    , \qquad
    \bm{r}_{\varepsilon,i}  \sim \Lap(\bm{0},  \diag(\bSTRAIN))  \label{eq:model_strain}
\end{align}
\noindent
\textbf{PDE Residual force modeling}
\vspace{-0.5em}
\begin{align}
        \mathcal{F}_\varepsilon \left(\hat{\bm{\varepsilon}}(\X_{i};\nnParae), \hat{\bm{E}}(\X_i;\nnParaE) \right)  &=  \bm{r}_{f,i} , \qquad
    \bm{r}_{f,i}\sim  \Lap(\bm{0},  \diag(\bPDE))  \label{eq:model_pde} 
\end{align}
where $\bm{r}_{u,i} \in \mathbb{R}^{2}$, $\bm{r}_{\varepsilon,i} \in \mathbb{R}^{3}$, $\bm{r}_{f,i} \in \mathbb{R}^{2}$ 
denote the pointwise residual vectors associated with displacement fitting, strain discrepancy, and PDE residual force, respectively. The corresponding Laplace scale parameters are $\bm{b}_{u,i} \in \mathbb{R}^{2}$
, which are modeled as pointwise scale parameters, whereas  $\bm{b}_{\varepsilon} \in \mathbb{R}^{3}$, and $\bm{b}_{f} \in \mathbb{R}^{2}$
are treated as global scale parameters.
Here, $\odot$ denotes componentwise multiplication, and $\CAUHCHY(\bm{0},\cdot)$ denotes a componentwise two-dimensional half-Cauchy distribution. That is, a random vector following this distribution has independent half-Cauchy components, whose scale parameters are given componentwise by the second argument. In \eqref{eq:hs_disp}, $\uEta_i$ and $\uTau$ denote the local and global shrinkage variables, respectively, as part of the horseshoe+ hierarchy.


The proposed probabilistic model consists of three components: displacement observation modeling, strain-discrepancy modeling, and residual-force modeling. 
The errors, or residuals, in all three components are modeled using Laplace distributions. 
This choice is important for inverse problems, where ill-posedness, observational noise, and imperfect enforcement of physics constraints can generate large errors.
Compared with a Gaussian model, the heavier tails of the Laplace distribution reduce sensitivity to large errors and improve recovery of the underlying signal in the presence of noise and model mismatch. This formulation is also consistent with IE-PINN \citep{Srikitrungruang2025}, which used $L_1$-norm losses for robust estimation.

The displacement observation model is given in \eqref{eq:model_disp}--\eqref{eq:hs_disp}. 
Robust mean estimation from noisy, low-resolution displacement observations is critical but challenging. 
Strong physics regularization can induce underfitting, causing part of the underlying displacement signal to be absorbed into the residual term, and this issue is exacerbated under low-resolution observations.
To improve robustness, we incorporate two additional designs: a B-spline-guided displacement network and a horseshoe+ hierarchical model for the Laplace scale parameters.
In practice, the displacement field 
${\bm{u}}$ 
is often modeled using a deep neural network as the sole function approximator \citep{Haghighat2021, Sahin2024}. 
Although deep neural networks have high representational capacity and can approximate complex functions \citep{Zhang2016}, they are prone to overfitting when trained with limited or low-resolution observations \citep{Jaehong2017}. 
To mitigate this issue, we define the predicted displacement field by combining a B-spline interpolation function, $\mathbf{B}_x(\X)^\top \splParaU \mathbf{B}_y(\X)$, with a neural network correction term, $\delta_{NN}(\X;\nnParaU)$, as in Equation~\eqref{eq:pred_disp}. 
Here, $\mathbf{B}_x(\X)$ and $\mathbf{B}_y(\X)$ denote B-spline basis vectors along the $x$ and $y$ directions, $\splParaU$ denotes the spline coefficient matrix, and $\nnParaU$ denotes the neural network parameters.
The B-spline component provides a smooth global approximation of the displacement field, while the neural network correction captures local variations that cannot be represented by the spline basis alone. This hybrid architecture reduces sensitivity to measurement noise and supports the recovery of high-resolution displacement fields from noisy, low-resolution observations. 

For effective fitting of the observed displacement data, a sparsity-enhancing horseshoe+ hierarchical model is employed for the Laplace scale parameters $\bm{b}_{u,i}$ in the displacement error model, as shown in Equation~\eqref{eq:hs_disp}.
The heavy tails of the horseshoe+ distribution model the large residuals induced by noise, model mismatch, or the trade-off between data fitting and physics consistency, while its sharp concentration near zero shrinks small residuals toward zero. 
This combination helps distinguish severe fitting errors from informative displacement signals, enabling robust recovery of the hidden mean field under substantial noise and low-resolution observations.

The strain-discrepancy model in \eqref{eq:model_strain} links the strain network output $\hat{\bm{\varepsilon}}(\X;\nnParae)$ with the strain field computed from the displacement through $\mathcal{E}(\hat{\bm{u}}(\X))$, following the IE-PINN formulation shown to stabilize inverse estimation under the $L_1$ loss \citep{Srikitrungruang2025}. The governing physics is incorporated in \eqref{eq:model_pde} through the residual-force model.

The proposed hierarchical model provides a joint probabilistic representation of the displacement observations, strain discrepancy, and equilibrium residuals. Based on this formulation, the PIE-PINN framework provides a systematic approach to training PINNs while explicitly accounting for fitting errors and noise in both observations and latent physical quantities. However, estimating all parameters simultaneously can be unstable in noisy and low-resolution inverse problems. Therefore, the next section presents an alternating estimation procedure for joint likelihood maximization.

%% file: Contents/4_ParamEst.tex
\section{Parameter Estimation} \label{sec:estimation}

The proposed model may be viewed as having a hierarchical Bayesian structure, especially through the horseshoe+ hierarchical model for the Laplace scale parameters. With additional prior specifications for the remaining mean-function and scale parameters, one may naturally consider full Bayesian inference to estimate posterior distributions of the denoised functional variables and model parameters.
%
However, the inverse elasticity problem considered here is inherently ill-posed, and even the simpler deterministic formulation is numerically unstable and computationally demanding when multiple coupled neural-network components are involved \citep{Barabasz2014}. Full Bayesian inference over the coupled mean-function and scale parameters would further increase the computational burden and could be difficult to stabilize for this problem. Therefore, we do not pursue full Bayesian inference in this work and focus on estimating the denoised mean functions of displacement, strain, stress, and elastic properties through functional approximations, by minimizing the negative log-joint likelihood under the specified hierarchical probabilistic model, as detailed in Section~\ref{ssec:NLL}. We further provide an analytical comparison between the proposed formulation and existing adaptive weighting approaches in Section~\ref{ssec:lossAnalysis}. Full Bayesian uncertainty quantification is left for future research.

\subsection{Negative Log-likelihood Minimization} \label{ssec:NLL}

Based on the hierarchical probabilistic model introduced in Section~\ref{sec:proposedmodel}, the inverse elasticity problem is formulated as the minimization of the full joint negative log-likelihood with respect to both the model parameters (i.e., the neural-network weights and spline coefficients) and scale-related parameters.
We denote the function-approximator parameters as $\Theta = \{\nnParaU, \splParaU, \nnParae, \nnParaE\}$, and the scale-related parameters as $\Phi = \{\bDISP, \uEta, \uTau, \bSTRAIN, \bPDE\}$, where $\bDISP =  \{\bDISPi\}^{\nDisp}_{i=1}$ and $\uEta = \{\uEta_i\}^{\nDisp}_{i=1}$ denote the collection of pointwise displacement scale parameters and local shrinkage parameters from horseshoe+ distribution, respectively.

The resulting optimization problem is given by
\begin{align}
\min_{\Theta,\Phi}\;\mathcal{L}_{\text{total}}(\Theta,\Phi)
= \min_{\Theta,\Phi}\;\left[
\LOSS_u + \LOSS_{\varepsilon} + \LOSS_f + \mathcal{L}_E
\right],
\label{eq:pie_total_loss}
\end{align}
where $\LOSS_u$, $\LOSS_{\varepsilon}$, and $\LOSS_f$ are derived from the probabilistic models in \eqref{eq:model_disp}-\eqref{eq:model_pde}, and $\mathcal{L}_E$ is the deterministic mean-modulus constraint inherited from IE-PINN. 

We next describe the four loss terms.
The negative log-likelihood for the displacement observation model $\LOSS_{u}$ is
\begin{align}
\LOSS_{u}
&=
\sum_{j\in \{x,y\}}
\left[
\sum_{i=1}^{\nDisp}
\log(2\sbDISP)
+
\sum_{i=1}^{\nDisp}
\frac{
\left|
\hat{{u}}_{i,j}
-{{u}}_{i,j}^*
\right|
}{\sbDISP}
\right]
+
l_{\mathcal{H}+}(\bm{b}_{u}, \uTau, \uEta),
\label{eq:pie_disp_loss}
\end{align}
where ${{u}}_{i,j}^*$ and $\hat{{u}}_{i,j}$ denote the observed and predicted displacement values in the $j$ direction at the $i$th observation point, respectively, $\bm{b}_{u,i,j}$ denotes the corresponding pointwise Laplace scale parameter, and $j\in \{x,y\}$ identifies the displacement component. The term $l_{\mathcal{H}+}(\bm{b}_u, \uTau, \uEta)$ is the negative log-likelihood associated with the horseshoe+ hierarchical model, which employs a hierarchical half-Cauchy distribution for the latent displacement error scales. Specifically, $l_{\mathcal{H}+}(\bm{b}_u, \uTau, \uEta)$ is
\begin{align}
\sum_{j\in \{x,y\}}
\left[
\sum_{i=1}^{\nDisp}
\left[
-\log
\left\{
\frac{2\sqrt{2}}{\pi \suTau \suEta}
\left(
1+\frac{2\sbDISP^2}{\suTauSqn \suEtaSqn}
\right)^{-1}
\right\}
-
\log
\left\{
\frac{2}{\pi(1+\suEtaSqn)}
\right\}
\right]
-
\log
\left\{
\frac{2}{\pi(1+\suTauSqn)}
\right\} \right],
\label{eq:pie_hs_loss}
\end{align}
where $\suEta$ and $\suTau$ denote the local and global shrinkage parameters, respectively. While $\suEta$ varies across observation points, $\suTau$ is shared globally within each displacement component.

The first bracketed term in \eqref{eq:pie_disp_loss} is the negative log-likelihood of the displacement observations under the Laplace residual model. The term $l_{\mathcal{H}+}(\bm{b}_u, \uTau, \uEta)$ induces global-local shrinkage on the pointwise Laplace error scales, allowing sparse large residuals while shrinking most scales toward the global scale. If separate global scales are used for different displacement components, the final global-scale term in \eqref{eq:pie_hs_loss} should be modified accordingly.

The negative log-likelihood of the strain-discrepancy model, $\LOSS_{\varepsilon}$, is 
\begin{align}
\LOSS_{\varepsilon}
=
\sum_{k\in \{xx,yy,xy\}}
\left[
\nStrain\log(2\sbSTRAIN)
+
\frac{1}{\sbSTRAIN}
\sum_{i=1}^{\nStrain}
\left|
\hat{\varepsilon}_{i,k}
-
{\varepsilon}_{i,k}^u
\right|
\right]
,
\label{eq:pie_strain_loss}
\end{align}
where ${\varepsilon}_{i,k}^u$ and $\hat{\varepsilon}_{i,k}$ denote the strain values derived from the predicted displacement field and predicted by the strain network, respectively, $\sbSTRAIN$ denotes the global Laplace scale parameter, and $k\in \{xx,yy,xy\}$  denotes the strain component.
Thus, $1/\sbSTRAIN$ acts as a learned inverse-scale weight for the strain-consistency residual, rather than as a manually selected tuning parameter.

The negative log-likelihood of the equilibrium residual, which is obtained from \eqref{eq:model_pde}, is
\begin{align}
\LOSS_f
=
\sum_{j\in \{x,y\}}
\left[
\nPDE\log(2\sbPDE)
+
\frac{1}{\sbPDE}
\sum_{i=1}^{\nPDE}
\left|
\frac{
r_{f,i,j}
}{\tilde{E}(\mathbf{x}_i)}
\right|
\right],
\label{eq:pie_force_loss}
\end{align}
where $r_{f,i,j}$ denotes the residual force at the $i$th point in the $j$th spatial direction ($j\in \{x,y\}$), and $\sbPDE$ denotes the global Laplace scale parameter associated with the residual force term. The equilibrium residual is also normalized by the predicted Young's modulus $\tilde{E}(\mathbf{x}_i)$ to facilitate elasticity estimation.

Whereas the mean-modulus constraint is not treated as a random residual. It is retained as a deterministic constraint term,
\begin{align}
\mathcal{L}_E
=
\lambda_E
\left|
\bar{E}-E_c
\right|,
\label{eq:pie_mean_modulus_loss}
\end{align}
because it constrains a single summary quantity rather than an observation-level residual distribution. The weight term, $\lambda_E$, is fixed throughout training. The primary role of this loss term is to constrain the average predicted Young's modulus to a prescribed value, thereby alleviating the ill-posedness of the inverse problem and preventing the predicted modulus field from collapsing to the trivial zero solution. Furthermore, previous IE-PINN studies have demonstrated that the training performance is relatively insensitive to the choice of this weight.


It is natural to consider minimizing \eqref{eq:pie_total_loss} with respect to both $\Theta$ and $\Phi$ together, which would address the long-standing challenge in PINNs that their performance depends heavily on prescribed weights among loss terms. However, due to the ill-posedness of the inverse problem, direct simultaneous minimization over both $\Theta$ and $\Phi$ can be numerically unstable and may drive the neural-network components toward divergent solutions. This instability is already observed in the simpler deterministic inverse PINN formulation \citep{Wang2021, Wang2025}, and it becomes more pronounced when neural-network components and scale parameters are updated simultaneously.

We therefore optimize the objective using an alternating optimization scheme inspired by the EM algorithm \citep{Moon1996}. The first step, referred to as the E-step, estimates the function-approximator parameters $\Theta$ given the observed data and the current scale parameters $\Phi$. In this step, the parameters for the mean functions, including neural-network parameters and the B-spline coefficient matrix, are updated to minimize the loss function, as in conventional PINN training, while the scale parameters are fixed.
The subsequent step, referred to as the M-step, estimates the scale parameters $\Phi$ while holding the mean-function parameters $\Theta$ fixed. In each M-step, the scale parameters are updated to reflect the magnitudes of the residuals associated with each loss component, and the corresponding weights are adjusted accordingly. 
The mathematical description of the proposed algorithm is summarized in Algorithm \ref{alg:em_piepinn}.

The full joint negative log-likelihood function in \eqref{eq:pie_total_loss} can be decomposed into $\mathcal{L}_{\Theta}(\Theta,\Phi)$, which includes the terms involving the mean-function parameters $\Theta$ along with $\Phi$, and $C(\Phi)$, which collects the terms that depend only on the scale parameters $\Phi$:
\begin{align}
\mathcal{L}_{\text{total}}(\Theta,\Phi)
=
\mathcal{L}_{\Theta}(\Theta,\Phi)
+
C(\Phi).
\end{align}
For fixed $\Phi$, the mean-function objective $\mathcal{L}_{\Theta}(\Theta,\Phi)$ is
\begin{align}
\sum_{j\in \{x,y\}}
\left[
\sum_{i=1}^{\nDisp}
\frac{
\left|
\hat{{u}}_{ij}
-
{{u}}^*_{ij}
\right|
}{\sbDISP}
+
\frac{1}{\sbPDE}
\sum_{i=1}^{\nPDE}
\left|
\frac{
r_{f,i,j}
}{\tilde{E}(\mathbf{x}_i)}
\right|
\right]
+
\sum_{k\in \{xx,yy,xy\} }
\left[
\frac{1}{\sbSTRAIN}
\sum_{i=1}^{\nStrain}
\left|
\hat{\varepsilon}_{i,k}
-
{\varepsilon}_{i,k}^u
\right|
\right]
+
\lambda_E
\left|
\bar{E}-E_c
\right|.
\label{eq:E-Loss}
\end{align}

Since the E-step optimizes only over $\Theta$, it minimizes $\mathcal{L}_{\Theta}(\Theta,\Phi)$ in \eqref{eq:E-Loss} with respect to $\Theta$ given fixed $\Phi$. 
%
Notice that \eqref{eq:E-Loss} has a form similar to the deterministic PINN loss in \eqref{eq:convPINNloss}, where the inverse scales $1/b_\cdot$ correspond to the loss weights $\lambda_\cdot$, and pointwise inverse scales are used for the displacement observations.
Therefore, the E-step, namely, minimizing $\mathcal{L}_{\text{total}}$ with respect to $\Theta$ given fixed $\Phi$, is equivalent to the deterministic PINN approach with the corresponding fixed loss weights. 
In conventional PINN training, these weights are prescribed manually. In PIE-PINN, they are estimated during training from the probabilistic residual and scale models in the M-step.



In the M-step, the scale-only component $C(\Phi)$ is also added to the objective function:
\begin{align}
C(\Phi)
=
\sum_{j\in \{x,y\} }
\left[
\sum_{i=1}^{\nDisp}
\log(2\sbDISP)
+
\nPDE\log(2\sbPDE)
\right]
+
\sum_{ k\in \{xx,yy,xy\} }
\left[
\nStrain\log(2\sbSTRAIN)
\right]
+
l_{\mathcal{H}+}(\bm{b}_u, \uTau, \uEta).
\label{eq:C_phi}
\end{align}
Here, $C(\Phi)$ consists of the likelihood-normalizing terms and the hierarchical half-Cauchy scale-model contribution. These terms balance the scale parameters. In minimizing $\mathcal{L}_{\Theta}(\Theta,\Phi)$, increasing the scale parameter $b$ reduces the corresponding weighted residual term. On the other hand, the likelihood-normalizing term $N\log(2b)$ prevents the scale parameter from increasing without bound. Thus, the likelihood structure yields data-adaptive weights without allowing the trivial collapse or divergence of the weights.

Therefore, the proposed EM-based algorithm provides a rigorous likelihood-based adaptive loss-weighting scheme for training PIE-PINN through two fundamental iterative steps. 
In the proposed training scheme, the model is first pretrained using an initial E-step for 100 epochs, followed by an M-step for 100 epochs. The training then alternates between E-steps of 25 epochs and M-steps of 100 epochs until the cumulative number of E-step epochs reaches 200.

\subsection{Analytical Comparison with Existing Adaptive Weighting Approaches}
\label{ssec:lossAnalysis}

The optimization problem in \eqref{eq:pie_total_loss} can be simplified as
\begin{align}
\min_{\Phi}
\left[
\min_{\Theta}
\sum_{k\in \{u,\varepsilon,f\}}
\lambda_k
l(\bm{r}_k;\Theta)
+
C(\Phi)
\right],
\label{eq:prob_loss_in_simple_form}
\end{align}
where $l(\cdot)$ is a residual loss function, $\bm{r}_k$ denotes the residual of the $k$th type, and $\lambda_k\in\Phi$ is the corresponding inverse-scale weight  induced by $\Phi$. In our problem, $l(\cdot)$ is the $L_1$ norm, and $\lambda_k$ corresponds to $1/b_k$. For simplicity, \eqref{eq:prob_loss_in_simple_form} assumes a scalar scale parameter for the displacement residual, although the proposed displacement observation model uses pointwise scales.

Conventional PINNs minimize
\begin{align}
\min_{\Theta}
\sum_{k\in \{u,\varepsilon,f\}}
\lambda_k
l(\bm{r}_k;\Theta)
\end{align}
with prescribed weights $\{\lambda_k\}$. The proposed method, on the other hand, estimates the weights that best describe not only the residuals from data fitting, such as $r_u$, but also the residuals from physics consistency, such as $r_f$, under the hierarchical probabilistic model.

In contrast, Self-Adaptive PINN (SA-PINN), one of the widely used adaptive PINN approaches among various dynamic weighting strategies \citep{Deresse2025, Gao2025, Xiang2022, Anagnostopoulos2024, McClenny2023, Song2024}, formulates the learning problem as a minimax optimization problem \citep{McClenny2023}:
\begin{align}
\max_{\lambda_k>0}
\min_{\Theta}
\sum_k
\lambda_k
l(\bm{r}_k;\Theta).
\label{eq:self_adaptive_PINN}
\end{align}
SA-PINN assigns higher weights to larger residuals to reduce them. This strategy may work for noiseless observations, such as observations from numerical simulations, especially for simpler forward PINN problems. However, when the problem is ill-posed with noisy observations, as in the present inverse problem, training becomes vulnerable to noise, and this vulnerability can worsen when larger weights are assigned to larger residuals. Similar pointwise adaptive-weighting approaches are used in \cite{Chen2026}, where pointwise weights are assigned instead of loss-term-wise weights.

Our proposed method differs fundamentally from this minimax residual-amplification strategy because it minimizes the objective with respect to both the network parameters and the scale-induced weights, which is made well-defined by $C(\Phi)$. 
Without $C(\Phi)$, a direct min-min formulation would lead to the trivial solution $\lambda_k=0$ for all $k$. 
The term $C(\Phi)$ makes the min-min formulation well-defined. 
As a result, persistently large residuals can be assigned smaller weights, equivalently larger scales, making the model more stable against noisy observations and more robust to errors while maintaining the likelihood-based interpretation of the loss weights.


\begin{algorithm}[!t]
\caption{Probabilistic model-driven adaptive weighting algorithm for PIE-PINN training}
\label{alg:em_piepinn}
\begin{algorithmic}
\REQUIRE Observed displacement data $\bm{u}$; initial function-approximator parameters $\Theta=\{\nnParaU, \splParaU, \nnParae, \nnParaE\}$; and initial probabilistic distribution parameters $\Phi=\{\bDISP, \uEta, \uTau, \bSTRAIN, \bPDE\}$
\REPEAT
    \STATE \textbf{E-step:}
    \STATE Fix $\Phi$ and solve
    \[
    \min_{\Theta}
    \mathcal{L}_{\text{total}}(\Theta
    \mid \bm{u}, \Phi)
    \]

    \STATE \textbf{M-step:}
    \STATE Fix $\Theta$ and solve
    \[
    \min_{\Phi}
    \mathcal{L}_{\text{total}}(\Phi
    \mid \bm{u}, \Theta)
    \]
\UNTIL{convergence}
\end{algorithmic}
\end{algorithm}

%% file: Contents/4_caseStudies.tex
\section{Case studies}\label{sec:caseStudies}

We systematically evaluate the performance of the proposed PIE-PINN model using a benchmark dataset from the ElastNet study \citep{Chen2023}, in which the ground-truth spatial distribution of Young’s modulus exhibits a dragon-shaped pattern, while Poisson’s ratio follows a dog-shaped pattern. The corresponding displacement fields are simulated using the finite element method (FEM).
The displacement observations are then preprocessed by applying spatial downsampling to reduce the observation resolution, followed by the addition of zero-mean Gaussian noise to emulate noisy measurement conditions, as detailed in Supplementary Note \ref{sup:dataDescription}. Unless otherwise stated, the downsampling factor is initially set to 2, corresponding to a 50\% reduction in spatial resolution. The noise standard deviation is set to 1\% of the average displacement magnitude, resulting in a signal-to-noise ratio (SNR) of 100.

We compare the proposed model with five benchmark methods, which can be categorized into two groups: (1) state-of-the-art PINN-based models designed for inverse elasticity problems and (2) state-of-the-art general PINN-based models that have been widely applied to physics-constrained inverse problems. Brief descriptions of the benchmark methods used in the numerical study are provided below.

\noindent\textbf{Our Proposed Method:}
 \begin{itemize}
    \item \textbf{PIE-PINN} (Probabilistic Inverse Elasticity Physics-Informed Neural Network): Our proposed method employs a probabilistic PINN framework trained with an adaptive loss-weighting capability. The model is enhanced by a sparsity-promoting loss function and a B-spline-guided neural network used as the function approximator for robust estimation under noisy and low-resolution data.
\end{itemize}

\noindent\textbf{State-of-the-art PINN for inverse elasticity:}
 \begin{itemize}
    \item \textbf{IE-PINN} (Inverse Elasticity Physics-Informed Neural Network): IE-PINN is a state-of-the-art PINN-based method for the inverse elasticity problem that estimates elastic properties from noisy displacement observations \citep{Srikitrungruang2025}. 
    The framework requires prespecified loss weights, limiting its adaptability to different noise characteristics in the data, and it has not been designed for observations that are both noisy and low-resolution.
    \item \textbf{Self-Adaptive IE-PINN} (Self-Adaptive Inverse Elasticity Physics-Informed Neural Network): 
    This benchmark is an extension of IE-PINN where the loss-term weights are not fixed as in the original formulation. Instead, we introduce an adaptive weighting strategy inspired by the Self-Adaptive Physics-Informed Neural Networks framework \citep{McClenny2023}. In this approach, the weights are trained jointly with the neural network parameters through a minimax optimization objective. During training, the neural network parameters are optimized to minimize the loss, whereas the weights are optimized to maximize it, encouraging larger weights for loss terms with higher residual errors. This strategy improves the adaptability of IE-PINN to different training conditions. 
    \item \textbf{ElastNet} : ElastNet is one of the earliest deep-learning approaches for inverse elasticity with spatially heterogeneous elastic fields based on displacements \citep{Chen2023}. The method demonstrates successful estimation when noise-free displacement observations are available, and it directly utilizes the observed displacement field to estimate elasticity parameters. However, its performance degrades significantly when the displacement data are contaminated with noise, because noise propagates through numerical differentiation to the strain and stress calculations, leading to unstable elasticity estimation. Furthermore, ElastNet requires the true mean Young's modulus to recover absolute elasticity values, which is impractical in real-world applications.
    \item \textbf{EI-UNet}: EI-UNet is another deep-learning-based framework for inverse elasticity that employs a convolutional neural network (CNN) architecture with automatic differentiation to enforce the governing physical laws \citep{Kamali2024}. The model requires the strain field as input when the displacement observations are noisy. If the strain field is computed through numerical differentiation of noisy displacement data, the resulting strain measurements become highly noisy, which leads to inaccurate elasticity estimation. In addition, EI-UNet requires the true stress distribution to obtain accurate elasticity values, which is typically unavailable in practical applications.
\end{itemize}

\noindent\textbf{State-of-the-art general PINN-based models:}
 \begin{itemize}
    \item \textbf{gPINNs} (Gradient-enhanced physics-informed neural networks): gPINNs represent a state-of-the-art extension of standard PINNs that improves training efficiency and solution accuracy by incorporating gradient information of the PDE residual as an additional loss constraint \citep{Yu2022}. In this study, we integrate the elasticity network into the gPINN framework to enable a fair comparison under consistent modeling assumptions.
\end{itemize}

Estimation accuracy is quantified using the mean absolute error (MAE), defined as the average absolute discrepancy between the predicted and ground-truth values:
\begin{equation*}
    MAE = \frac{1}{N_E} \sum_{i=1}^{N_E}  \left( | \hat{E}_i- E_{i} | \right)   
\end{equation*}
where $\hat{E}_i$ denotes the estimated Young’s modulus, $E_{i}$ denotes the corresponding ground-truth value, and $N_E$ denotes the number of evaluation points in the high-resolution spatial grid.
The estimation accuracy of Poisson’s ratio is evaluated in the same way. The experimental hyperparameters are provided in Table~\ref{table:param_piepinn} of the Supplementary Material, and the corresponding computational costs, evaluated on a single NVIDIA RTX A2000 GPU, are summarized in Table~\ref{table:resourceComputation}. All probabilistic loss components are initially assigned weights equal to 1.0.

In the following sections, the proposed model is compared with the benchmarking models introduced in Section~\ref{ssec:benchmark}. We then systematically evaluate the robustness of the proposed approach under varying levels of observational noise and spatial resolution (Section~\ref{ssec:robustness}). In addition, ablation studies are conducted to isolate and assess the contribution of each key component of the PIE-PINN framework. Specifically, we examine the B-spline-guided displacement network (Section~\ref{ssec:importanceBS}), the horseshoe$+$ hierarchical model (Section~\ref{ssec:importanceHS}), and the probabilistic model-driven loss with adaptive weight
algorithm (Section~\ref{ssec:importanceEM}).

\subsection{Benchmark Comparison} \label{ssec:benchmark}

\begin{figure}[t]
    \centering
    \includegraphics[width=\textwidth]{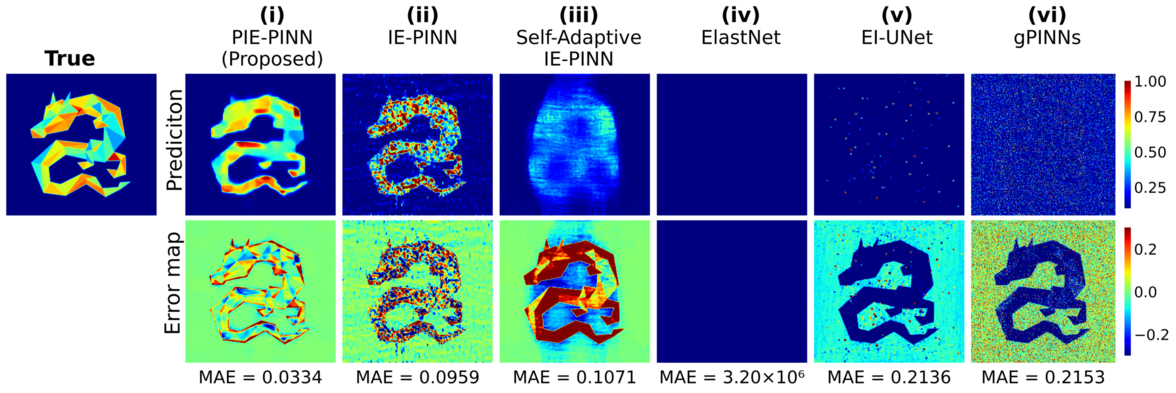}
    \justifying
    \caption{Comparison of Young’s modulus ($E$) estimates from different models inferred from noisy, low-resolution displacement data (50\% resolution, SNR = 100).}
\label{Fig:Benchmark_n2_snr100_E}
\end{figure}

\begin{figure}[t]
    \centering
    \includegraphics[width=\textwidth]{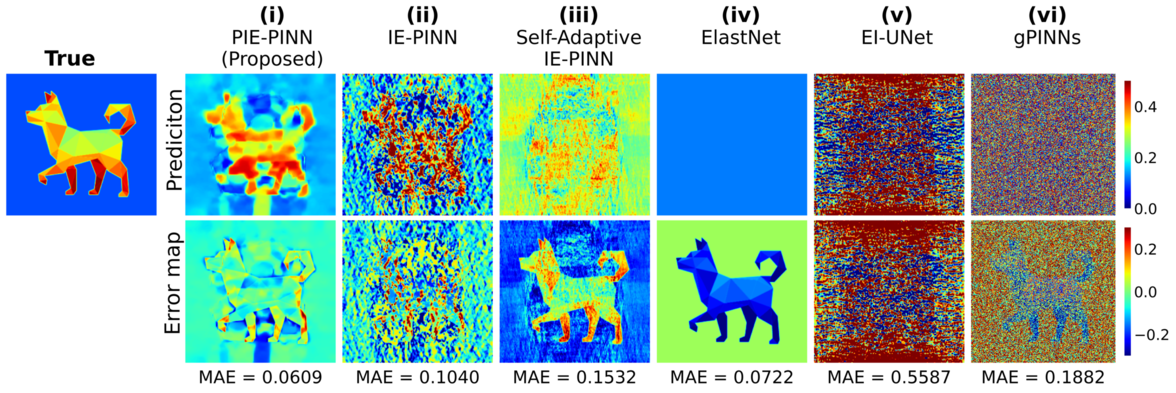}
    \justifying
    \caption{Comparison of Poisson's ratio $(\nu)$ estimates from different models inferred from noisy, low-resolution displacement data (50\% resolution, SNR = 100). }
\label{Fig:Benchmark_n2_snr100_v}
\end{figure}

The estimation fields and error of the proposed model are demonstrated in comparison to the benchmark methods  mentioned earlier in Figures~\ref{Fig:Benchmark_n2_snr100_E} and \ref{Fig:Benchmark_n2_snr100_v}. They present the estimated fields of Young’s modulus ($E$) and Poisson’s ratio ($\nu$), respectively, inferred from displacement data at 50\% spatial resolution with an SNR of 100. The standard PINN variant, gPINN, fails to accurately estimate the elastic properties due to numerical instability, as shown in Figures~\ref{Fig:Benchmark_n2_snr100_E}(vi) and \ref{Fig:Benchmark_n2_snr100_v}(vi). Similarly, the established PINN-based elasticity models, ElastNet and EI-UNet, are unable to recover the elastic parameters under these conditions, as illustrated in Figures~\ref{Fig:Benchmark_n2_snr100_E}(iv-v) and \ref{Fig:Benchmark_n2_snr100_v}(iv-v).

In comparison with IE-PINN, a recent and effective approach for elasticity estimation from noisy displacement data, the proposed PIE-PINN achieves consistently higher accuracy in estimating both Young’s modulus and Poisson’s ratio. While IE-PINN exhibits noticeable performance degradation under noisy and low-resolution conditions (Figures~\ref{Fig:Benchmark_n2_snr100_E}(ii) and \ref{Fig:Benchmark_n2_snr100_v}(ii)), PIE-PINN maintains stable and accurate reconstructions.

Moreover, the Self-Adaptive IE-PINN, which augments IE-PINN with dynamic loss weighting, fails to reliably estimate elastic properties under severe noise and limited spatial resolution. When the spatial resolution is reduced to 25\%, or the noise level is increased to an SNR of 20, IE-PINN deteriorates substantially, whereas PIE-PINN consistently demonstrates superior performance (Figures~\ref{Fig:Benchmark_n2_snr20} and \ref{Fig:Benchmark_n4_snr100}, Supplementary Material).

\begin{figure}[t!]
    \centering
    \begin{subfigure}[t!]{0.49\textwidth}
        \includegraphics[width=1.0\textwidth]{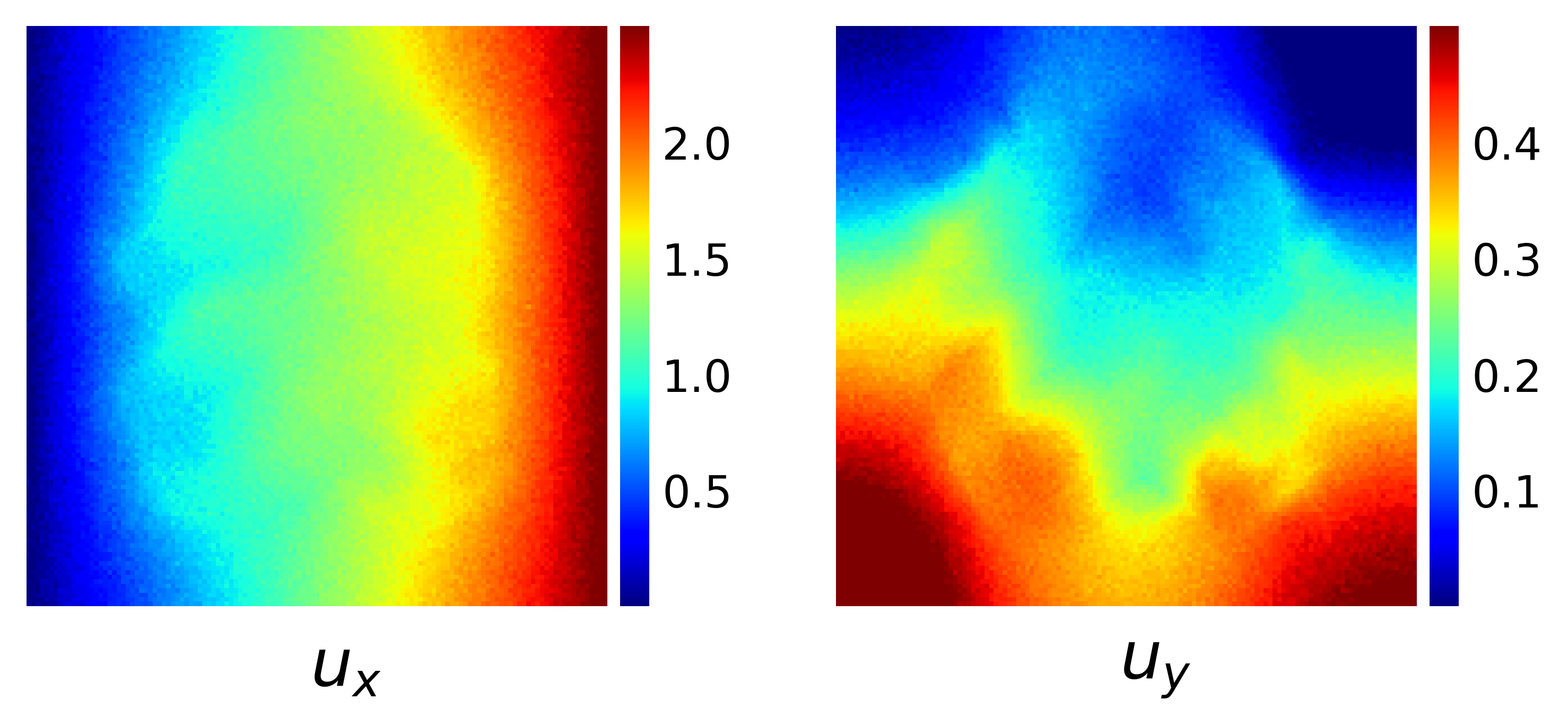}
        \caption{\centering The observed axial\\displacement (mm) fields}
        \label{fig:snr100-n2-mech-u-obs}
    \end{subfigure}
    \vspace{0.5em}
    \begin{subfigure}[t!]{0.49\textwidth}
        \includegraphics[width=1.0\textwidth]{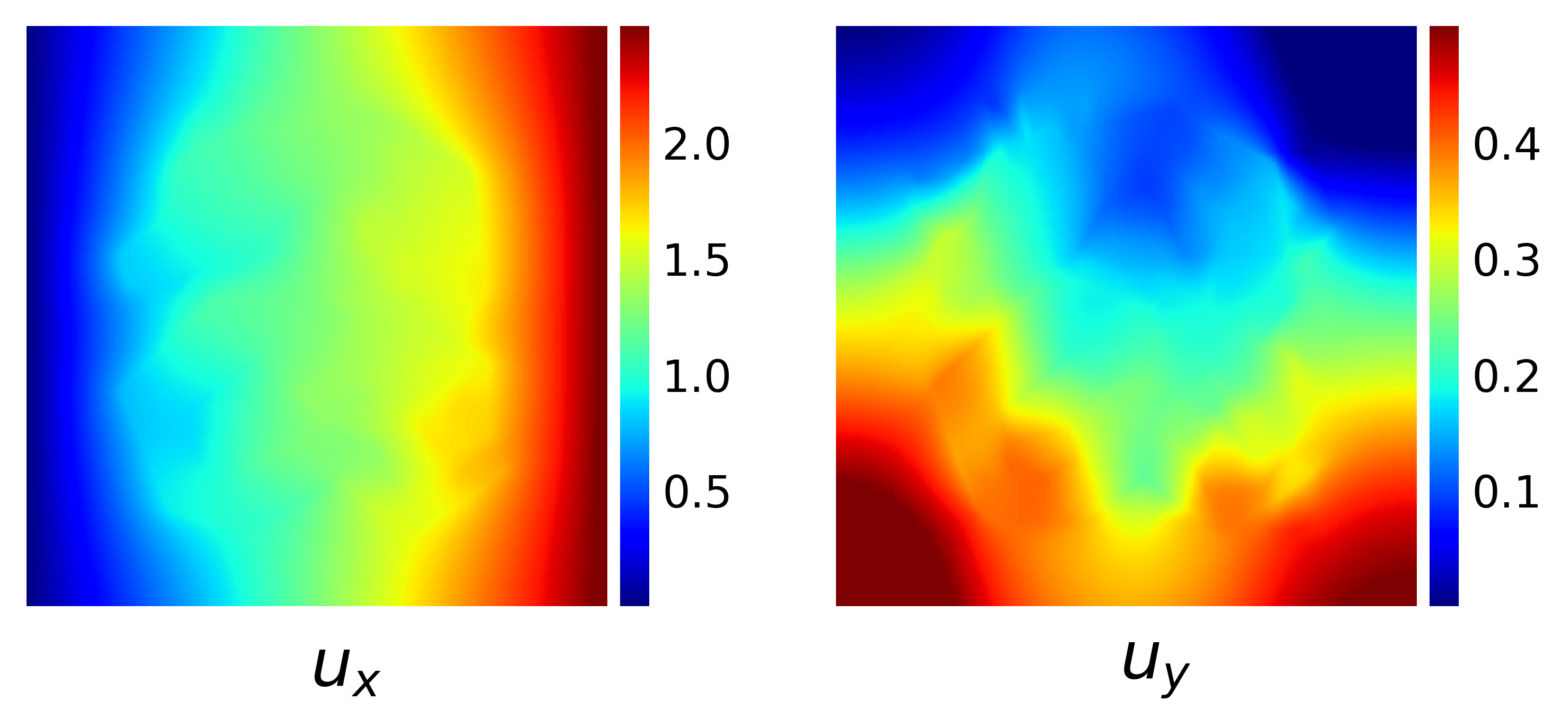}
        \caption{\centering The true axial\\displacement (mm) fields}
        \label{fig:snr100-n2-mech-u-true}
    \end{subfigure}
    \vspace{0.5em}
    \begin{subfigure}[t!]{0.49\textwidth}
        \includegraphics[width=1.0\textwidth]{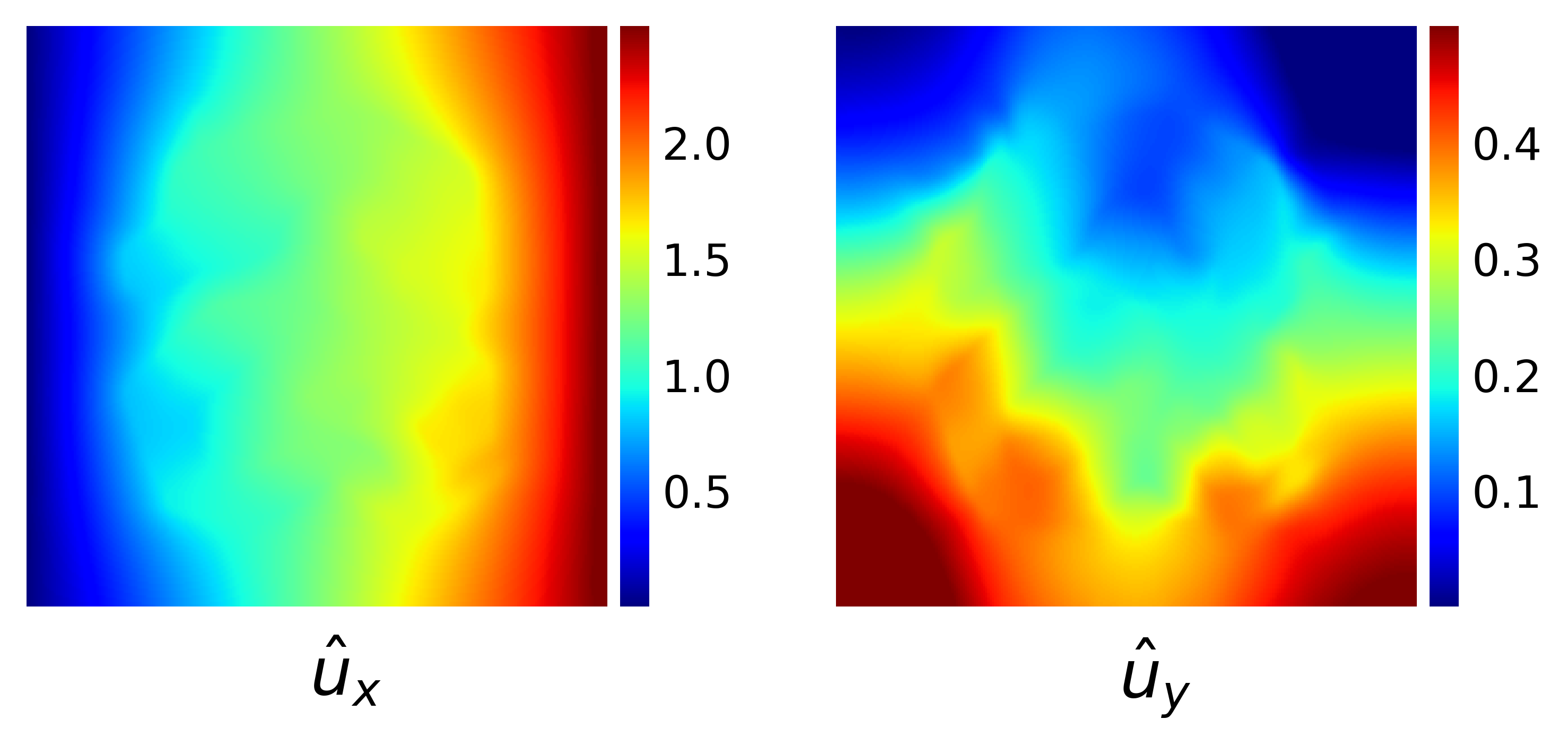}
        \caption{\centering The predicted axial\\displacement (mm) fields}
        \label{fig:snr100-n2-mech-u-pred}
    \end{subfigure}
    \vspace{0.5em}
    \begin{subfigure}[t!]{0.49\textwidth}
        \includegraphics[width=1.0\textwidth]{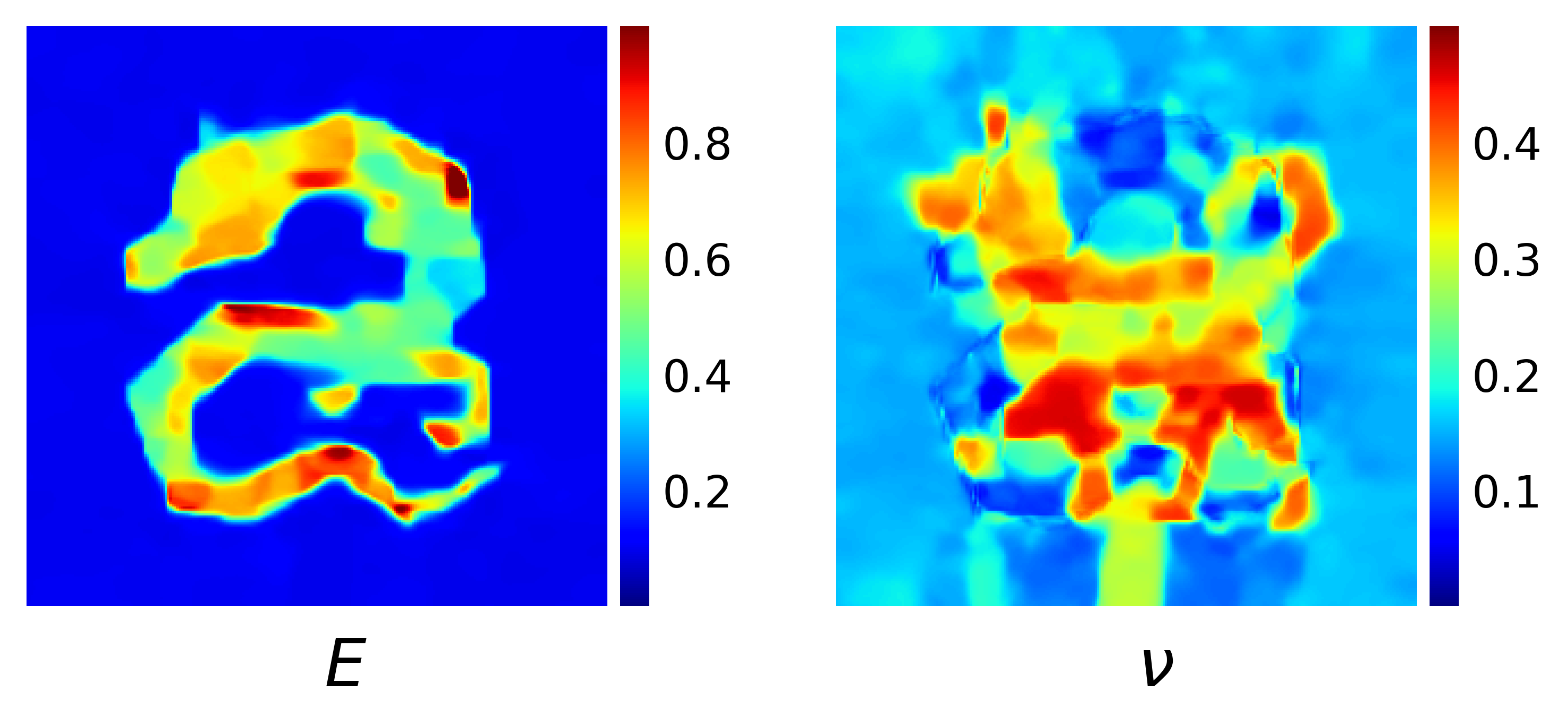}
        \caption{\centering The predicted Young’s modulus (MPa),\\ Poisson’s ratio}
        \label{fig:snr100-n2-mech-elas}
    \end{subfigure}
    \caption{Related mechanical fields in the proposed PIE-PINN under noisy, low-resolution displacement data (50\% resolution, SNR = 100).}
    \label{fig:snr100-n2-mech-new} 
\end{figure}

The effectiveness of PIE-PINN in estimating elastic properties under noisy, low-resolution observation conditions is demonstrated in Figure~\ref{fig:snr100-n2-mech-new}. Despite the degraded displacement observations shown in Figure~\ref{fig:snr100-n2-mech-u-obs}, the proposed framework can reconstruct high-resolution, denoised displacement fields (Figure~\ref{fig:snr100-n2-mech-u-pred}) that closely match the ground truth (Figure~\ref{fig:snr100-n2-mech-u-true}). This reconstruction subsequently yields accurate elasticity estimates, as illustrated in Figure~\ref{fig:snr100-n2-mech-elas}. 

The predicted fields and corresponding error maps for all mechanical quantities, including the denoised high-resolution displacement, elasticity, strains, and stresses, are provided in Figures~\ref{fig:snr100-n2-pred} and~\ref{fig:snr100-n2-error} in the Supplementary Material. These results highlight a key strength of the proposed framework: in addition to accurately estimating heterogeneous elastic properties, the model can reconstruct noise-free, high-resolution displacement fields and the associated mechanical quantities from degraded observations.

\begin{figure}[t]
    \centering
    \includegraphics[width=0.5\textwidth]{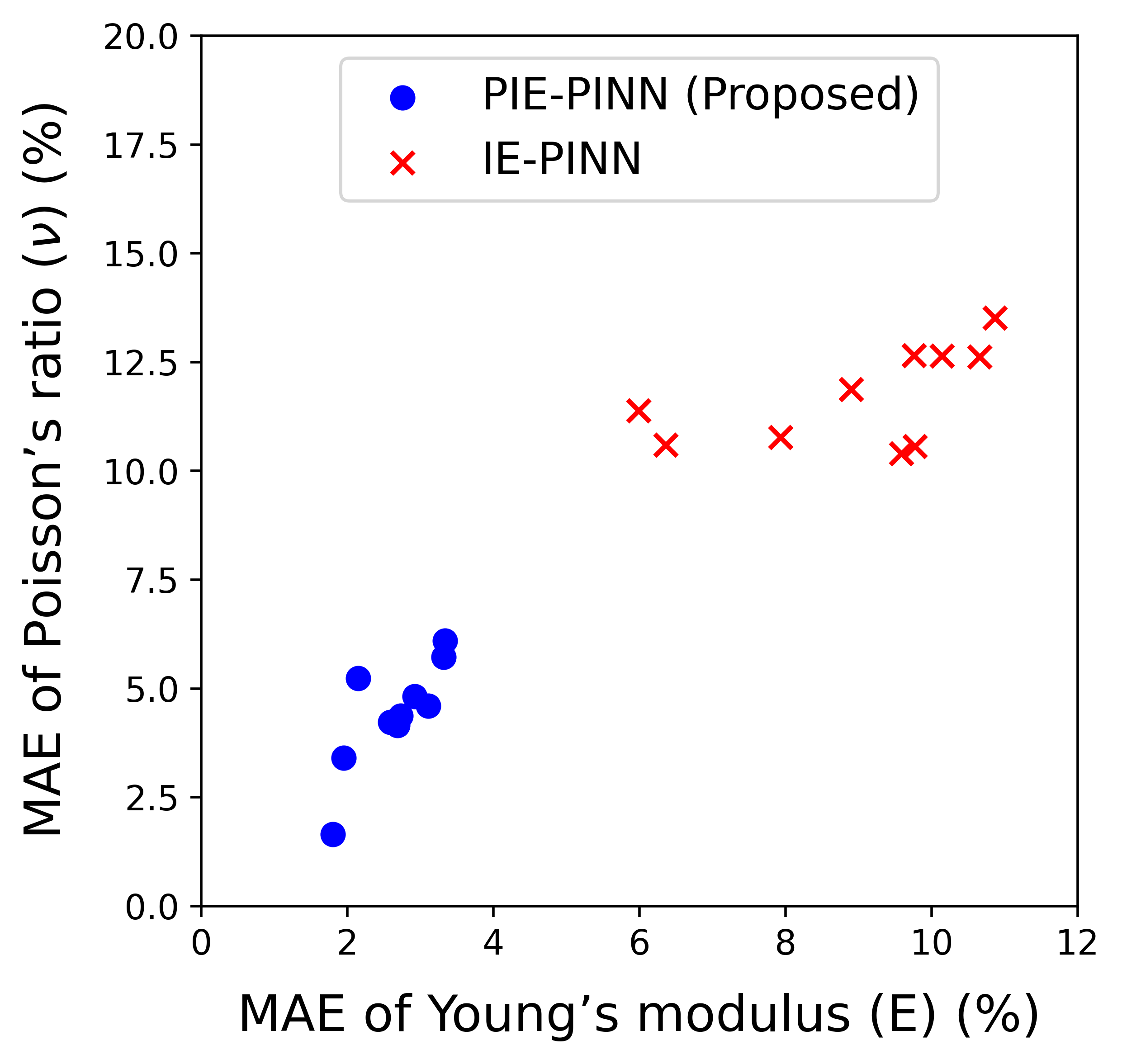}
    \caption{Elasticity estimation errors across multiple elastic distribution patterns obtained from noisy, low-resolution displacement data (50\% resolution, SNR = 100), comparing the proposed PIE-PINN and IE-PINN. }
\label{fig:robustnessDataset}
\end{figure}

Furthermore, evaluations across multiple datasets with distinct spatial elasticity distributions confirm that PIE-PINN consistently outperforms IE-PINN in both robustness and estimation accuracy (Figure~\ref{fig:robustnessDataset}).


\subsection{Robustness to noise \& resolution} \label{ssec:robustness}
The presence of noise is known to degrade estimation accuracy, as demonstrated in previous studies \citep{Srikitrungruang2025}, and low spatial resolution further exacerbates the difficulty of inverse elasticity estimation. To evaluate the robustness of the proposed model under varying noise levels and spatial resolutions, we investigate its performance from two complementary perspectives.

\begin{figure}[t!]
    \centering
    \begin{subfigure}[t!]{0.49\textwidth}
        \includegraphics[width=1.0\textwidth]{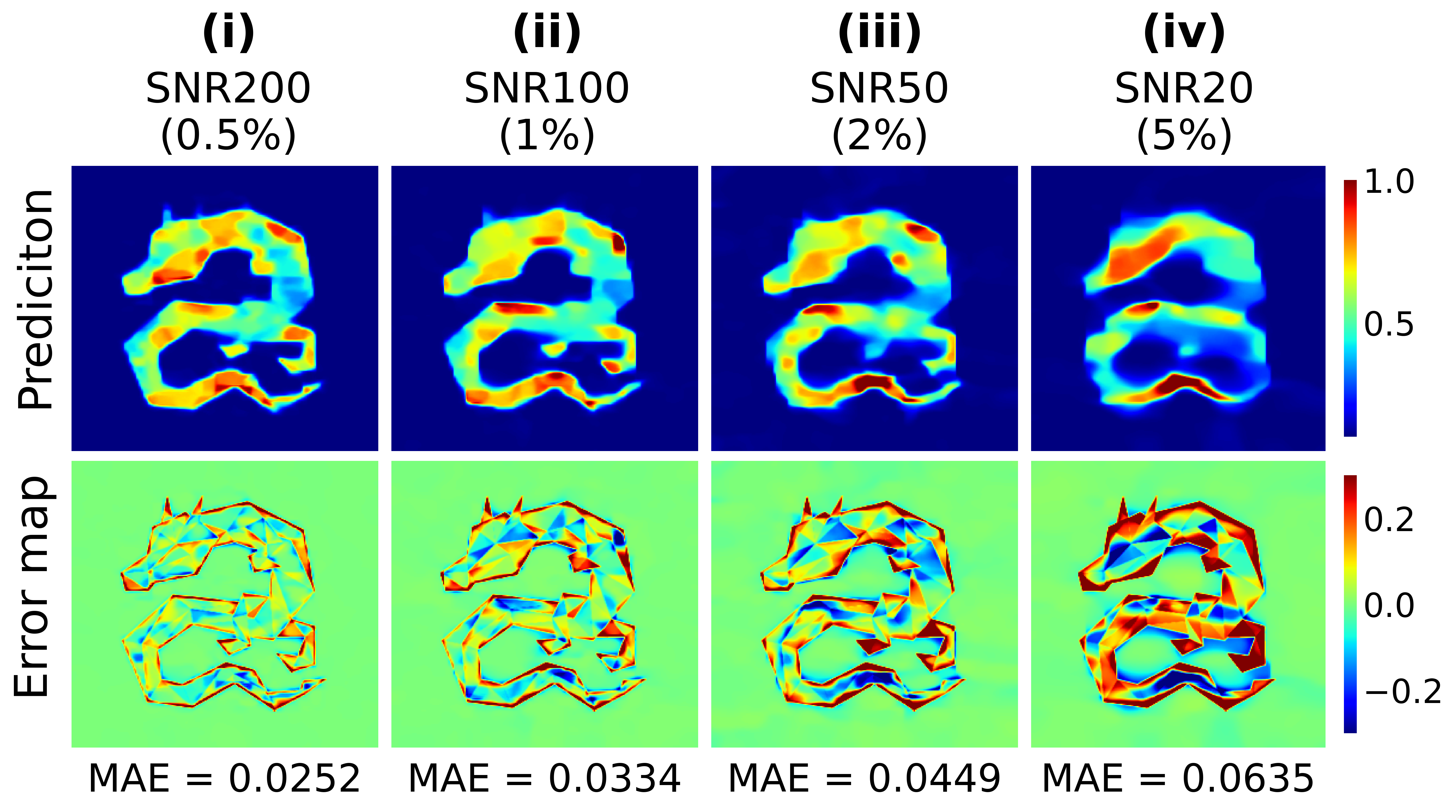}
        \caption{Estimated Young's modulus }
        \label{Fig:robustnessNoise_E}
    \end{subfigure}
    \vspace{0.5em}
    \begin{subfigure}[t!]{0.49\textwidth}
        \includegraphics[width=1.0\textwidth]{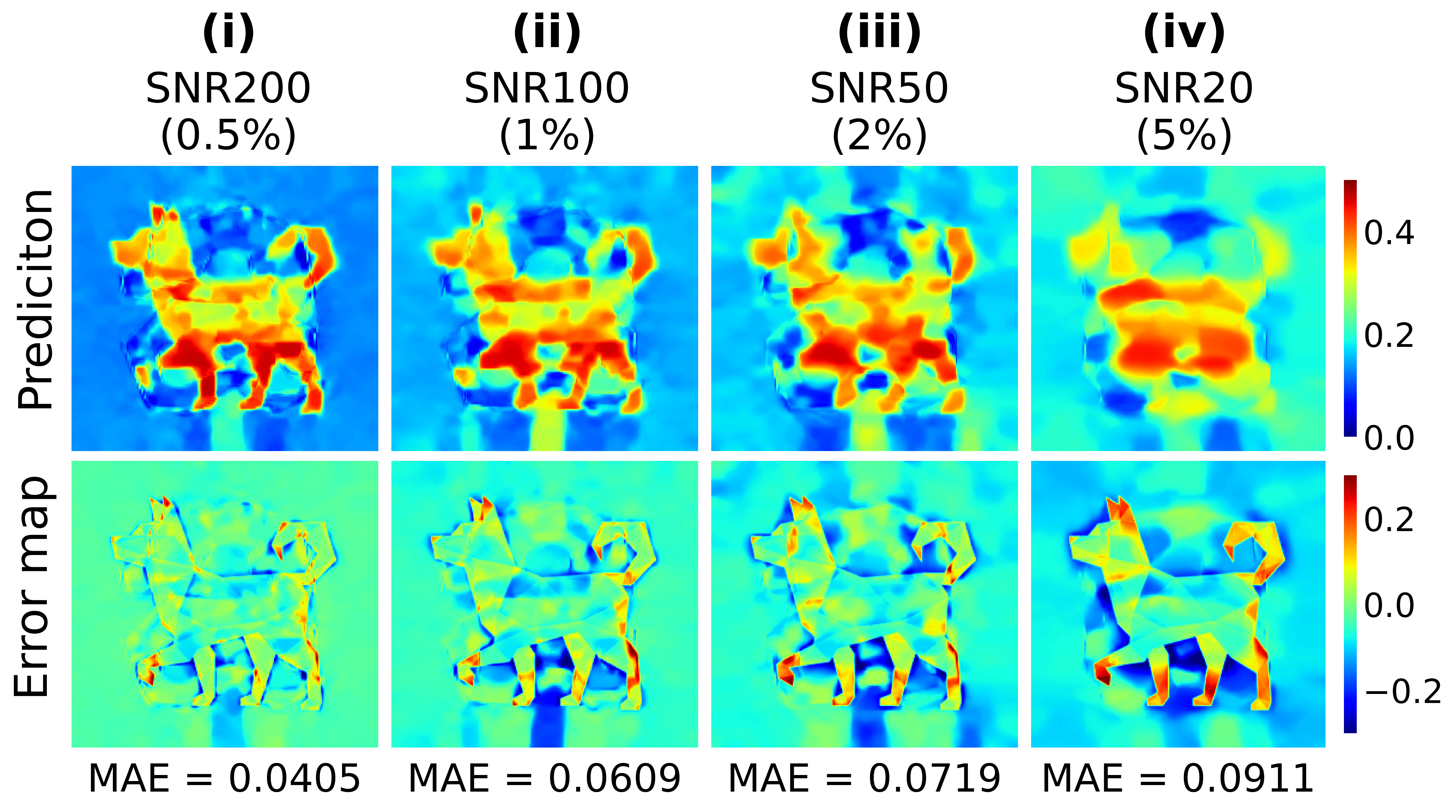}
        \caption{Estimated Poisson's ratio }
        \label{Fig:robustnessNoise_v}
    \end{subfigure}
    \caption{Elasticity maps estimated by the proposed PIE-PINN model using 50\% low-resolution displacement data across varying noise levels.}
    \label{Fig:robustnessNoise} 
\end{figure}

First, we fix the spatial resolution of the observations at 50\% and vary the noise level across four signal-to-noise ratios (SNRs): 200, 100, 50, and 20. These SNRs correspond to average displacement errors of approximately 0.5\%, 1.0\%, 2.0\%, and 5.0\%, respectively, relative to the mean displacement magnitude. As shown in Figure~\ref{Fig:robustnessNoise}, the estimation errors remain consistently low across all noise levels, indicating strong robustness in recovering elastic properties. As the noise level increases to 5\%, the errors in both Young’s modulus and Poisson’s ratio increase moderately; however, the proposed model is still able to accurately capture the overall spatial patterns and general structure of the elasticity maps.

\begin{figure}[t!]
    \centering
    \begin{subfigure}[t!]{0.49\textwidth}
        \includegraphics[width=1.0\textwidth]{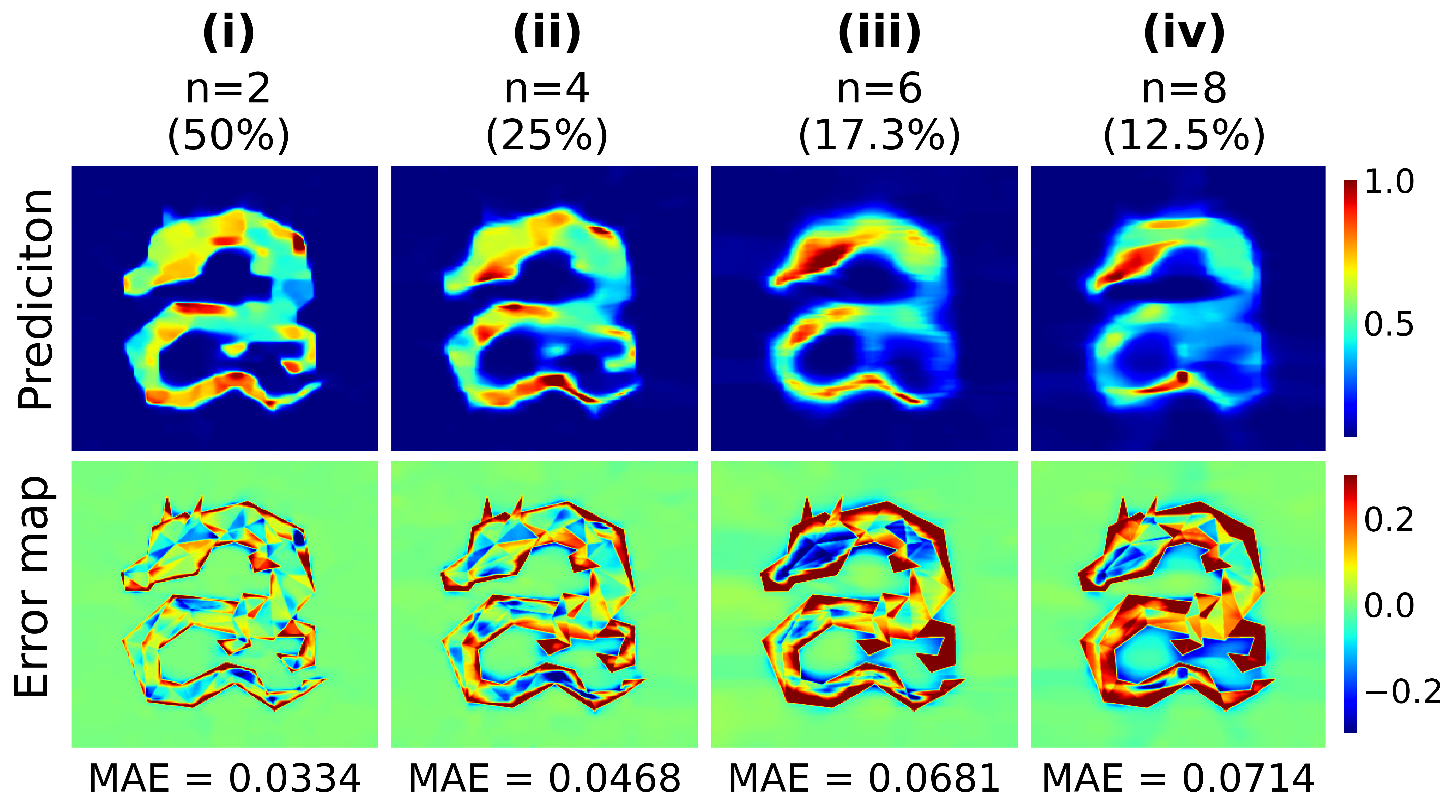}
        \caption{Estimated Young's modulus }
        \label{Fig:robustnessResolution_E}
    \end{subfigure}
    \vspace{0.5em}
    \begin{subfigure}[t!]{0.49\textwidth}
        \includegraphics[width=1.0\textwidth]{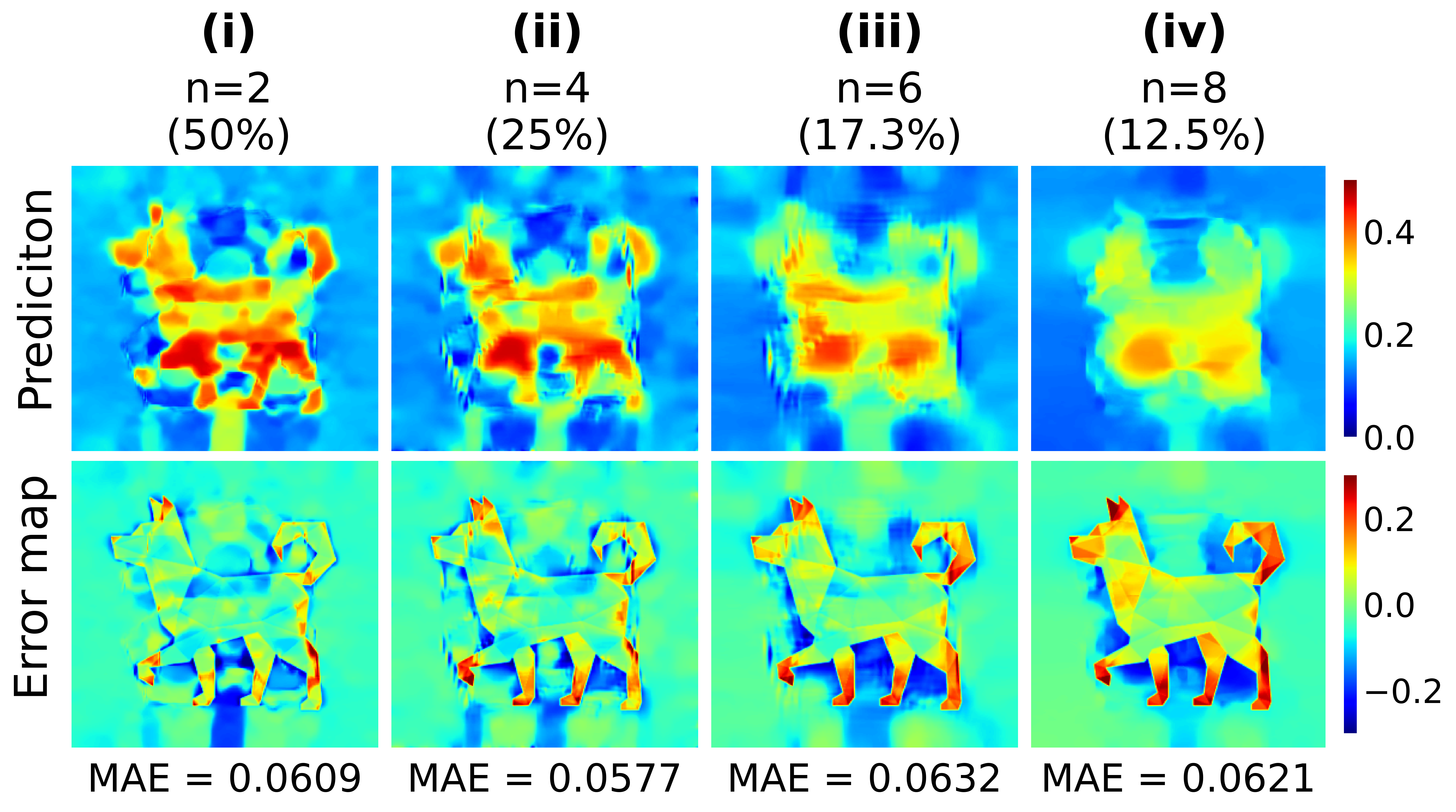}
        \caption{Estimated Poisson's ratio}
        \label{Fig:robustnessResolution_v}
    \end{subfigure}
    \caption{Elasticity maps estimated by the proposed PIE-PINN under noisy, low-resolution displacement data at SNR = 100 across varying resolutions.}
    \label{Fig:robustnessResolution} 
\end{figure}

Next, we examine robustness with respect to spatial resolution while fixing the noise level at 1\% (SNR = 100). The observation resolution is progressively reduced from 50\% to 25\%, 17.3\%, and 12.5\%. The corresponding predicted fields and error distributions are presented in Figure~\ref{Fig:robustnessResolution}. Despite a substantial reduction in available observation points, the proposed model maintains stable estimation performance, demonstrating its ability to reliably infer elastic properties even under severely low-resolution conditions.

\subsection{Importance of B-spline guided architecture}
\label{ssec:importanceBS}

\begin{figure}[t!]
    \centering
    \begin{subfigure}[t!]{0.49\textwidth}
        \includegraphics[width=1.0\textwidth]{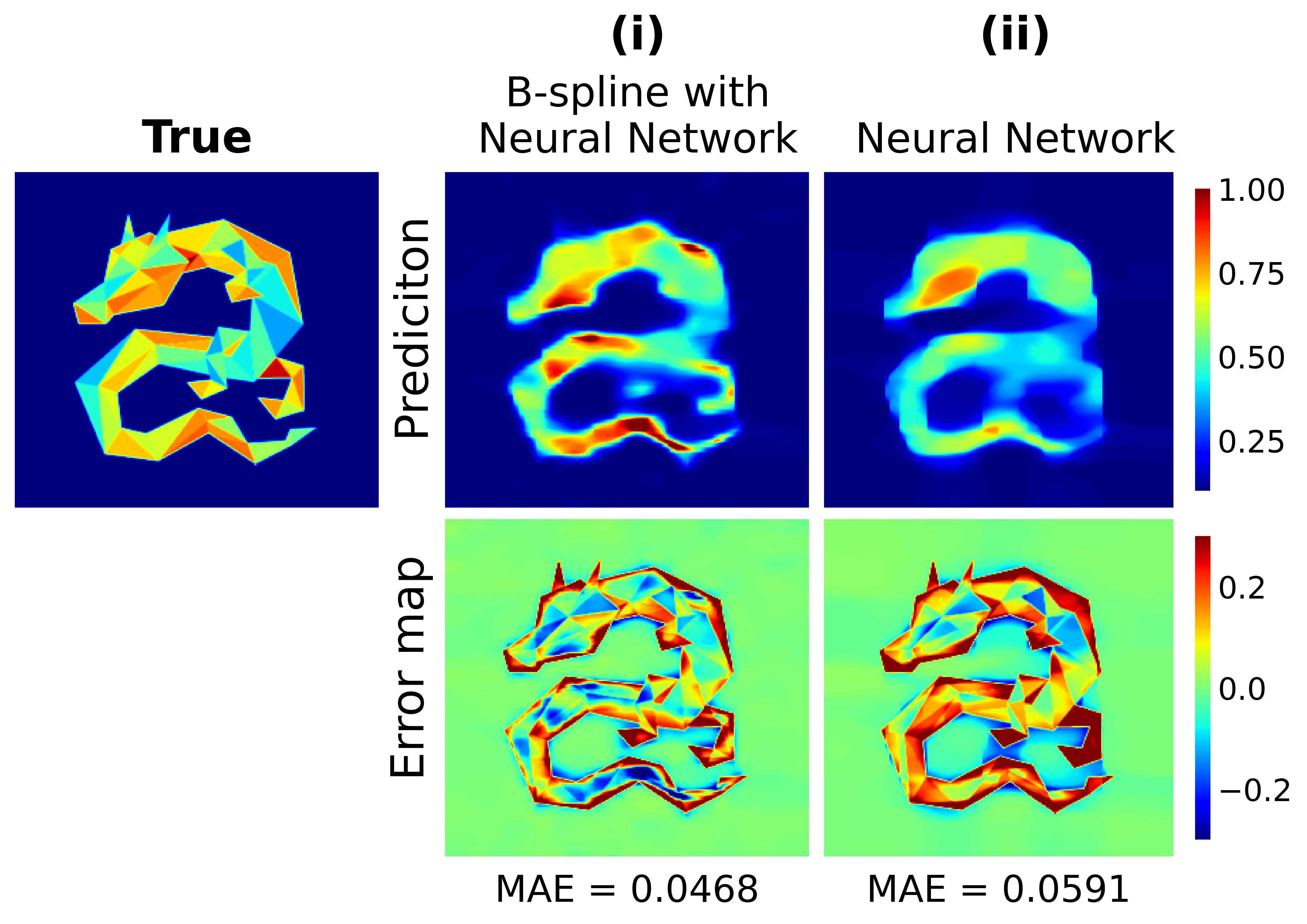}
        \caption{Estimated Young's modulus}
        \label{Fig:importanceBspline_n4_E}
    \end{subfigure}
    \vspace{0.5em}
    \begin{subfigure}[t!]{0.49\textwidth}
        \includegraphics[width=1.0\textwidth]{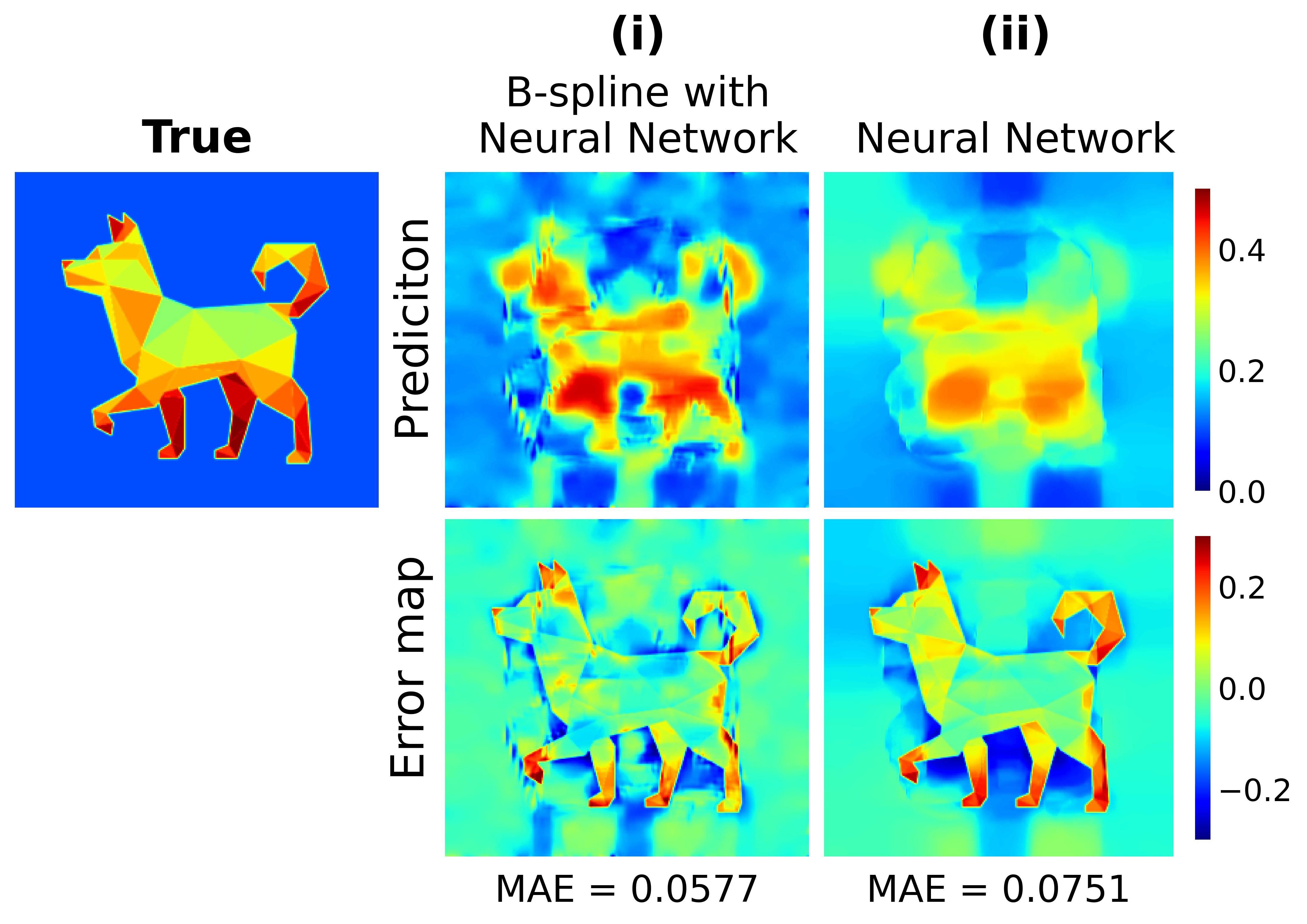}
        \caption{Estimated Poisson's ratio}
        \label{Fig:importanceBspline_n4_v}
    \end{subfigure}
    \caption{Comparison of elasticity maps inferred from noisy, low-resolution displacement data (25\% resolution, SNR = 100) between the proposed B-spline-guided neural network architecture and the standard neural network architecture.}
    \label{Fig:importanceBspline_n4} 
\end{figure}

B-spline representations are well known for their effectiveness in interpolation, particularly under sparse sampling conditions. As demonstrated by the results of the proposed model, we compare two architectures: (i) the proposed hybrid B-spline-guided neural network for displacement fitting and (ii) a standard neural network architecture without B-spline interpolation. The spatial resolution of the displacement observations is varied across 25\%, 17.3\%, and 12.5\%. As shown in Figure~\ref{Fig:importanceBspline_n4}, when the displacement observations are available at 25\% spatial resolution, the hybrid B-spline-guided neural network achieves higher estimation accuracy and successfully captures fine-scale elasticity features. In contrast, the conventional neural network produces overly smooth predictions and fails to recover detailed spatial variations. As the spatial resolution is further reduced to 17.3\% and 12.5\%, the vanilla neural network is no longer able to reliably infer elasticity, as illustrated in Figures~\ref{Fig:importanceBspline_n6} and~\ref{Fig:importanceBspline_n8} in the Supplementary Material. By comparison, the proposed B-spline-guided model remains robust and continues to recover meaningful elasticity structures even under these severely low-resolution conditions.


\subsection{Impact of the Sparsity-Enhanced distribution on Displacement Fitting Residuals}
\label{ssec:importanceHS}

\begin{figure}[t!]
    \centering
    \begin{subfigure}[t!]{0.49\textwidth}
        \includegraphics[width=1.0\textwidth]{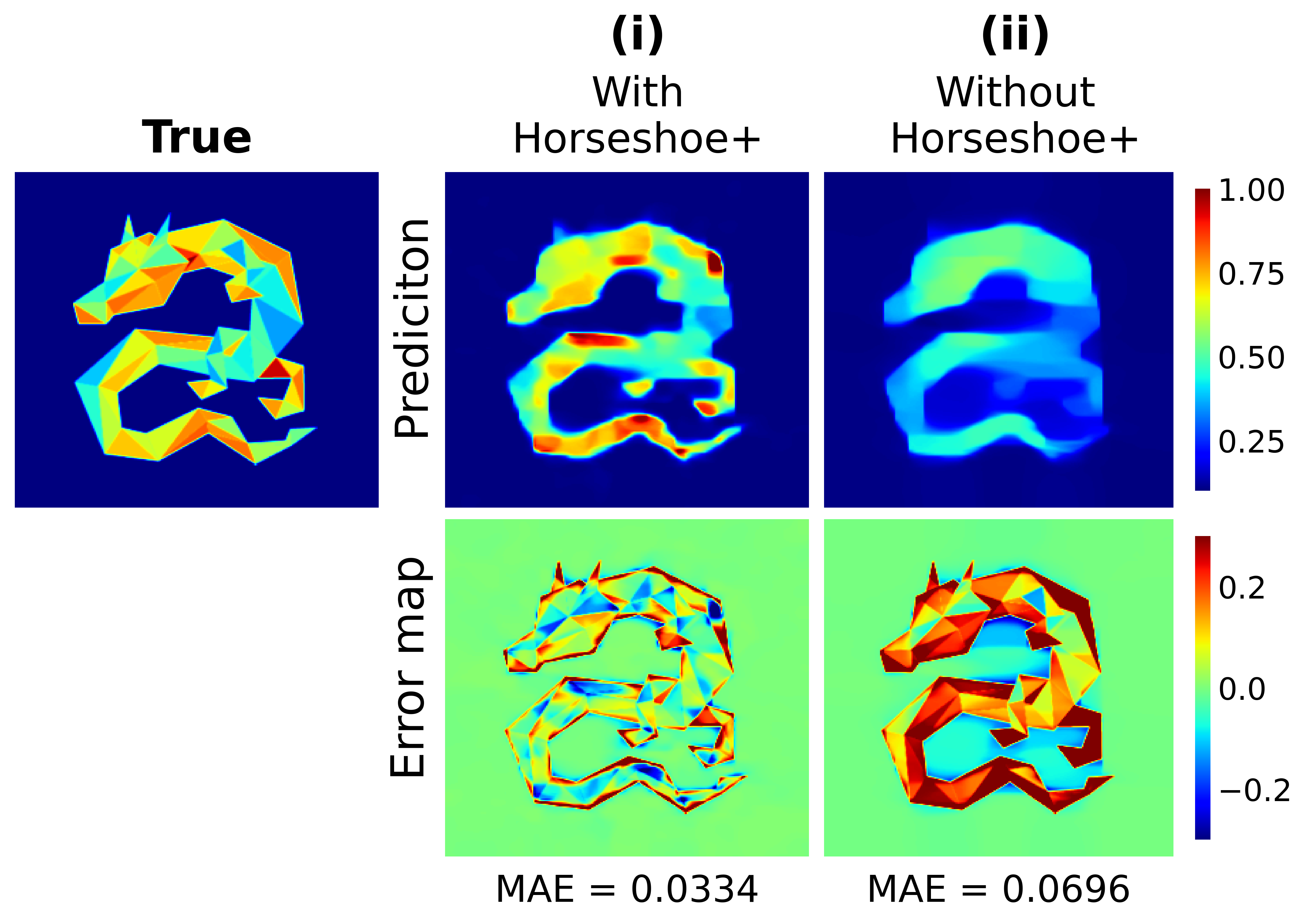}
        \caption{Estimated Young's modulus}
        \label{Fig:importanceHSPlus_E}
    \end{subfigure}
    \vspace{0.5em}
    \begin{subfigure}[t!]{0.49\textwidth}
        \includegraphics[width=1.0\textwidth]{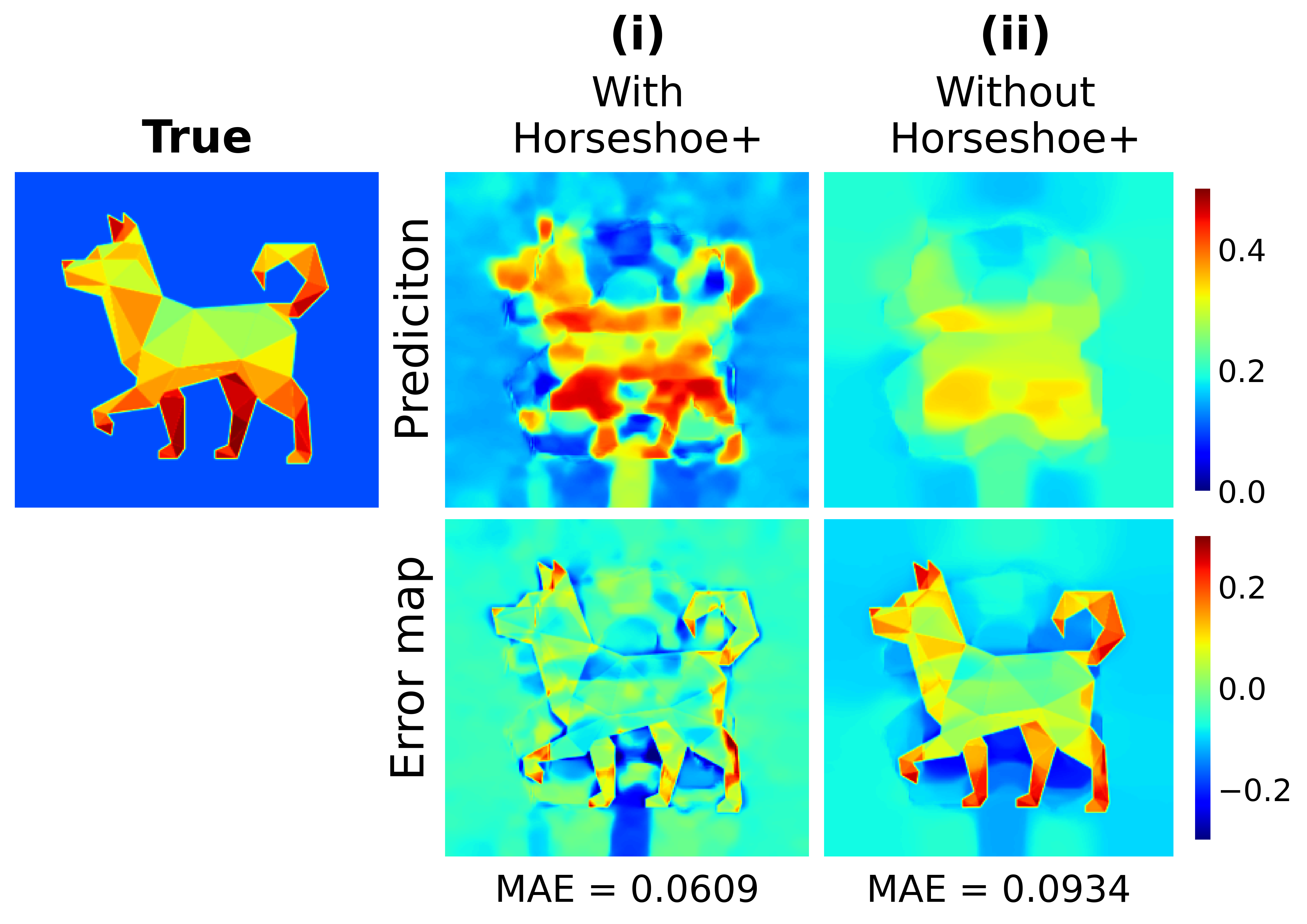}
        \caption{Estimated Poisson's ratio}
        \label{Fig:importanceHSPlus_v}
    \end{subfigure}
    \caption{Comparison of elasticity maps inferred from noisy, low-resolution displacement data (50\% resolution, SNR = 100) using the proposed model with and without the horseshoe+ hierarchy.}
    \label{Fig:importanceHSPlus} 
\end{figure}

Low-resolution observations are inherently sparse, and when trained within the PINN framework, the physics-based loss acts as a regularizer that constrains the solution to satisfy the governing physical laws. However, under sparse low-resolution observations, this regularization often leads to overly smooth estimates, which can obscure fine-scale spatial features. To address this limitation, we impose a horseshoe+ hierarchical distribution on the pointwise scale parameters of the displacement likelihood. 
The importance of the horseshoe+ hierarchy is demonstrated by estimating the elasticity fields using the proposed model
both with and without the horseshoe+ hierarchical model, under noisy and low-resolution displacement observations (50\% spatial resolution, SNR = 100). As shown in Figure~\ref{Fig:importanceHSPlus}(a.ii) and (b.ii), the model without the horseshoe+ hierarchy produces overly smooth predictions and fails to recover detailed elasticity structures. In contrast, when the horseshoe+ hierarchy is incorporated, the proposed model yields more accurate estimates and successfully captures fine-scale spatial details, as illustrated in Figure~\ref{Fig:importanceHSPlus}(a.i) and (b.i).

\subsection{Importance of the Probabilistic Loss with the adaptive weight algorithm}
\label{ssec:importanceEM}

\begin{figure}[t!]
    \centering
    \includegraphics[width=0.8\textwidth]{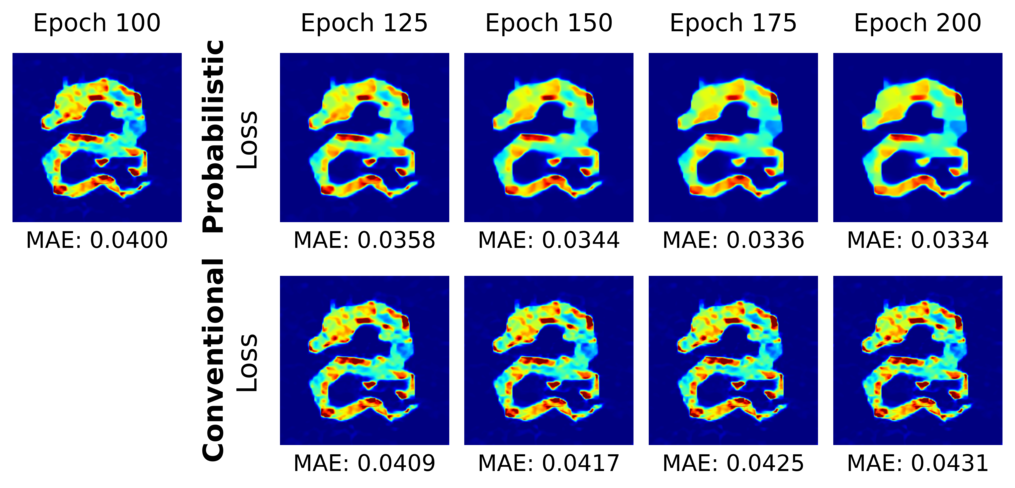}
    \caption{Evolution of Young’s modulus estimates inferred from noisy, low-resolution displacement data (50\% spatial resolution, SNR = 100) using the proposed architecture. Results are compared between the conventional deterministic loss and the proposed probabilistic loss combined with the EM-style adaptive weight algorithm. The conventional loss leads to overfitting, producing coarse and unstable elasticity fields, whereas the proposed approach yields smoother and more physically consistent estimates.}
\label{fig:evolution_n2_snr100_E}
\end{figure}

\begin{figure}[t!]
    \centering
    \includegraphics[width=0.8\textwidth]{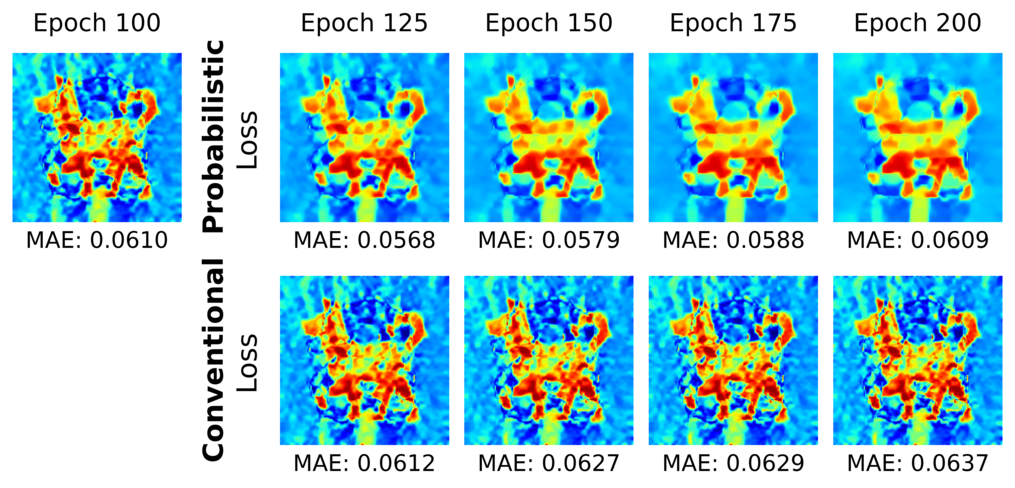}
    \caption{Evolution of Poisson's ratio estimates inferred from noisy, low-resolution displacement data (50\% spatial resolution, SNR = 100) using the proposed architecture. Results are compared between the conventional deterministic loss and the proposed probabilistic loss combined with the EM-style adaptive weight algorithm. The conventional loss exhibits overfitting, resulting in coarse and unstable estimates, whereas the proposed approach yields smoother and more physically consistent Poisson’s ratio fields, consistent with the behavior observed for Young’s modulus.}
\label{fig:evolution_n2_snr100_v}
\end{figure}

Conventionally, PINNs are trained using a weighted sum of residual-based loss terms with fixed, manually specified weights. However, improper weight initialization can lead to inaccurate estimates or even training failure in elasticity identification problems. Moreover, fixed weighting schemes lack adaptability to data quality and noise levels. To address these limitations, we introduce a probabilistic loss with an adaptive weight algorithm that enables dynamic, data-driven weight adaptation during training. To highlight the importance of the proposed probabilistic loss with adaptive weight algorithm, we evaluate elasticity estimation performance using noisy, low-resolution displacement data (50\% spatial resolution, SNR = 100), as shown in Figures~\ref{fig:evolution_n2_snr100_E} and~\ref{fig:evolution_n2_snr100_v}. The results demonstrate that the proposed model can effectively adapt loss weights during training and yield more accurate elasticity estimates compared with the conventional fixed-weight approach. When the noise level is further increased (SNR = 20), the proposed model continues to recover the general structure of the elasticity fields, as illustrated in Figures~\ref{fig:evolution_n2_snr20_E} and~\ref{fig:evolution_n2_snr20_v} in the Supplementary Material.

\begin{figure}[t!]
    \centering
    \begin{subfigure}[t!]{0.49\textwidth}
        \includegraphics[width=1.0\textwidth]{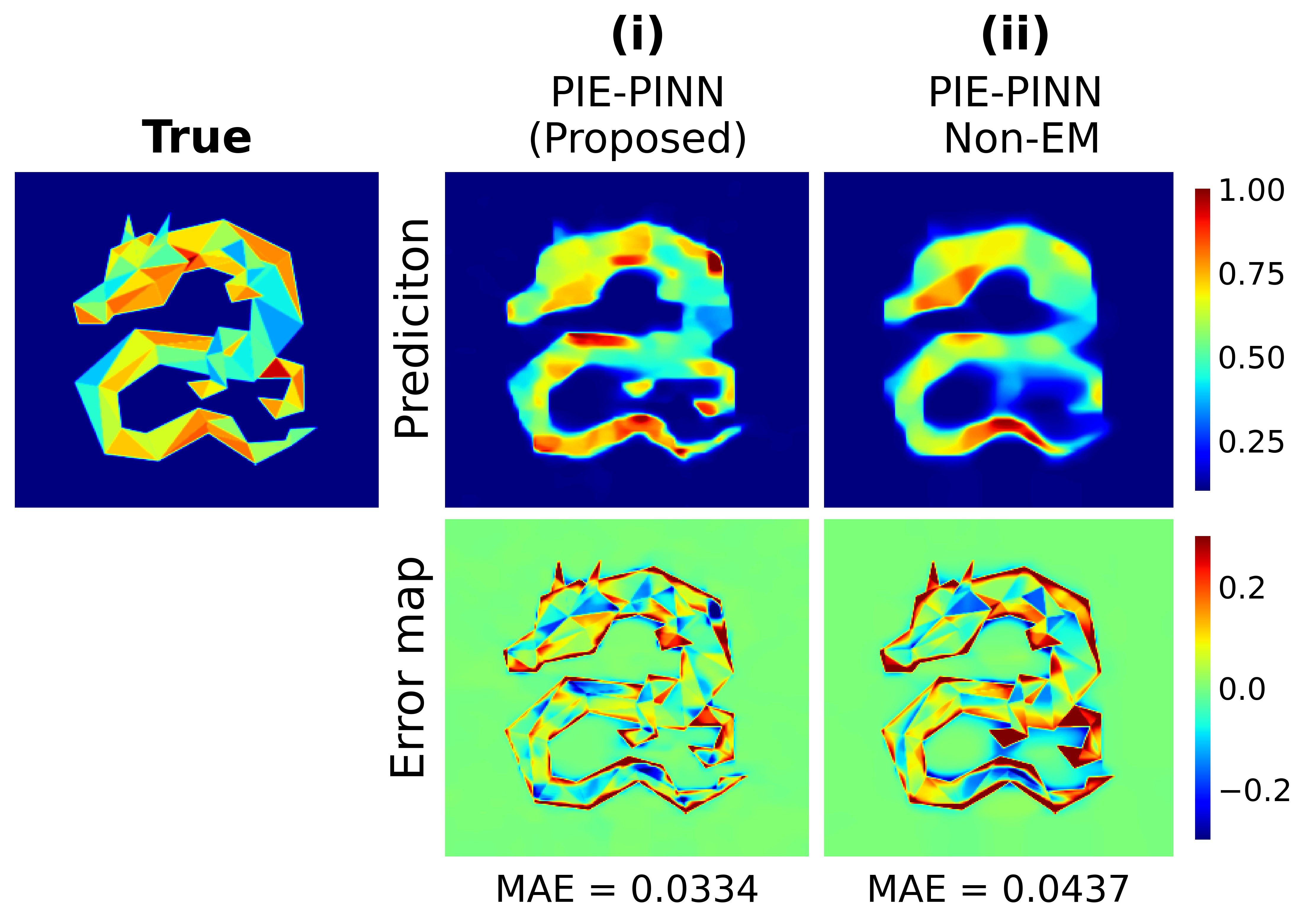}
        \caption{Estimated Young's modulus}
        \label{fig:ProblossWithEM_n2_snr100_E}
    \end{subfigure}
    \vspace{0.5em}
    \begin{subfigure}[t!]{0.49\textwidth}
        \includegraphics[width=1.0\textwidth]{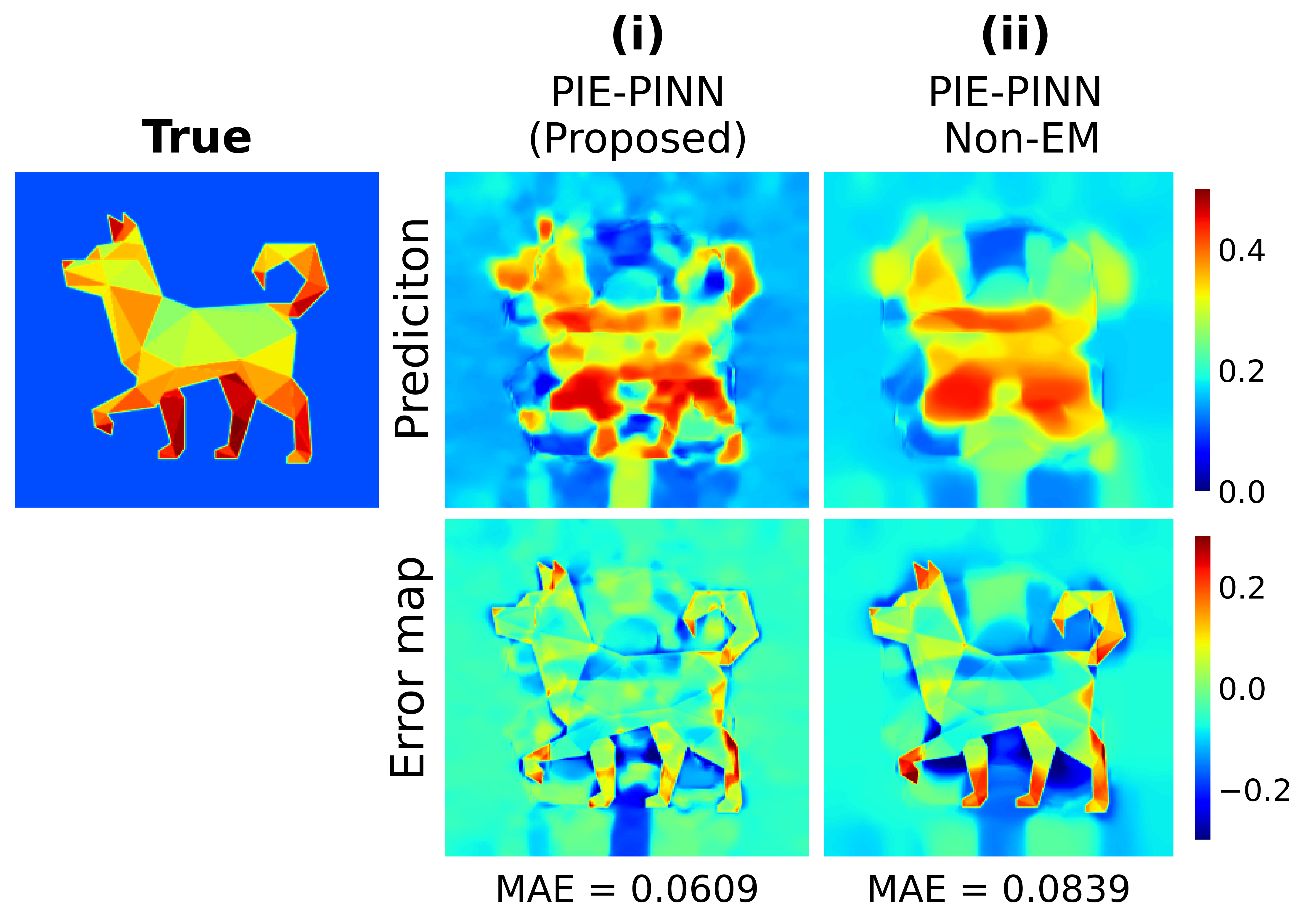}
        \caption{Estimated Poisson's ratio}
        \label{fig:ProblossWithEM_n2_snr100_v}
    \end{subfigure}
    \caption{Comparison of elasticity maps inferred from noisy, low-resolution displacement data (50\% spatial resolution, SNR = 100) using the proposed model with the probabilistic loss, with and without the EM-style adaptive weight algorithm.}
    \label{fig:ProblossWithEM_n2_snr100} 
\end{figure}

To further isolate the effect of the probabilistic model-driven adaptive weight algorithm inspired by the EM algorithm, we conduct additional experiments comparing the elasticity estimation obtained using the proposed probabilistic loss with and without the EM-style adaptive weight algorithm. Figure~\ref{fig:ProblossWithEM_n2_snr100} presents results under noisy, low-resolution observations (50\% resolution, SNR = 100). The model incorporating the adaptive weight algorithm achieves higher accuracy than the variant without EM, in which the function approximator parameters and distribution parameters are trained simultaneously. Under more severely degraded observation conditions, the model without EM-style adaptive weight algorithm fails to reliably learn elasticity, whereas the proposed EM-based approach remains robust and successfully captures fine-scale elasticity features, as shown in Figures~\ref{fig:ProblossWithEM_n2_snr20} (50\% resolution, SNR = 20) and~\ref{fig:ProblossWithEM_n4_snr100} (25\% resolution, SNR = 100) in the Supplementary Material.

%% file: Contents/5_conclusion.tex
\section{Conclusion} \label{sec:conclusion}
In this work, we present a Probabilistic Inverse Elasticity.
Physics-Informed Neural Networks (PIE-PINN) for estimating heterogeneous elastic properties, including Young’s modulus and Poisson’s ratio, from noisy and low-resolution displacement data. Challenges arising from limited observations (low-resolution data), severe measurement noise, and heuristic loss-weighting strategies are addressed through a probabilistic physics-informed formulation of the loss function, coupled with a carefully designed B-spline-guided neural network architecture. All residual components in the loss function are modeled probabilistically. In particular, the residual of the displacement fitting is further endowed with the horseshoe+ distribution to promote sparsity, thereby mitigating underfitting under noisy and limited observation settings. Unlike fully Bayesian frameworks that target posterior uncertainty quantification and often incur substantial computational cost, the proposed formulation focuses on enhancing estimation accuracy while maintaining computational efficiency.

In addition, a training scheme of probabilistic model-driven adaptive weighting inspired by the Expectation-Maximization algorithm was introduced to adaptively balance the contributions of individual loss components, enabling robust learning under harsh observational conditions. Numerical results demonstrated that PIE-PINN consistently outperforms existing methods across a range of noise levels and observation resolutions. These results suggest that the proposed framework is well-suited for practical inverse elasticity problems, particularly in applications such as medical imaging and in-situ material characterization, where measurements are inherently noisy and of limited resolution due to equipment or environmental constraints.

Despite its promising performance, several directions merit further investigation. First, extending the framework to three-dimensional elasticity and nonlinear material models would broaden its applicability to more complex real-world scenarios. Second, incorporating principled uncertainty quantification within a Bayesian framework while maintaining computational efficiency is an important avenue for future research. Additionally, improving scalability for large-scale domains and exploring adaptive spline resolutions may further enhance the flexibility and practicality of the proposed approach. Pursuing these directions will advance probabilistic physics-informed learning for challenging inverse problems in computational mechanics.

%% file: Contents/6_suffix.tex
\section{Disclosure statement}\label{sec:disclosure-statement}

The authors declare no conflict of interest.

\section{Data Availability Statement}\label{sec:data-availability-statement}

The code supporting the findings of this study will be made available in a public repository upon publication of the article. The data analyzed in this study are derived from publicly available resources associated with DOI: https://doi.org/10.1073/pnas.2102721118.



%% file: Contents/9__supplementary.tex
\setcounter{page}{1}

\newcounter{supnote}
\renewcommand{\thesupnote}{S\arabic{supnote}}
\newcommand{\supnote}[1]{
  \refstepcounter{supnote}
  \subsection*{Supplementary Note \thesupnote: #1}
  \addcontentsline{toc}{subsection}{Supplementary Note \thesupnote: #1}
}

\renewcommand{\theequation}{S\arabic{equation}}

\setcounter{figure}{0}
\renewcommand{\thefigure}{S\arabic{figure}}
\captionsetup[figure]{
  labelformat=simple,
  labelsep=colon,
  name={Supplementary Figure},
  labelfont=bf,
  textfont=normalfont,
  width=0.95\linewidth
}

\setcounter{table}{0}
\renewcommand{\thetable}{S\arabic{table}}
\captionsetup[table]{
  labelformat=simple,
  labelsep=colon,
  name={Supplementary Table},
  labelfont=bf,
  textfont=normalfont,
  width=0.95\linewidth
}

\begin{center}
    {\large\bf SUPPLEMENTARY MATERIAL}
\end{center}

\section*{Supplementary Notes}

\input{Contents/Supplementary/sup_FD}

\newpage
\input{Contents/Supplementary/sup_dataDescriptions}

\clearpage
\section*{Supplementary Figures}
\input{Contents/9_2_sup_fig}

\clearpage
\section*{Supplementary Tables}
\input{Contents/9_3_sup_table}

%% file: Contents/Supplementary/sup_FD.tex
\supnote{Finite difference}\label{sup:FD}

The finite difference method serves as a numerical approach for estimating derivatives, where differential operators are applied using predetermined kernels over structured grids \citep{Tadmor2012}. These kernels can be implemented efficiently through convolution operations, which simplify computation, leverage GPU parallelism, and fit seamlessly within modern deep learning libraries \citep{goodfellow2016deep}. Derivative values at each grid point $(i,j)$  are obtained by convolving the chosen field with finite-difference kernels across the two-dimensional domain, as formulated below.
\begin{align*}
    (f \ast w )(i,j) = \sum_{a=1}^A \sum_{b=1}^B \big[w(a,b) \times f(i+a-1,j+b-1) \big]
\end{align*}
where $f$ represents the target field and $w \in \mathbb{R}^{A\times B} $ is a convolution kernel that approximates a specific differential operator.

Within the proposed PIE-PINN framework, finite-difference operators are employed to evaluate the displacement-strain and equilibrium relations. Specifically, the strain components are obtained by applying convolution-based derivative operators to the predicted displacement field.
    \begin{equation*}
           {\varepsilon}^u_{xx}(i,j) = \sum_{a=1}^2 \sum_{b=1}^2  {w_{x}}(a,b)\hat{u}_x(i+a-1,j+b-1) 
    \end{equation*}
    \begin{equation*}
       {\varepsilon}^u_{yy}(i,j) = \sum_{a=1}^2 \sum_{b=1}^2  {w_{y}}(a,b)\hat{u}_y(i+a-1,j+b-1) 
    \end{equation*}
    \begin{equation*}
       {\varepsilon}^u_{xy}(i,j) = \sum_{a=1}^2 \sum_{b=1}^2  {w_{y}}(a,b)\hat{u}_x(i+a-1,j+b-1) + {w_{x}}(a,b)\hat{u}_y(i+a-1,j+b-1) 
    \end{equation*}
where ${w_{x}}$ and ${w_{y}}$ denote the finite-difference kernels associated with the spatial derivatives in the x- and y-directions, respectively, and are given by
    \begin{equation*}
        {w_{x}} = \begin{bmatrix}
                                -0.5 & 0.5 \\
                                -0.5 & 0.5 \\
                                \end{bmatrix} ,
                        {w_{y}} = \begin{bmatrix}
                                0.5 & 0.5\\
                                -0.5 & -0.5\\
                                \end{bmatrix}.  \notag
    \end{equation*}

Similarly, the equilibrium residuals are evaluated by applying convolution-based finite-difference operators to the stress field. The equilibrium residual at the grid point \((i, j)\) is computed as 
    \begin{align*}
        r(i,j) = & \sum_{a=1}^3 \sum_{b=1}^3 \{ {w_{xx}}(a,b)\sigma_{xx}(i+a-1,j+b-1) \notag \\
        & + {w_{yy}}(a,b)\sigma_{yy}(i+a-1,j+b-1) \notag \\
        & + {w_{xy}}(a,b)\sigma_{xy}(i+a-1,j+b-1) \}/ht ,\label{eq:residualPDE}
    \end{align*} 
where ${w_{xx}}$, ${w_{yy}}$, and ${w_{xy}}$ are the kernels for the finite difference to evaluate the stress components $\sigma_{xx}$, $\sigma_{yy}$, and $\sigma_{xy}$, respectively. $h$ and $t$ denote the vertical and horizontal spacing between neighboring displacement data points.
The kernels for $r_x$ are defined below.
    \begin{equation*}
    w_{xx} = \begin{bmatrix}
                            -1 & 0 & 1\\
                            -1 & 0 & 1\\
                            -1 & 0 & 1 \\
                            \end{bmatrix} ,
                    w_{yy} = \begin{bmatrix}
                            0 & 0 & 0\\
                            0 & 0 & 0\\
                            0 & 0 & 0 \\
                            \end{bmatrix} ,
                    w_{xy} = \begin{bmatrix}
                            1 & 1 & 1\\
                            0 & 0 & 0\\
                            -1 & -1 & -1 \\
                            \end{bmatrix} \label{eq:kernelResidual_x}
    \end{equation*}
and kernels for $r_y$ are defined below.
    \begin{equation*}
        w_{xx} = \begin{bmatrix}
                                0 & 0 & 0\\
                                0 & 0 & 0\\
                                0 & 0 & 0 \\
                                \end{bmatrix} ,
                        w_{yy} = \begin{bmatrix}
                                1 & 1 & 1\\
                                0 & 0 & 0\\
                                -1 & -1 & -1 \\
                                \end{bmatrix} ,
                        w_{xy} = \begin{bmatrix}
                                -1 & 0 & 1\\
                                -1 & 0 & 1\\
                                -1 & 0 & 1 \\
                                \end{bmatrix} 
                                \label{eq:kernelResidual_y}
    \end{equation*}

%% file: Contents/Supplementary/sup_dataDescriptions.tex
\supnote{Data Descriptions}\label{sup:dataDescription}

The simulation datasets are generated based on the publicly available data reported by Chen and Gu \citep{Chen2021, Chen2023}. To construct noisy and low-resolution displacement observations, the high-resolution reference displacement field is uniformly downsampled. Specifically, displacement measurements are retained at regularly spaced grid points, while the remaining points are discarded, resulting in sparse observation sets.

Additive Gaussian noise is then incorporated into the downsampled displacement field obtained from finite element analysis (FEA). Let $u(i,j)$ denote the noise-free displacement at spatial location $(i,j)$. The noisy observation $u^*(i,j)$ is generated as \citep{Gupta2018, Mafi2018, Krissian2005, Flaschel2021}
\begin{align}
u^*(i,j) = u(i,j) + \epsilon_{i,j},
\end{align}
where $\epsilon_{i,j}$ is a zero-mean Gaussian random variable. The noise standard deviation $\sigma$ is determined according to a prescribed signal-to-noise ratio (SNR), defined as
\begin{align*}
\sigma = \left| \frac{\bar{u}}{\mathrm{SNR}} \right|,
\end{align*}
where $\bar{u}$ denotes the mean magnitude of the displacement measurements.

%% file: Contents/9_2_sup_fig.tex
\begin{figure}[H] 
    \centering
    \includegraphics[width=\textwidth]{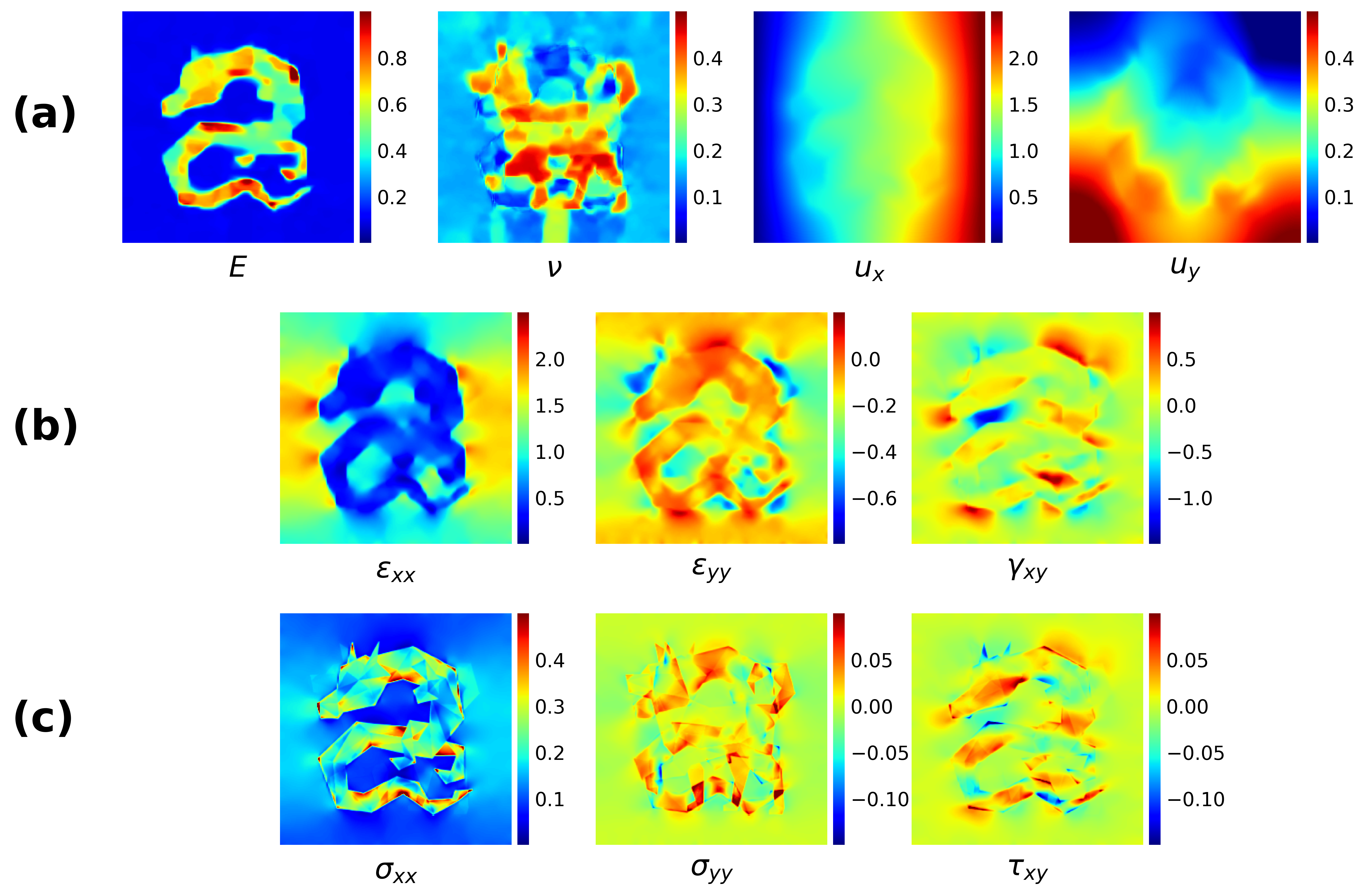}
    \justifying
    \caption{Prediction fields of predicted mechanical quantities from the proposed PIE-PINN under noisy, low-resolution displacement data (50\% resolution, SNR = 100). (a) Predicted Young’s modulus (MPa), Poisson’s ratio, and axial displacement (mm) fields. (b) Predicted strain fields (\%). (c) Predicted stress fields (MPa).}
\label{fig:snr100-n2-pred}
\end{figure}

\clearpage \newpage
\begin{figure}[H] 
    \centering
    \includegraphics[width=\textwidth]{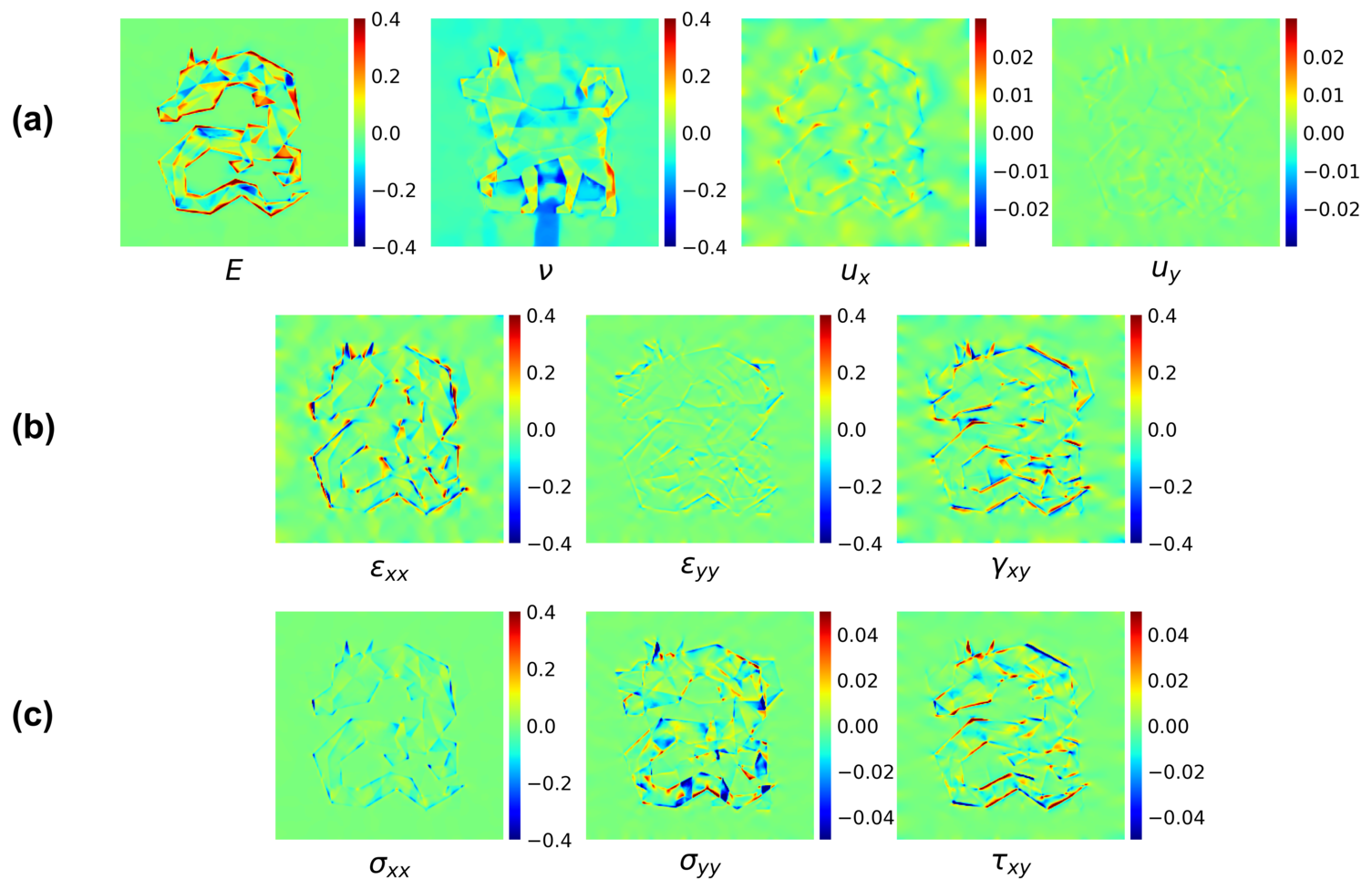}
    \justifying
    \caption{Error fields of predicted mechanical quantities from the proposed PIE-PINN under noisy, low-resolution displacement data (50\% resolution, SNR = 100). (a) Predicted Young’s modulus (MPa), Poisson’s ratio, and axial displacement (mm) fields. (b) Predicted strain fields (\%). (c) Predicted stress fields (MPa).}
\label{fig:snr100-n2-error}
\end{figure}

\clearpage \newpage

\begin{figure}[H] 
    \centering
    \begin{subfigure}[t!]{0.49\textwidth}
        \includegraphics[width=1.0\textwidth]{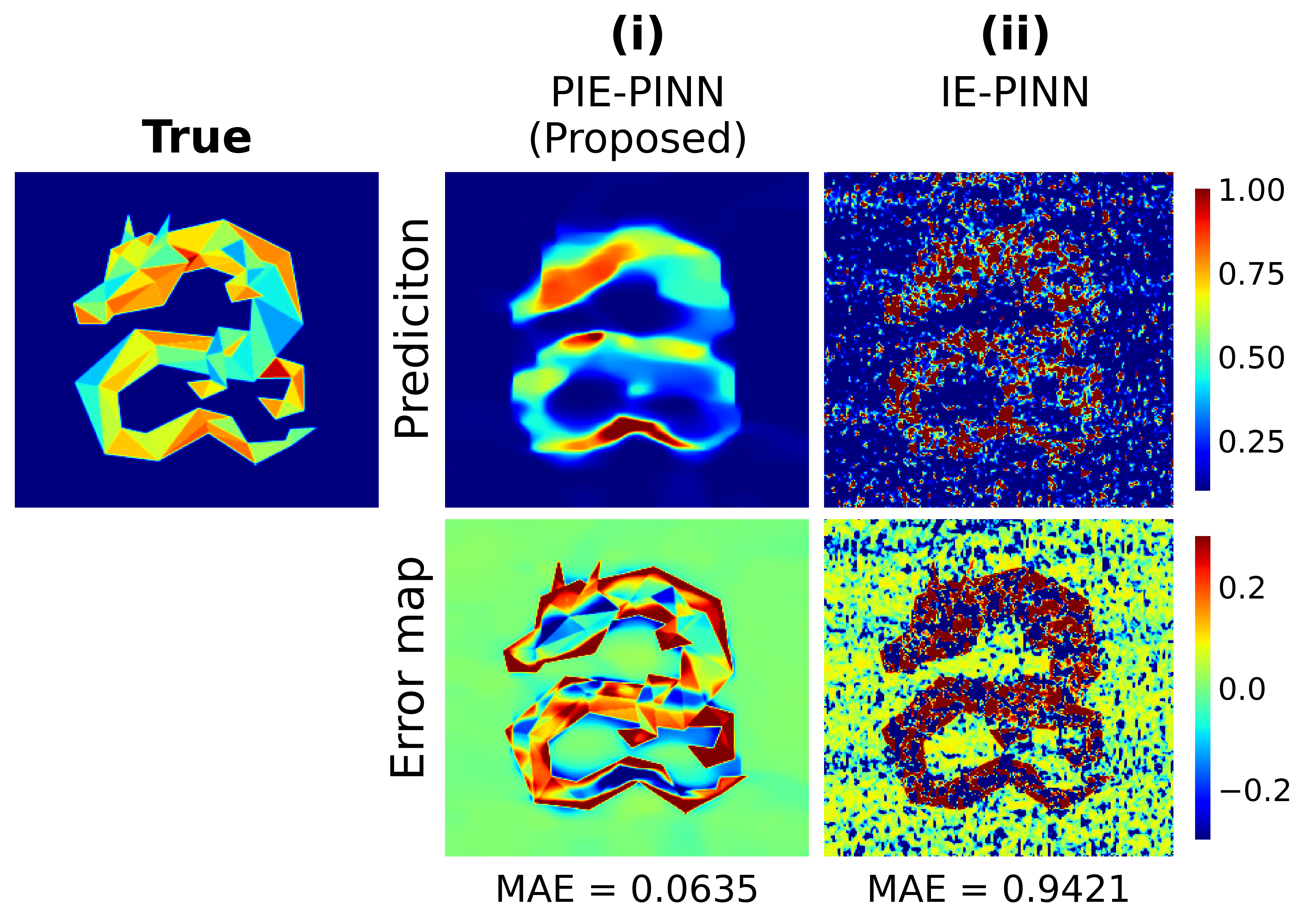}
        \caption{Estimated Young's modulus}
        \label{Fig:Benchmark_n2_snr20_E}
    \end{subfigure}
    \vspace{0.5em}
    \begin{subfigure}[t!]{0.49\textwidth}
        \includegraphics[width=1.0\textwidth]{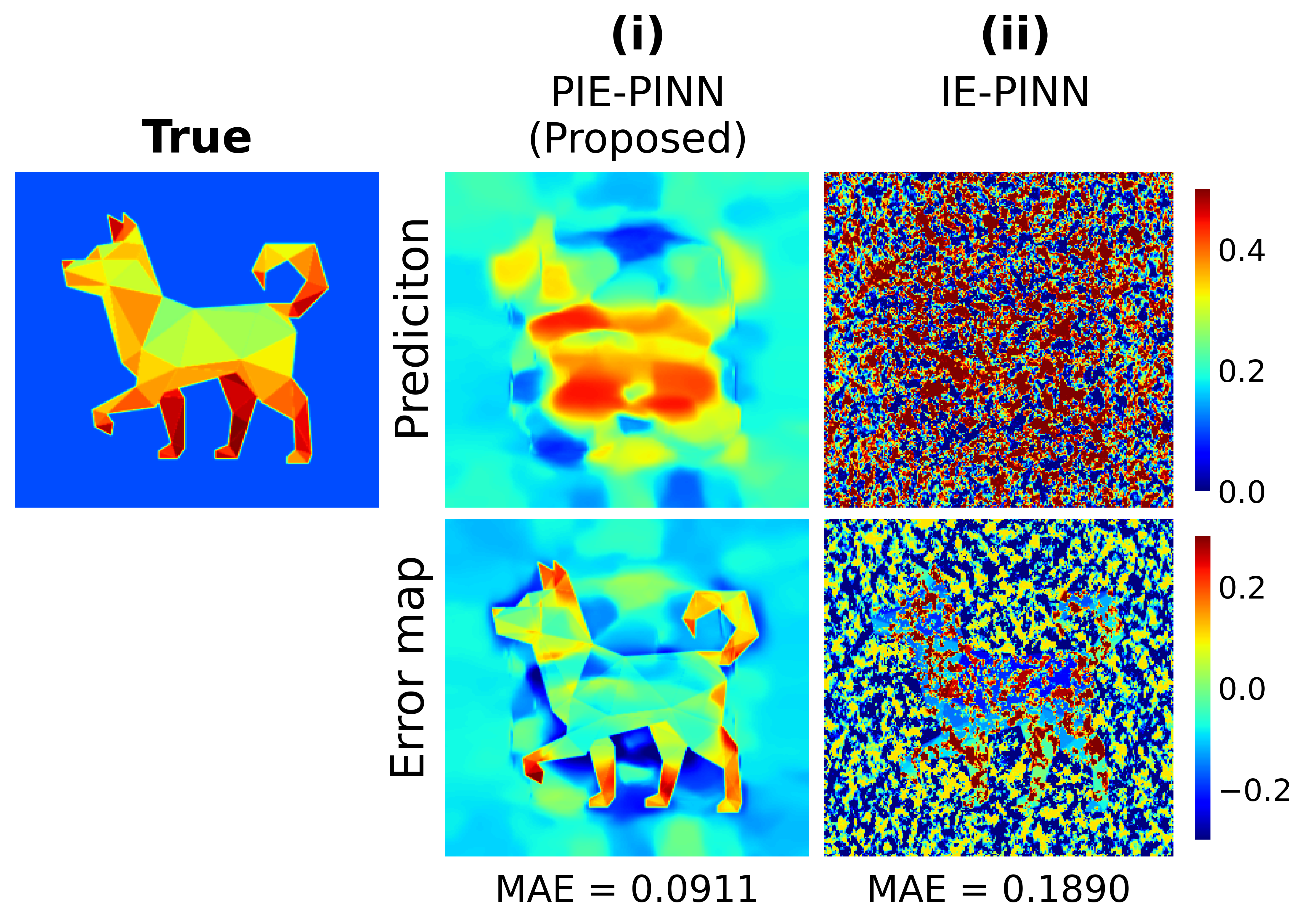}
        \caption{Estimated Poisson's ratio }
        \label{Fig:Benchmark_n2_snr20_v}
    \end{subfigure}
    \caption{Comparison of elasticity maps from different models inferred from noisy, low-resolution displacement data (50\% resolution, SNR = 20). }
    \label{Fig:Benchmark_n2_snr20} 
\end{figure}

\begin{figure}[H] 
    \centering
    \begin{subfigure}[t!]{0.49\textwidth}
        \includegraphics[width=1.0\textwidth]{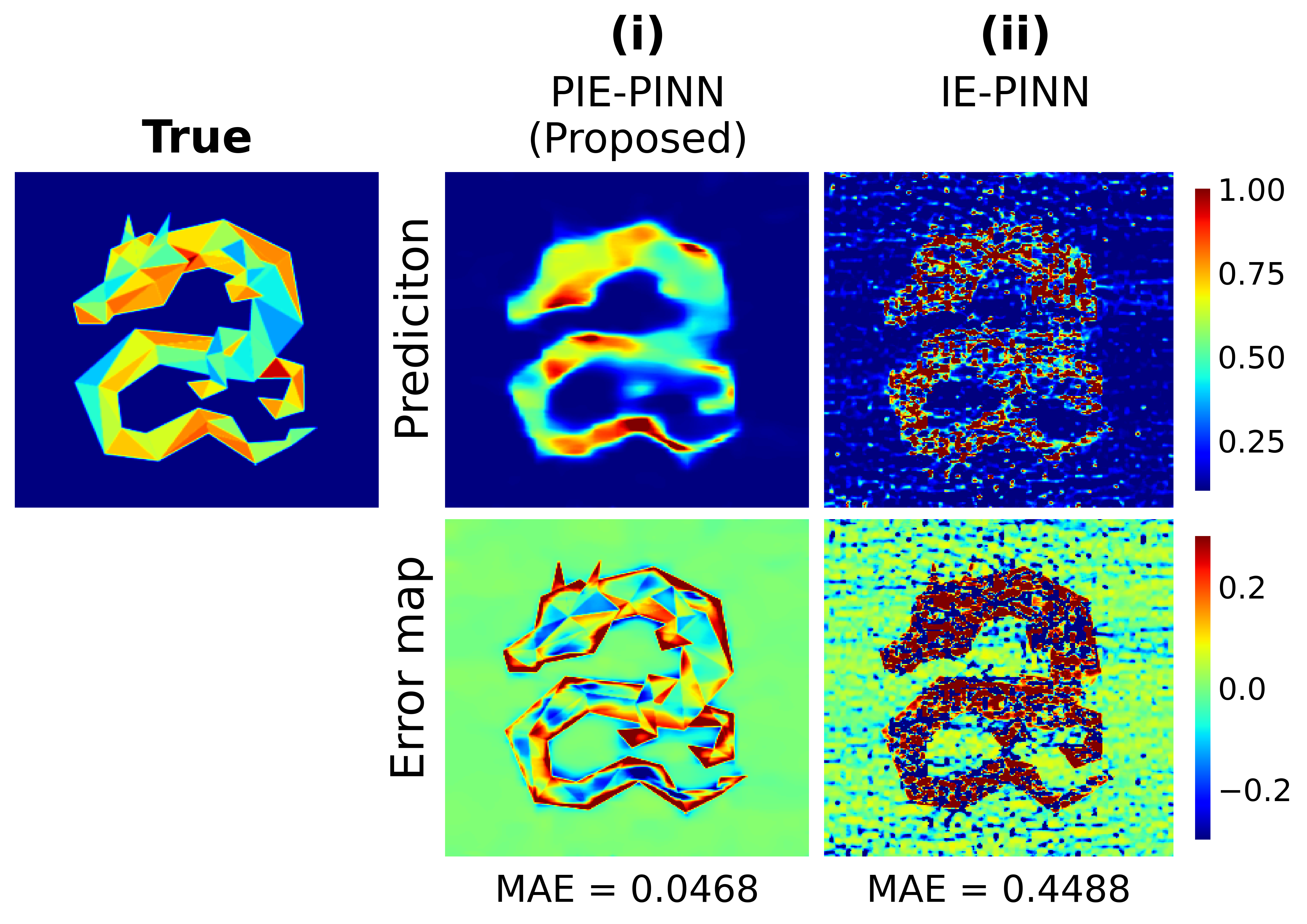}
        \caption{Estimated Young's modulus}
        \label{Fig:Benchmark_n4_snr100_E}
    \end{subfigure}
    \vspace{0.5em}
    \begin{subfigure}[t!]{0.49\textwidth}
        \includegraphics[width=1.0\textwidth]{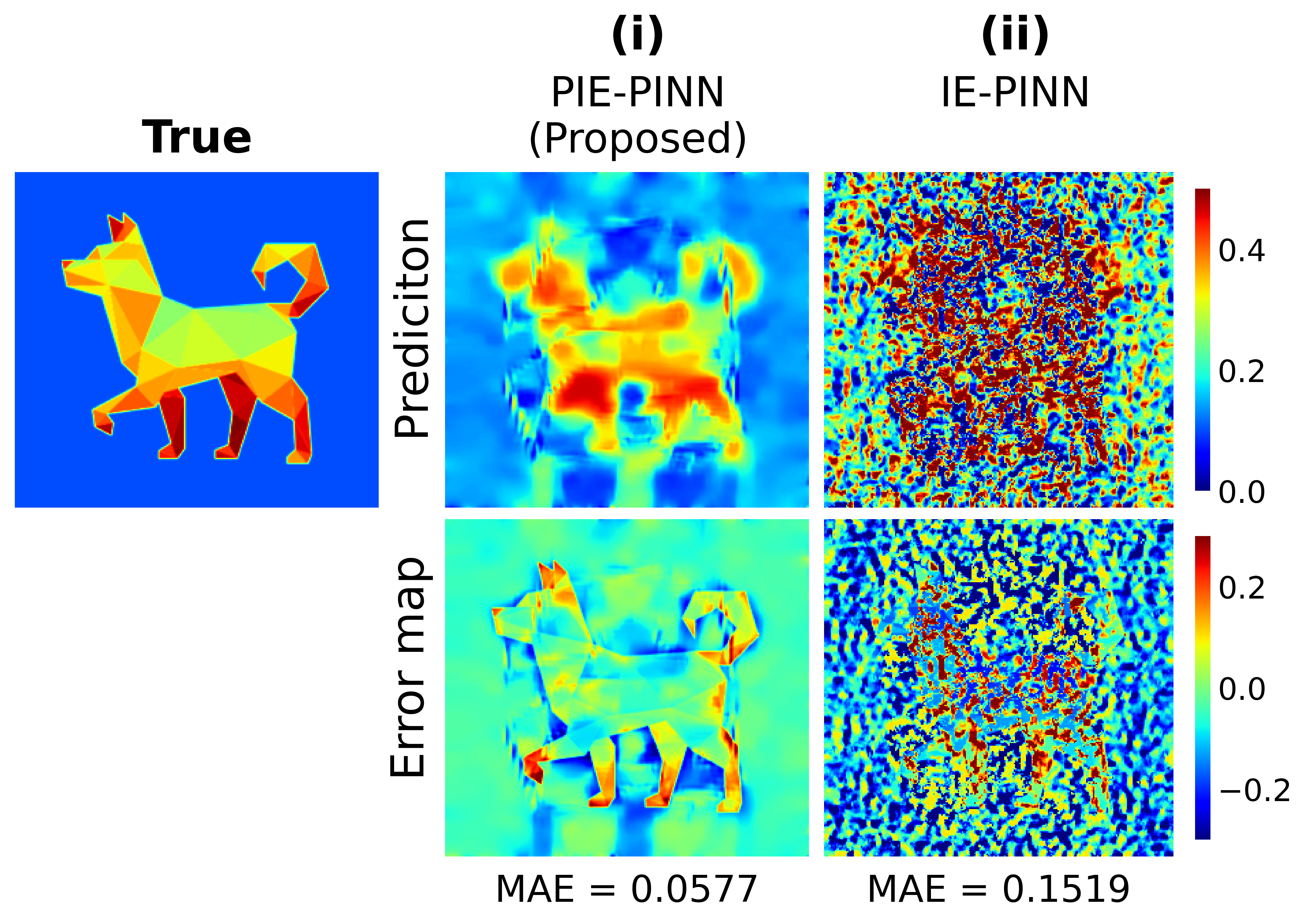}
        \caption{Estimated Poisson's ratio }
        \label{Fig:Benchmark_n4_snr100_v}
    \end{subfigure}
    \caption{Comparison of elasticity maps from different models inferred from noisy, low-resolution displacement data (25\% resolution, SNR = 100). }
    \label{Fig:Benchmark_n4_snr100} 
\end{figure}

\clearpage \newpage

\begin{figure}[t!]
    \centering
    \begin{subfigure}[t!]{0.49\textwidth}
        \includegraphics[width=1.0\textwidth]{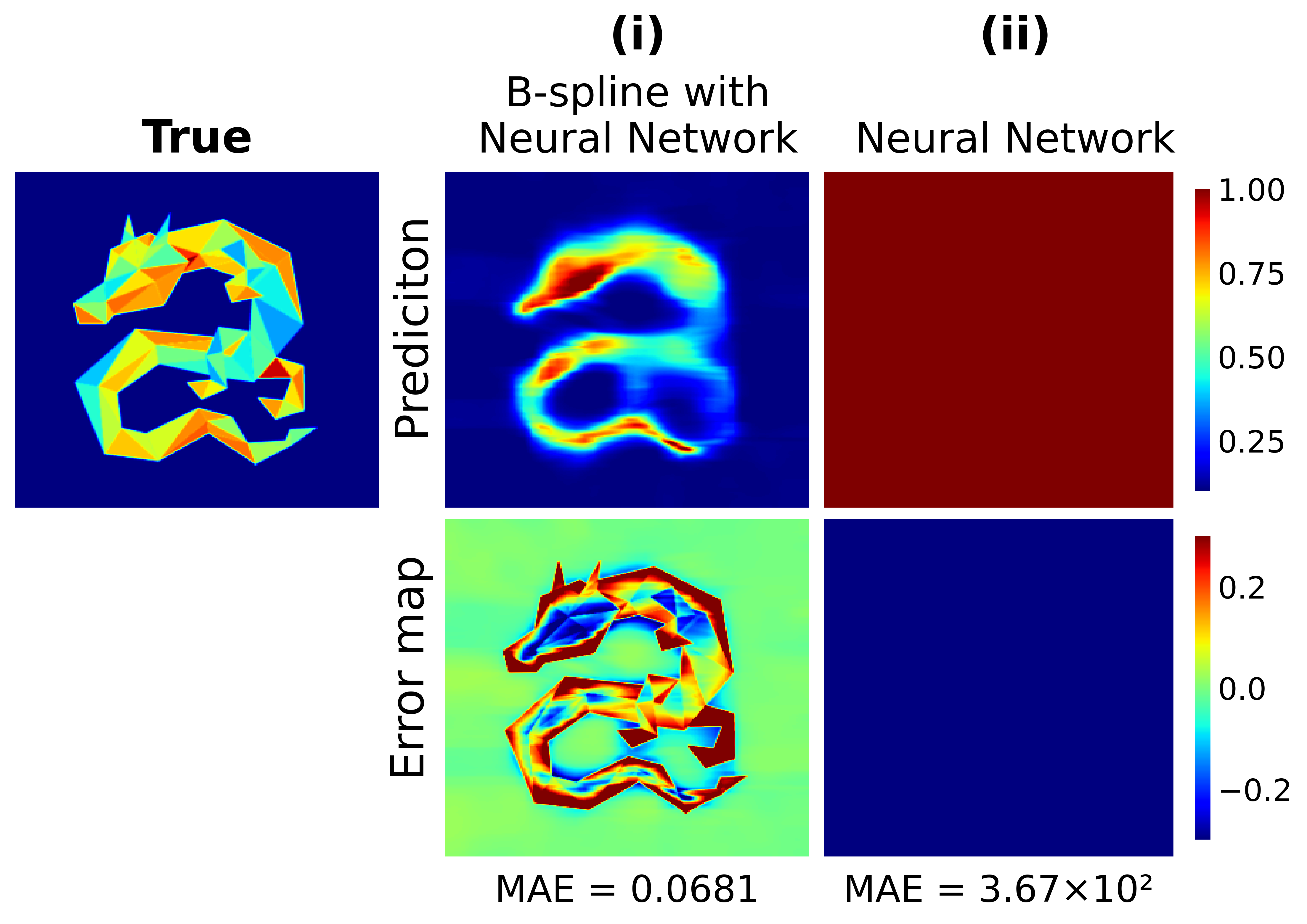}
        \caption{Estimated Young's modulus}
        \label{Fig:importanceBspline_n6_E}
    \end{subfigure}
    \vspace{0.5em}
    \begin{subfigure}[t!]{0.49\textwidth}
        \includegraphics[width=1.0\textwidth]{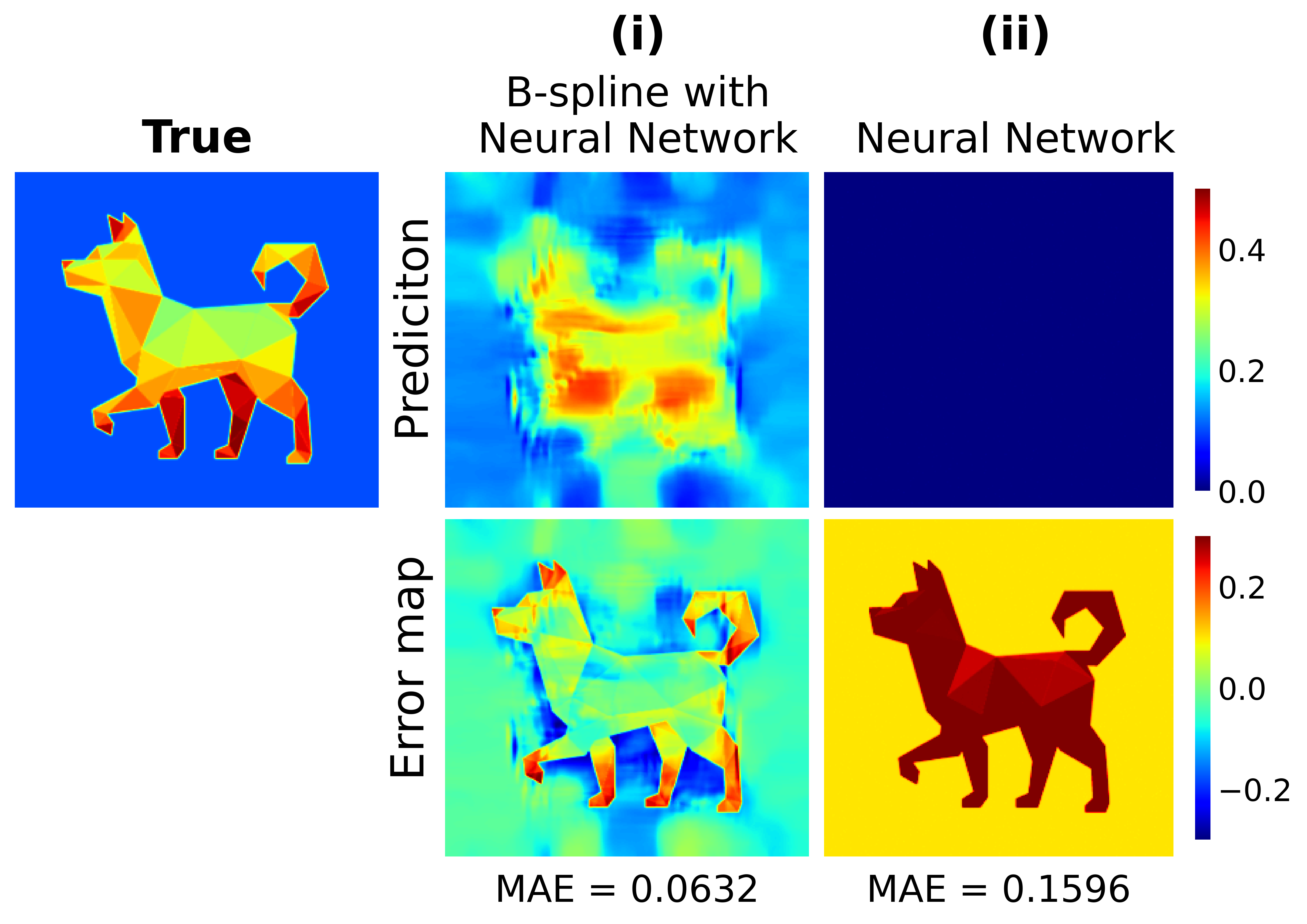}
        \caption{Estimated Poisson's ratio}
        \label{Fig:importanceBspline_n6_v}
    \end{subfigure}
    \caption{Comparison of elasticity maps inferred from noisy, low-resolution displacement data (17.3\% resolution, SNR = 100) between the proposed B-spline-guided neural network architecture and the standard neural network architecture.}
    \label{Fig:importanceBspline_n6} 
\end{figure}

\begin{figure}[t!]
    \centering
    \begin{subfigure}[t!]{0.49\textwidth}
        \includegraphics[width=1.0\textwidth]{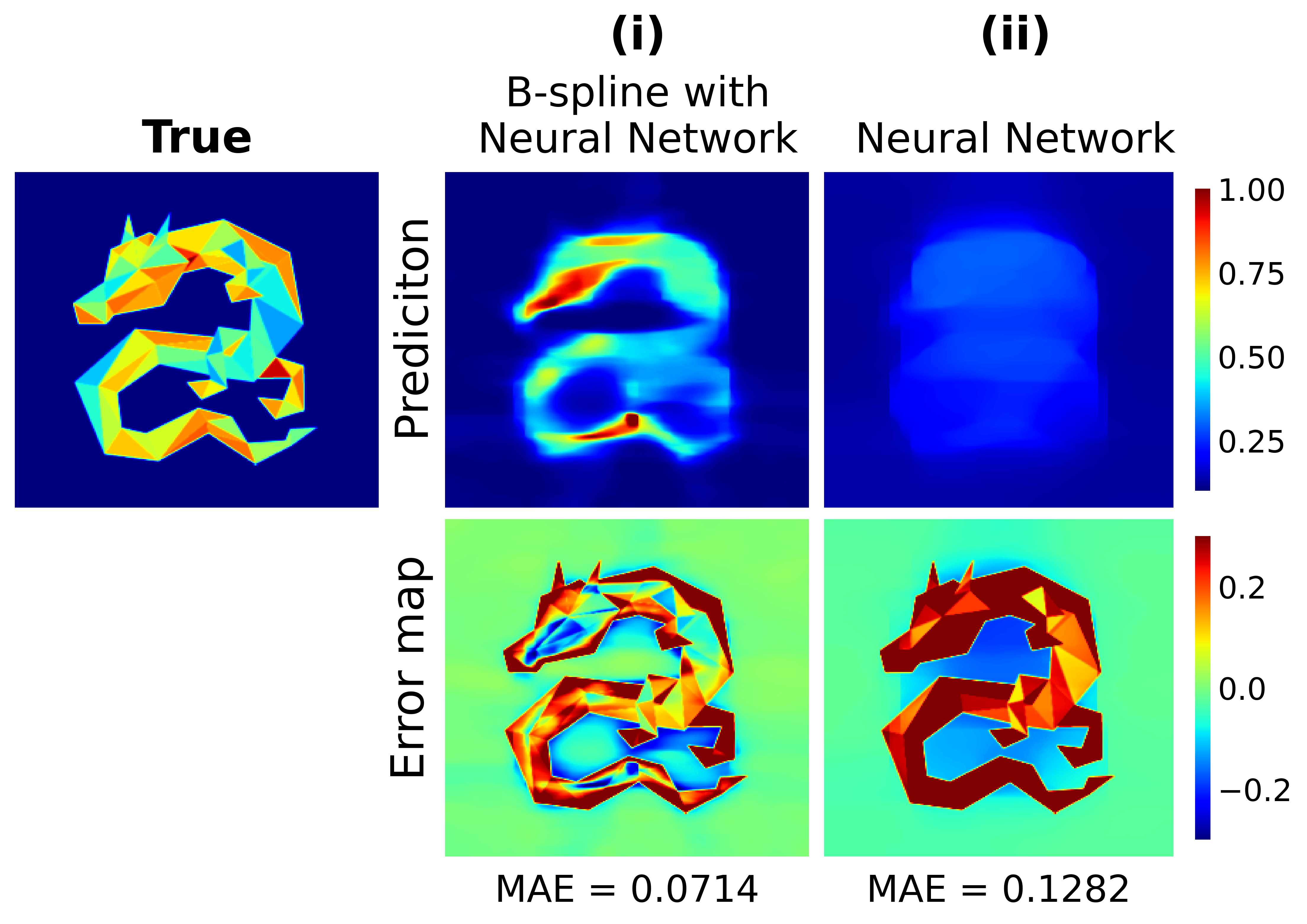}
        \caption{Estimated Young's modulus}
        \label{Fig:importanceBspline_n8_E}
    \end{subfigure}
    \vspace{0.5em}
    \begin{subfigure}[t!]{0.49\textwidth}
        \includegraphics[width=1.0\textwidth]{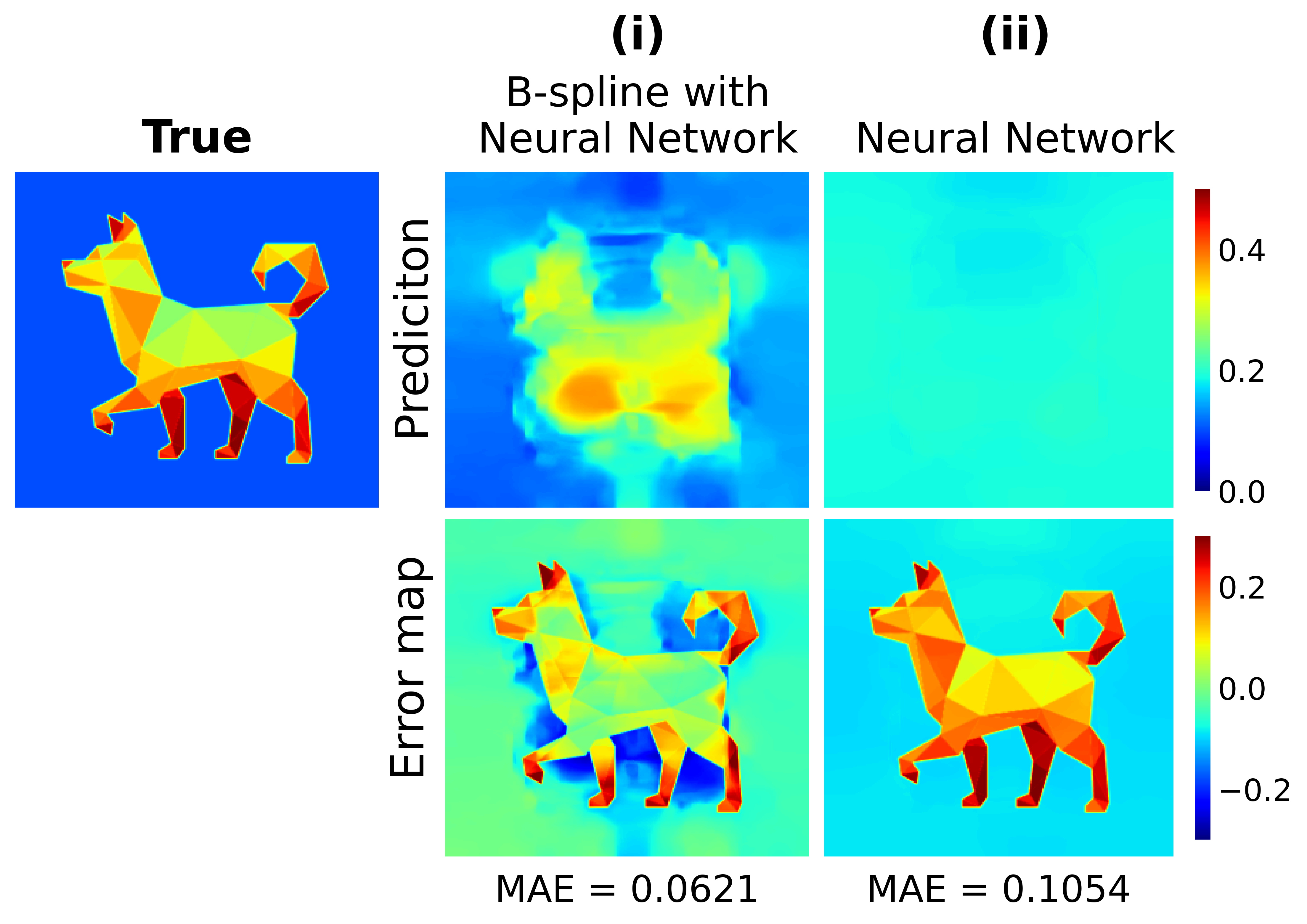}
        \caption{Estimated Poisson's ratio}
        \label{Fig:importanceBspline_n8_v}
    \end{subfigure}
    \caption{Comparison of elasticity maps inferred from noisy, low-resolution displacement data (12.5\% resolution, SNR = 100) between the proposed B-spline-guided neural network architecture and the standard neural network architecture.}
    \label{Fig:importanceBspline_n8} 
\end{figure}

\clearpage \newpage

\begin{figure}[t!]
    \centering
    \includegraphics[width=1.0\textwidth]{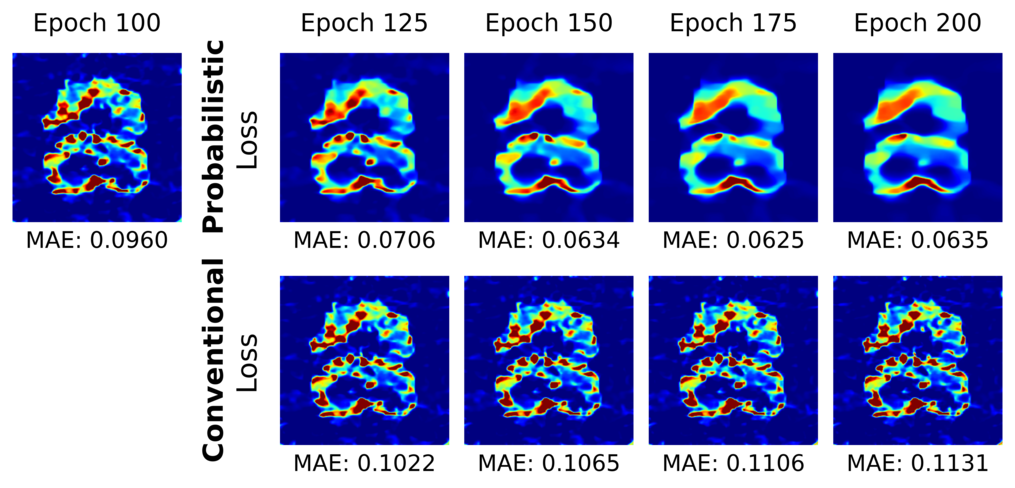}
    \justifying
    \caption{Evolution of Young’s modulus estimates inferred from noisy, low-resolution displacement data (50\% spatial resolution, SNR = 20) using the proposed architecture. Results are compared between the conventional deterministic loss and the proposed probabilistic loss combined with the EM-style adaptive weight algorithm.}
\label{fig:evolution_n2_snr20_E}
\end{figure}


\clearpage \newpage

\begin{figure}[t!]
    \centering
    \includegraphics[width=1.0\textwidth]{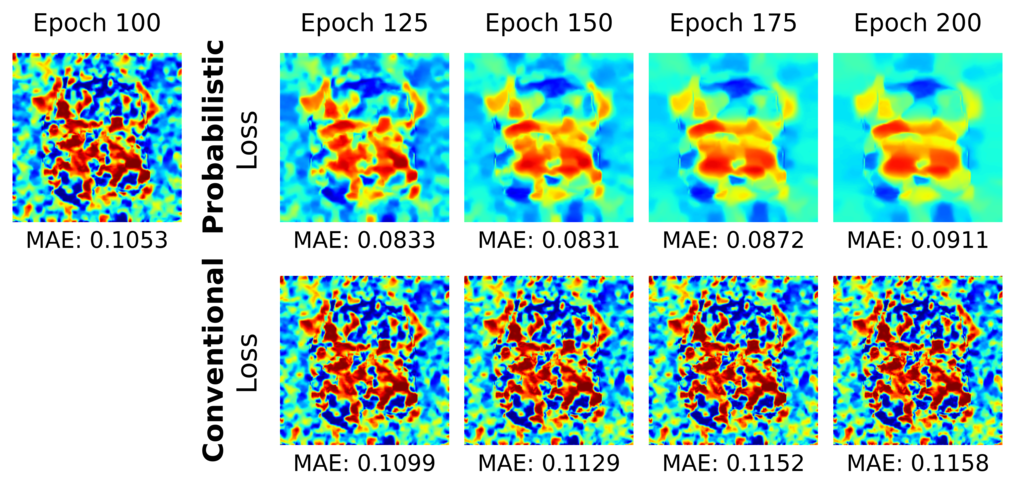}
    \justifying
    \caption{Evolution of Poisson's ratio estimates inferred from noisy, low-resolution displacement data (50\% spatial resolution, SNR = 20) using the proposed architecture. Results are compared between the conventional deterministic loss and the proposed probabilistic loss combined with the EM-style adaptive weight algorithm.}
\label{fig:evolution_n2_snr20_v}
\end{figure}


\clearpage \newpage

\begin{figure}[t!]
    \centering
    \begin{subfigure}[t!]{0.49\textwidth}
        \includegraphics[width=1.0\textwidth]{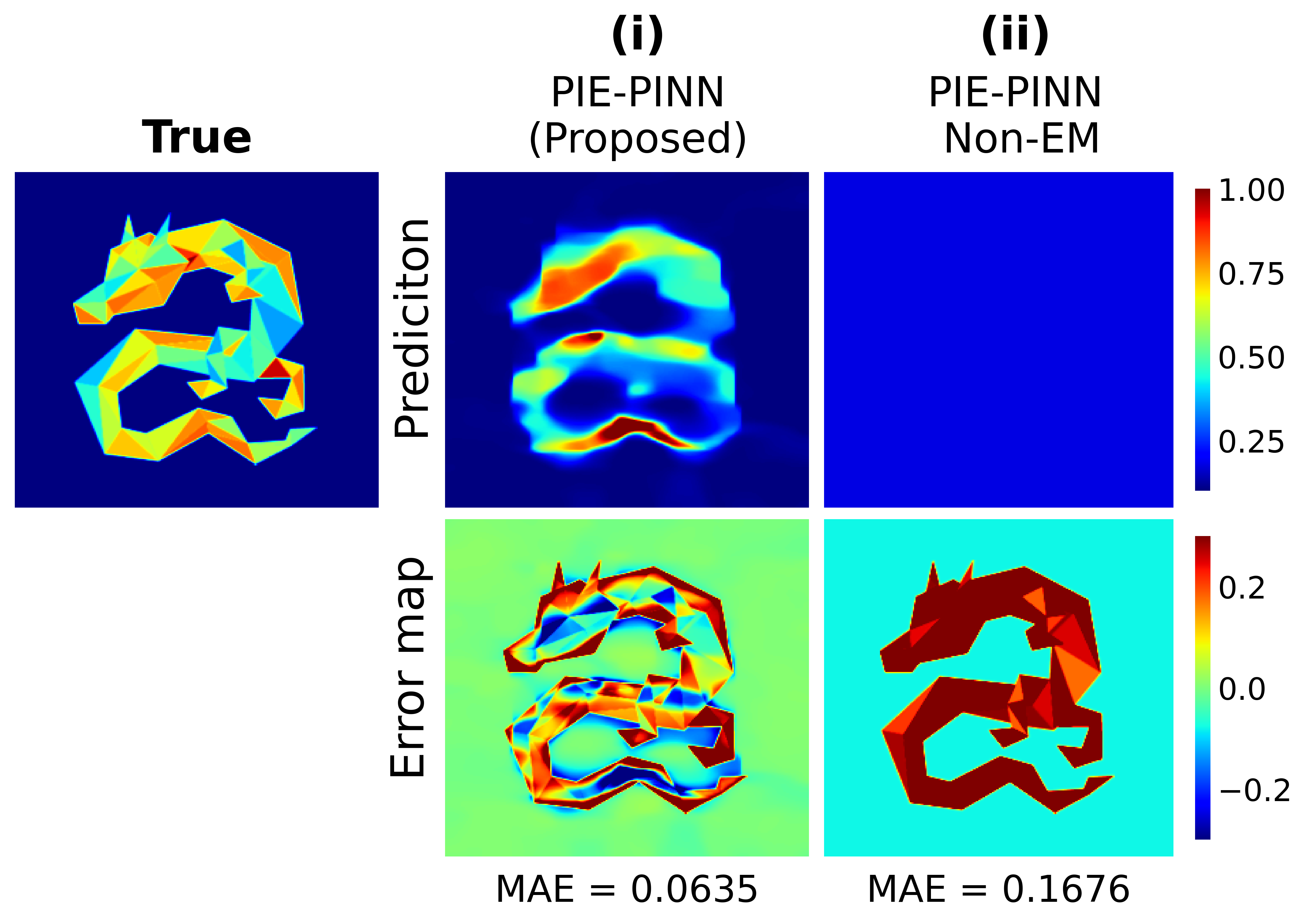}
        \caption{Estimated Young's modulus}
        \label{fig:ProblossWithEM_n2_snr20_E}
    \end{subfigure}
    \vspace{0.5em}
    \begin{subfigure}[t!]{0.49\textwidth}
        \includegraphics[width=1.0\textwidth]{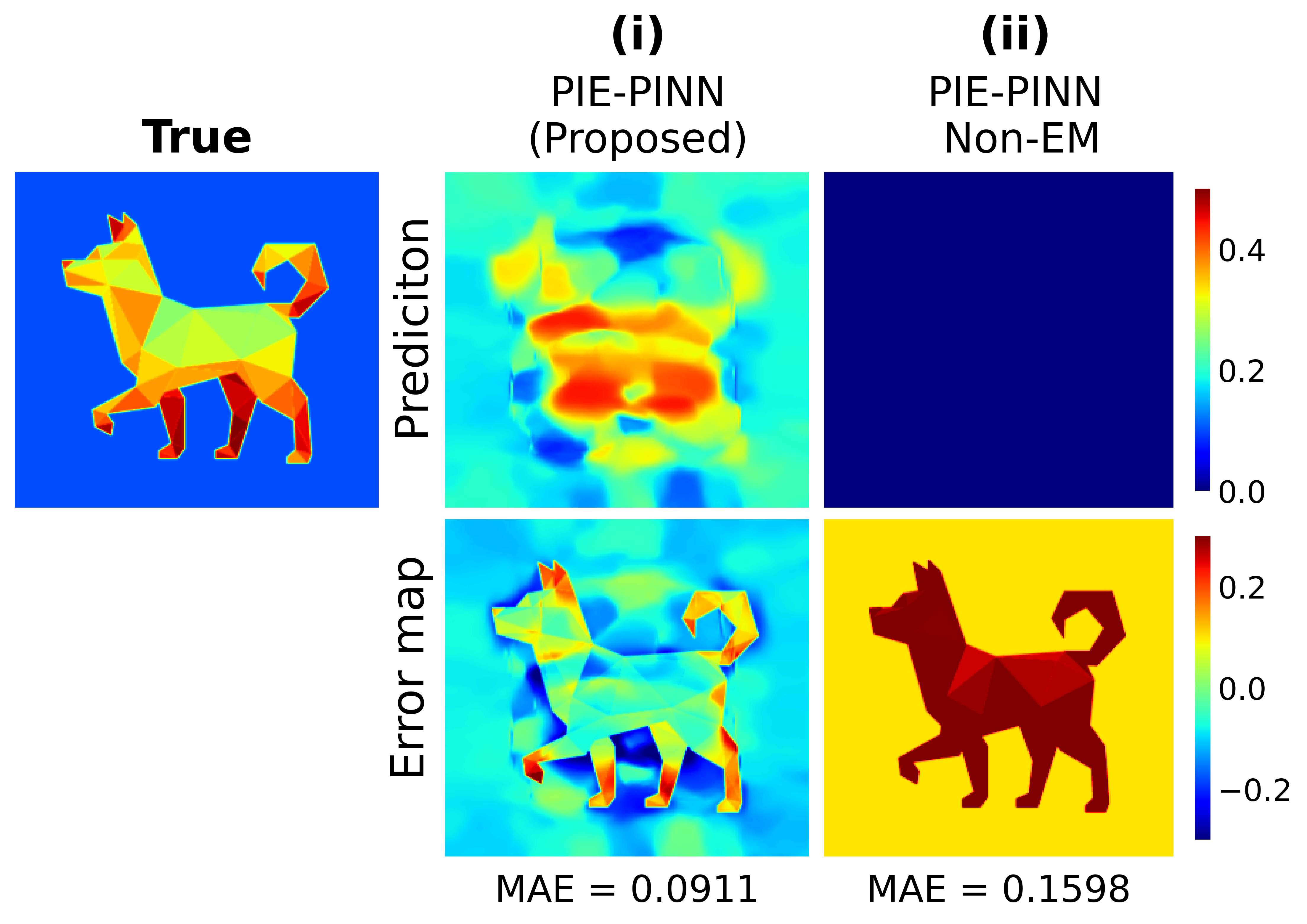}
        \caption{Estimated Poisson's ratio}
        \label{fig:ProblossWithEM_n2_snr20_v}
    \end{subfigure}
    \caption{Comparison of elasticity maps inferred from noisy, low-resolution displacement data (50\% resolution, SNR = 20) using the proposed model with the probabilistic loss, with and without the EM-style adaptive weight algorithm.}
    \label{fig:ProblossWithEM_n2_snr20} 
\end{figure}

\clearpage \newpage

\begin{figure}[t!]
    \centering
    \begin{subfigure}[t!]{0.49\textwidth}
        \includegraphics[width=1.0\textwidth]{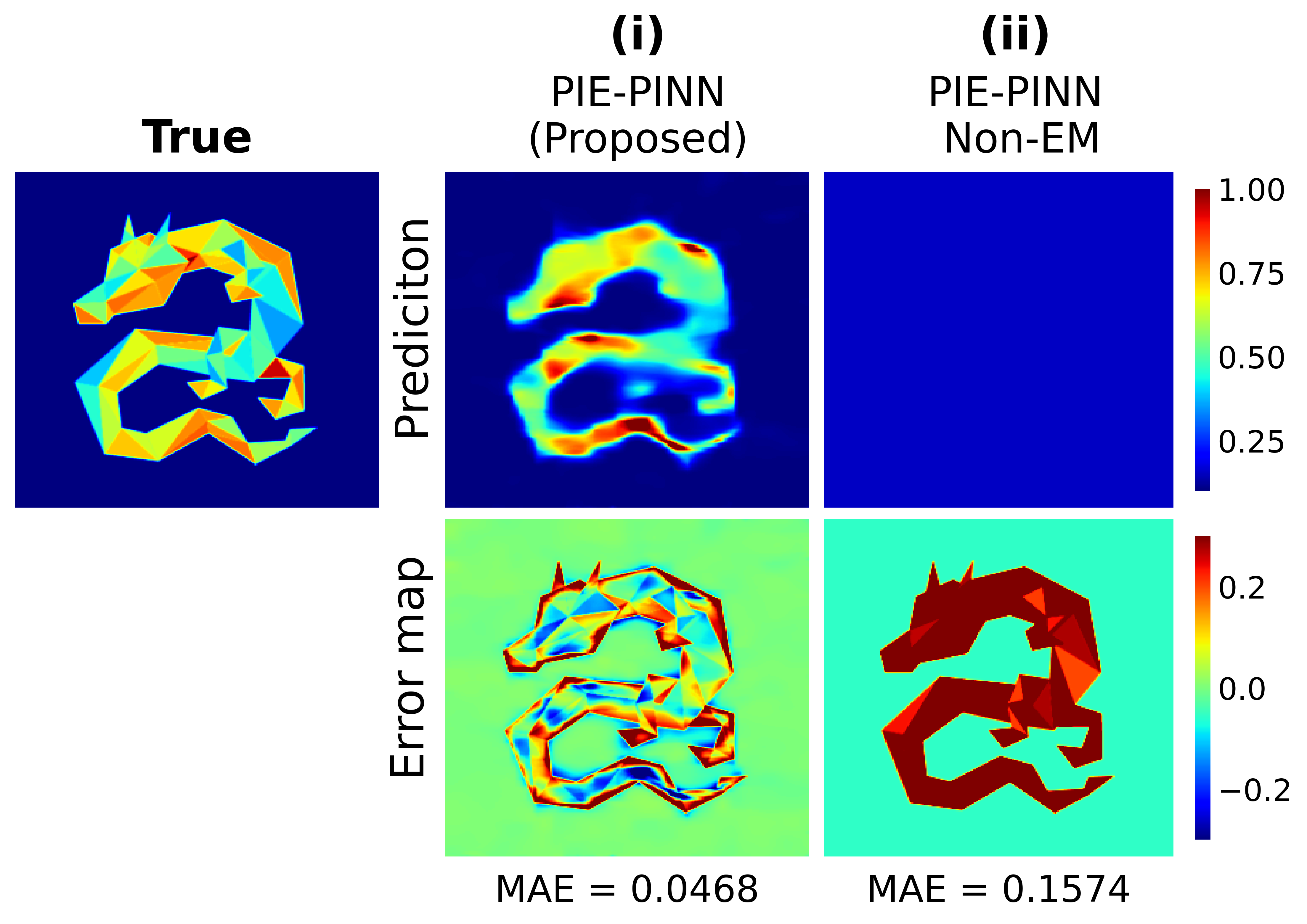}
        \caption{Estimated Young's modulus}
        \label{fig:ProblossWithEM_n4_snr100_E}
    \end{subfigure}
    \vspace{0.5em}
    \begin{subfigure}[t!]{0.49\textwidth}
        \includegraphics[width=1.0\textwidth]{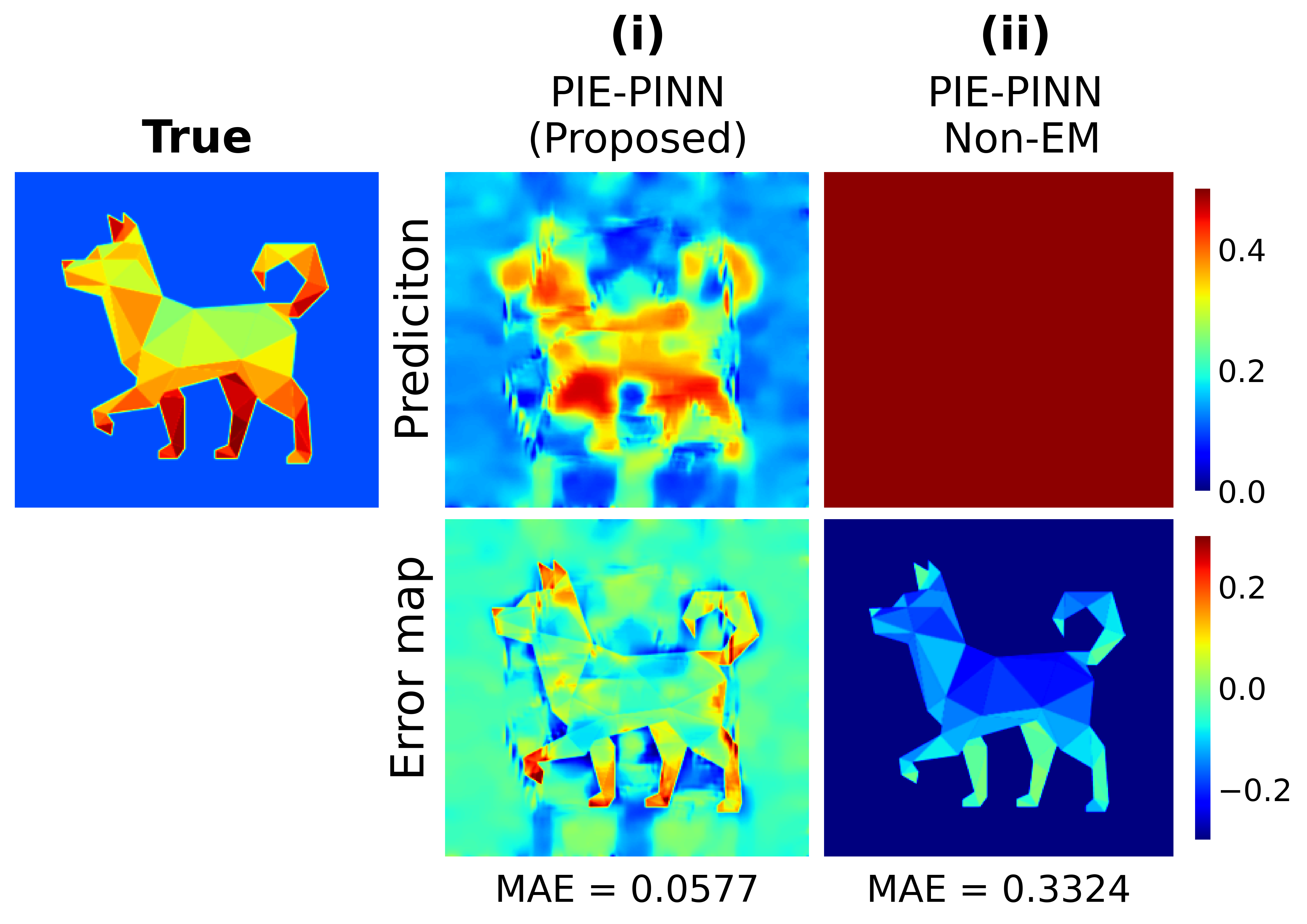}
        \caption{Estimated Poisson's ratio}
        \label{fig:ProblossWithEM_n4_snr100_v}
    \end{subfigure}
    \caption{Comparison of elasticity maps inferred from noisy, low-resolution displacement data (25\% resolution, SNR = 100) using the proposed model with the probabilistic loss, with and without the EM-style adaptive weight algorithm.}
    \label{fig:ProblossWithEM_n4_snr100} 
\end{figure}

%% file: Contents/9_3_sup_table.tex
\begin{table}[ht]
\centering
\caption{\textbf{Training parameters of PIE-PINN} \label{table:param_piepinn}}
{
\begin{tabular}{p{8.0 cm}|p{6.0cm}}
\midrule
\parbox{8.0cm}{ Parameter}& \parbox{6.0cm}{Value}  \\ 
\midrule
\parbox{8.0cm}{\textit{Common}}& \parbox{6.0cm}{}  \\ 
\parbox{8.0cm}{\quad learning rate ($\alpha$) }& \parbox{6.0cm}{$ 10^{-3}$}  \\ 
\parbox{8.0cm}{\quad optimizer}& \parbox{6.0cm}{Adam}  \\
\midrule
\parbox{8.0cm}{\textit{B-spline-guided Displacement network architecture}}& \parbox{6.0cm}{}  \\ 
\parbox{8.0cm}{\quad degree of the B-spline }& \parbox{6.0cm}{2}  \\
\parbox{8.0cm}{\quad number of basis functions}& \parbox{6.0cm}{dimension of observation}  \\
\parbox{8.0cm}{\quad network structure}& \parbox{6.0cm}{fully connected neural network}  \\
\parbox{8.0cm}{\quad number of hidden layers}& \parbox{6.0cm}{16}  \\
\parbox{8.0cm}{\quad number of neurons per layer}& \parbox{6.0cm}{128}  \\
\parbox{8.0cm}{\quad linearity}& \parbox{6.0cm}{SIREN \citep{Vaswani2017}}  \\
\parbox{8.0cm}{\quad activation function of output}& \parbox{6.0cm}{-}  \\
\midrule
\parbox{8.0cm}{\textit{Strain network architecture}}& \parbox{6.0cm}{}  \\ 
\parbox{8.0cm}{\quad network structure}& \parbox{6.0cm}{fully connected neural network}  \\
\parbox{8.0cm}{\quad number of hidden layers}& \parbox{6.0cm}{16}  \\
\parbox{8.0cm}{\quad number of neurons per layer}& \parbox{6.0cm}{128}  \\
\parbox{8.0cm}{\quad linearity}& \parbox{6.0cm}{SIREN \citep{Vaswani2017}}  \\
\parbox{8.0cm}{\quad activation function of output}& \parbox{6.0cm}{-}  \\
\midrule
\parbox{8.0cm}{\textit{Elasticity network architecture}}& \parbox{6.0cm}{}  \\ 
\parbox{8.0cm}{\quad network structure}& \parbox{6.0cm}{fully connected neural network}  \\
\parbox{8.0cm}{\quad number of hidden layers}& \parbox{6.0cm}{16}  \\
\parbox{8.0cm}{\quad number of neurons per layer}& \parbox{6.0cm}{128}  \\
\parbox{8.0cm}{\quad linearity}& \parbox{6.0cm}{SIREN \citep{Vaswani2017}}  \\
\parbox{8.0cm}{\quad activation function of output}& \parbox{6.0cm}{Softplus function}  \\
\midrule
\parbox{8.0cm}{\textit{Initial weighting scheme}}& \parbox{6.0cm}{}  \\ 
\parbox{8.0cm}{\quad weight of displacement fitting loss ($\lambda_u$)}& \parbox{6.0cm}{1}  \\
\parbox{8.0cm}{\quad weight of of strain fitting loss ($\lambda_{\varepsilon}$)}& \parbox{6.0cm}{1}  \\
\parbox{8.0cm}{\quad weight of residual of equilibrium loss ($\lambda_r$)}& \parbox{6.0cm}{1}  \\
\parbox{8.0cm}{\quad weight of mean modulus constraint loss ($\lambda_E$)}& \parbox{6.0cm}{0.1}  \\
\bottomrule
\end{tabular}
}
{}
\end{table}

\begin{table}[ht]
\centering
\caption{\textbf{Computational resources for PIE-PINN to estimate elasticity maps inferred from noisy, low-resolution displacement data (50\% resolution, SNR = 100)}}\label{table:resourceComputation}
{
\begin{tabular}{p{6.0 cm}|p{2.2cm}|p{2.8cm}|p{2.8cm}}
\midrule
\parbox{6.0cm}{Phase} 
    & \parbox{2.2cm}{Elapsed time (h:mm)} 
    & \parbox{2.8cm}{Peak usage\\ memory (MB)} 
    & \parbox{2.8cm}{Allocated usage\\ memory (MB)} \\ 
\midrule
    \parbox{6.0cm}{\textit{Pretraining stage}}
    & \parbox{2.2cm}{}
    & \parbox{2.8cm}{}
    & \parbox{2.8cm}{}  \\ 
    \parbox{6.0cm}{\quad Displacement fitting}
    & \parbox{2.2cm}{0:15}
    & \parbox{2.8cm}{390.31}
    & \parbox{2.8cm}{644.00}  \\ 
    \parbox{6.0cm}{\quad Strain fitting}
    & \parbox{2.2cm}{1:45}
    & \parbox{2.8cm}{2484.98}
    & \parbox{2.8cm}{2730.00}  \\
\midrule
\parbox{6.0cm}{\textit{Physic-informed training stage}}
    & \parbox{2.2cm}{}
    & \parbox{2.8cm}{}
    & \parbox{2.8cm}{}  \\ 
    \parbox{6.0cm}{\quad Elasticity learning}
    & \parbox{2.2cm}{11:00}
    & \parbox{2.8cm}{4404.12}
    & \parbox{2.8cm}{4726.0}  \\
\bottomrule
\end{tabular}
}
{}
\end{table}